\documentclass{article} 
\usepackage{iclr2025_conference,times}

\usepackage{amsmath,amsfonts,bm}

\def\eqref#1{equation~\ref{#1}}

\def\1{\bm{1}}

\def\vr{{\bm{r}}}
\def\vs{{\bm{s}}}

\DeclareMathAlphabet{\mathsfit}{\encodingdefault}{\sfdefault}{m}{sl}
\SetMathAlphabet{\mathsfit}{bold}{\encodingdefault}{\sfdefault}{bx}{n}
\newcommand{\tens}[1]{\bm{\mathsfit{#1}}}
\def\tA{{\tens{A}}}
\def\tB{{\tens{B}}}
\def\tC{{\tens{C}}}
\def\tD{{\tens{D}}}
\def\tE{{\tens{E}}}

\usepackage{hyperref}
\usepackage{url}
\usepackage{adjustbox}
\usepackage{graphicx}
\usepackage{subcaption}
\usepackage{booktabs}
\usepackage{multirow}
\usepackage{booktabs}
\usepackage{wrapfig}

\title{PointNet with KAN versus PointNet with MLP for 3D Classification and Segmentation of Point Sets}

\author{Antiquus S.~Hippocampus, Natalia Cerebro \& Amelie P. Amygdale \thanks{ Use footnote for providing further information
about author (webpage, alternative address)---\emph{not} for acknowledging
funding agencies.  Funding acknowledgements go at the end of the paper.} \\
Department of Computer Science\\
Cranberry-Lemon University\\
Pittsburgh, PA 15213, USA \\
\texttt{\{hippo,brain,jen\}@cs.cranberry-lemon.edu} \\
\And
Ji Q. Ren \& Yevgeny LeNet \\
Department of Computational Neuroscience \\
University of the Witwatersrand \\
Joburg, South Africa \\
\texttt{\{robot,net\}@wits.ac.za} \\
\AND
Coauthor \\
Affiliation \\
Address \\
\texttt{email}
}

\author{Ali Kashefi \\
Stanford University, Stanford, 94305, CA, USA \\
\texttt{kashefi@stanford.edu}\\
\url{https://github.com/Ali-Stanford/PointNet_KAN_Graphic}
}

\iclrfinalcopy 
\begin{document}

\maketitle

\begin{abstract}
Kolmogorov-Arnold Networks (KANs) have recently gained attention as an alternative to traditional Multilayer Perceptrons (MLPs) in deep learning frameworks. KANs have been integrated into various deep learning architectures such as convolutional neural networks, graph neural networks, and transformers, with their performance evaluated. However, their effectiveness within point-cloud-based neural networks remains unexplored. To address this gap, we incorporate KANs into PointNet for the first time to evaluate their performance on 3D point cloud classification and segmentation tasks. Specifically, we introduce PointNet-KAN, built upon two key components. First, it employs KANs instead of traditional MLPs. Second, it retains the core principle of PointNet by using shared KAN layers and applying symmetric functions for global feature extraction, ensuring permutation invariance with respect to the input features. In traditional MLPs, the goal is to train the weights and biases with fixed activation functions; however, in KANs, the goal is to train the activation functions themselves. We use Jacobi polynomials to construct the KAN layers. We extensively and systematically evaluate PointNet-KAN across various polynomial degrees and special types such as the Lagrange, Chebyshev, and Gegenbauer polynomials. Our results show that PointNet-KAN achieves competitive performance compared to PointNet with MLPs on benchmark datasets for 3D object classification and part and semantic segmentation, despite employing a shallower and simpler network architecture. We also study a hybrid PointNet model incorporating both KAN and MLP layers. We hope this work serves as a foundation and provides guidance for integrating KANs, as an alternative to MLPs, into more advanced point cloud processing architectures.
\end{abstract}

\section{Introduction}
\label{Sect1}

Kolmogorov-Arnold Networks (KANs), introduced by \citet{liu2024kan}, have recently emerged as an alternative modeling framework to traditional Multilayer Perceptrons (MLPs) \citep{cybenko1989approximation,hornik1989multilayer}. KANs are based on the Kolmogorov-Arnold representation theorem \citep{kolmogorovSuperposition,arnold2009functions}. Unlike MLPs, which rely on fixed activation functions while training weights and biases, the objective in KANs is to train the activation functions themselves \citep{liu2024kan}.

The performance of KANs has been evaluated across various domains, including scientific machine learning tasks \citep{wang2024KANinformed,shukla2024comprehensive,abueidda2024deepokan,koenig2024kan}, image classification \citep{azam2024suitability,cheon2024kolmogorovRemote,lobanov2024hyperkan,yu2024kan,ExploreClassification}, image segmentation \citep{li2024UKAN,tang20243d}, image detection \citep{wang2024spectralkan}, audio classification \citep{yu2024kan}, and other applications. Additionally, from a neural network architecture perspective, KANs have been integrated into convolutional neural networks (CNNs) \citep{azam2024suitability,bodner2024CNNkan} and graph neural networks \citep{kiamari2024gkan,bresson2024kagnns,zhang2024graphKAN,de2024kolmogorovGraph}.

However, the efficiency of KANs for 3D point cloud data has not yet been explored. Point cloud data plays a critical role in various domains, including computer graphics, computer vision, robotics, and autonomous driving \citep{uy2018pointnetvlad,li2020deepReview,guo2020deep,zhang2023deep,zhang2023deepSurvay}. One of the most successful neural networks for deep learning on point cloud data is PointNet, introduced by \citet{qi2017pointnet}. Following this, several modified and advanced versions of PointNet have been developed \citep{qi2017pointnet++,shen2018mining,thomas2019kpconv,wang2019dynamic,zhao2021point}. To the best of our knowledge, the only existing work embedding KANs into PointNet involves 2D supervised learning in the context of computational fluid dynamics \citep{kashefi2024kolmogorov}. In this work, we integrate KANs into PointNet for the first time to evaluate its performance on classification and segmentation tasks for 3D point cloud data.

It is important to clarify that by embedding KANs into PointNet, we do not simply mean replacing every instance of MLPs with KANs. While such an approach could be considered a research case, our goal is to preserve and utilize the core principles upon which PointNet is built. First, we apply shared KANs, meaning that the same KANs are applied to all input points. Second, we utilize a symmetric function, such as the max function, to extract global features from the points. These two elements are fundamental to PointNet, and by maintaining them, we ensure that the network remains invariant to input permutations. Our objective is to propose a version of PointNet integrated with KANs that retains these two essential properties, which we refer to as PointNet-KAN throughout the rest of this article. First, we focus on PointNet \citep{qi2017pointnet} to directly and explicitly investigate the effect of KANs on the network's performance. Using more complex versions of PointNet \citep{qi2017pointnet++,shen2018mining,thomas2019kpconv,wang2019dynamic,zhao2021point} could introduce other factors that might obscure the direct influence of KANs, making it challenging to determine whether any performance changes are due to the KAN architecture or other components of the network. Next, we investigate a hybrid version of PointNet that incorporates KAN layers in the encoder and MLP layers in the decoder or classification head, motivated by the desire to combine the advantages of KAN and MLP components. Moreover, we explore embedding KAN layers into PointNet++ \citep{qi2017pointnet++}, an advanced point cloud-based neural network, for 3D object classification.

We use Jacobi polynomials to construct PointNet-KAN and investigate its performance across different polynomial degrees. Additionally, we examine the effect of special cases of Jacobi polynomials, including Legendre polynomials, Chebyshev polynomials of the first and second kinds, and Gegenbauer polynomials. The performance of PointNet-KAN is evaluated across classification, part segmentation, and segmentation tasks. Overall, the summary of our key contributions is as follows:

\begin{itemize}
\item We introduce PointNet with KANs (i.e., PointNet-KAN) for the first time and evaluate its performance against PointNet with MLPs.
\item We embed KAN into a point-cloud-based neural network for the first time, for computer vision tasks on unordered 3D point sets.
\item We conduct an extensive evaluation of the hyperparameters of PointNet-KAN, specifically the degree and type of polynomial used in constructing KANs.
\item We assess the efficiency of PointNet-KAN on benchmarks for 3D object classification and segmentation tasks.
\item We demonstrate that PointNet-KAN achieves competitive performance to PointNet, despite having a much shallower and simpler network architecture.
\item We additionally investigate a hybrid PointNet architecture that integrates both KAN and MLP components.
\item We release our code to support reproducibility and future research. 
\end{itemize}


\section{Related work}
\label{Sect5}

Relevant work on KANs can be discussed from two perspectives. The first focuses on using KANs for classification and segmentation tasks in computer graphics and computer vision. For classification, researchers \citep{cheon2024kolmogorovRemote,bodner2024CNNkan,azam2024suitability} have embedded KANs as a replacement for MLPs in various popular CNN-based neural networks for two-dimensional image classification, such as VGG16 \citep{simonyan2014very}, MobileNetV2 \citep{sandler2018mobilenetv2}, EfficientNet \citep{tan2019efficientnet}, ConvNeXt \citep{liu2022convnet}, ResNet-101 \citep{he2016deepRESNET}, and Vision Transformer \citep{dosovitskiy2020image}, and evaluated the performance of these networks with KANs. For 3D image segmentation tasks, KANs have been embedded into U-Net \citep{ronneberger2015uUNET} as a replacement for MLPs \citep{tang20243d,wu2024transukan}. However, no prior work has explored the use of KANs in point-cloud-based neural networks for 3D classification and segmentation of unordered point sets or evaluated their performance on complex benchmark datasets such as ModelNet40 \citep{wu20153d} and the ShapeNet Part dataset \citep{yi2016scalable}. From the second perspective, KANs were originally constructed using B-spline as the basis polynomial \citep{liu2024kan}, and researchers employed this type of polynomial for image classification and segmentation \citep{cheon2024kolmogorovRemote,bodner2024CNNkan,azam2024suitability}. However, studies have shown that B-splines are computationally expensive and pose difficulties in implementation \citep{howard2024finite,rigas2024adaptive}. To address these issues, recent advancements in scientific machine learning suggested the use of Jacobi polynomials as an alternative in KANs \citep{ss2024chebyshev,seydi2024exploring}. Accordingly, Jacobi polynomials are not only easier to implement but also computationally more efficient. However, no prior work has explored the use of KANs with Jacobi polynomials in computer vision for classification and segmentation tasks.

A key motivation for introducing KANs into point cloud-based networks is to reduce the complexity of current architectures, particularly in terms of the number of layers and overall network depth. For example, as shown in Fig. 2 of the PointNet article \citep{qi2017pointnet}, the classification branch consists of five shared MLP layers in the encoder and three MLP layers in the classification head. Moreover, the encoder includes two mini-PointNets, one used for input transformation and another for feature transformation, adding further complexity. Similarly, as illustrated in Fig. 9 of the PointNet paper \citep{qi2017pointnet}, the part segmentation network employs eight shared MLP layers in the encoder, while the decoder’s first layer concatenates the global feature vector with five intermediate features extracted from the encoder. Compared to an MLP layer, a KAN layer appears to be a more capable component. Its learnable activation function is governed by a polynomial, with parameters that can switch between different polynomial types, offering flexibility based on the dataset and task. Thus, we investigate the potential of using KANs to enable shallower networks with comparable or better performance.

In addition to the area of computer vision and computer graphics, \cite{kashefi2024kolmogorov} introduced KAPointNet to predict the velocity and pressure fields in fluid dynamics on irregular geometries. To construct KA‑PointNet, \cite{kashefi2024kolmogorov} modified the segmentation branch of PointNet by replacing shared MLP layers with shared KAN layers and used mean squared error as the loss function. However, KA‑PointNet \citep{kashefi2024kolmogorov} was tested only on two‑dimensional spatial domains, and the possibility of obtaining a lighter‑weight architecture through layer pruning was not explored. In contrast, the present work applies shared KAN layers to both the classification and segmentation branches of PointNet, operates on three‑dimensional coordinates, and adopts a cross‑entropy loss. We analyze the effect of network depth to assess whether sequential stacking of KAN layers is required, and we integrate KAN layers into the hierarchical PointNet++ framework \citep{qi2017pointnet++}, a more advanced point‑based neural architecture, for both classification and segmentation tasks.


\section{Kolmogorov-Arnold Network (KAN) layers}
\label{Sect2}

Inspired by the Kolmogorov-Arnold representation theorem \citep{kolmogorovSuperposition,arnold2009functions}, Kolmogorov-Arnold Network (KAN) has been proposed as a novel neural network architecture by \citet{liu2024kan}. According to the theorem, multivariate continuous function can be expressed as a finite composition of continuous univariate functions and additions. To describe the structure of KAN straightforwardly, consider a single-layer KAN. The network’s input is a vector $\vr$ of size $d_\text{input}$, and its output is a vector $\vs$ of size $d_\text{output}$. In this configuration, the single-layer KAN maps the input to the output as follows:

\begin{equation}
\vs_{d_\text{output}} = \tA_{d_\text{output}\times d_\text{input}} \vr_{d_\text{input}},
\label{Eq1}
\end{equation}

where the tensor $\tA_{d_\text{output}\times d_\text{input}}$ is expressed as:

\begin{equation}
\tA_{d_\text{output}\times d_\text{input}} = 
\left[
\begin{array}{cccc}
\psi_{1,1}(\cdot) & \psi_{1,2}(\cdot) & \cdots & \psi_{1,d_\text{input}}(\cdot) \\
\psi_{2,1}(\cdot) & \psi_{2,2}(\cdot) & \cdots & \psi_{2,d_\text{input}}(\cdot) \\
\vdots & \vdots & \ddots & \vdots \\
\psi_{d_\text{output},1}(\cdot) & \psi_{d_\text{output},2}(\cdot) & \cdots & \psi_{d_\text{output},d_\text{input}}(\cdot) \\
\end{array}
\right],
\label{Eq2}
\end{equation}

where each $\psi(\gamma)$ (the subscript is removed to lighten notation) is defined as:

\begin{equation}
\psi(\gamma) = \sum_{i=0}^n \omega_i  f_i^{(\alpha,\beta)}(\gamma),
\label{Eq3}
\end{equation}

where $f_i^{(\alpha,\beta)}(\gamma)$ represents the Jacobi polynomial of degree $i$, $n$ is the polynomial order of $\psi$, and $\omega_i$ are trainable parameters. Hence, the total number of trainable parameters embedded in $\tA$ is $(n+1)\times d_\text{input} \times d_\text{output}$. We implement $f_n^{(\alpha,\beta)}(\gamma)$ using a recursive relation \citep{Szego1939Orthogonal}:

\begin{equation}
f_n^{(\alpha,\beta)}(\gamma) = (a_n \gamma + b_n)f_{n-1}^{(\alpha,\beta)}(\gamma) + c_n f_{n-2}^{(\alpha,\beta)}(\gamma),
\label{Eq4}
\end{equation}

where the coefficients $a_n$, $b_n$, and $c_n$ are given by:

\begin{equation}
a_n = \frac{(2n+\alpha+\beta-1)(2n+\alpha+\beta)}{2n(n+\alpha+\beta)},
\label{Eq5}
\end{equation}

\begin{equation}
b_n = \frac{(2n+\alpha+\beta-1)(\alpha^2 - \beta^2)}{2n(n+\alpha+\beta)(2n+\alpha+\beta-2)},
\label{Eq6}
\end{equation}

\begin{equation}
c_n = \frac{-2(n+\alpha-1)(n+\beta-1)(2n+\alpha+\beta)}{2n(n+\alpha+\beta)(2n+\alpha+\beta-2)},
\label{Eq7}
\end{equation}

with the following initial conditions:

\begin{equation}
f_0^{(\alpha,\beta)}(\gamma) = 1,
\label{Eq8}
\end{equation}

\begin{equation}
f_1^{(\alpha,\beta)}(\gamma) = \frac{1}{2}(\alpha+\beta+2)\gamma + \frac{1}{2}(\alpha-\beta).
\label{Eq9}
\end{equation}

Since $f_n^{(\alpha,\beta)}(\gamma)$ is recursively constructed, the polynomials $f_i^{(\alpha,\beta)}(\gamma)$ for $0 \leq i \leq n$ are computed sequentially. Additionally, because the input to the Jacobi polynomials must lie within the interval $[-1, 1]$, the input vector $\vr$ needs to be scaled to fit this range before being passed to the KAN layer. To achieve this, we apply the hyperbolic tangent function. Finally, setting $\alpha = \beta = 0$ yields the Legendre polynomial \citep{MiltonHandbook,Szego1939Orthogonal}, while the Chebyshev polynomials of the first and second kinds are obtained with $\alpha = \beta = -0.5$ and $\alpha = \beta = 0.5$, respectively \citep{MiltonHandbook,Szego1939Orthogonal}. Additionally, the Gegenbauer (or ultraspherical) polynomials arise when $\alpha = \beta$ \citep{Szego1939Orthogonal}.

\section{Overview of PointNet and its key principles}
\label{SectReview}

Consider a point cloud $\mathcal{X}$ as an unordered set with $N$ points, defined as $\mathcal{X} = \left\{ \mathbf{x}_j \in \mathbb{R}^d \right\}_{j=1}^N$. The dimension (or number of features) of each $\mathbf{x}_j$ is shown by $d$. According to the Theorem 1 proposed in \cite{qi2017pointnet}, a set function $g : \mathcal{X} \to \mathbb{R}$ can be defined to map this set of points to a vector as follows:

\begin{equation}
   g(\mathbf{x}_1, \mathbf{x}_2, \dots, \mathbf{x}_N) = \tau \left( \max_{j=1, \dots, N} \{ h(\mathbf{x}_j) \} \right),
\label{Eq10}
\end{equation}
where $\max$ is a vector-wise max operator that takes $N$ vectors as input and returns a new vector, computed as the element-wise maximum. In PointNet \citep{qi2017pointnet}, the continuous functions $\tau$ and $h$ are implemented as MLPs. In this work, we replace $\tau$ and $h$ with KANs, resulting in PointNet-KAN. Note that the function $g$ is invariant to the permutation of input points. Details of this theorem and its proof can be found in \cite{qi2017pointnet}.

\section{Architecture of PointNet-KAN}
\label{Sect3}


\paragraph{Classification branch}

\begin{figure}[h]
\begin{center}
\includegraphics[width=13cm]{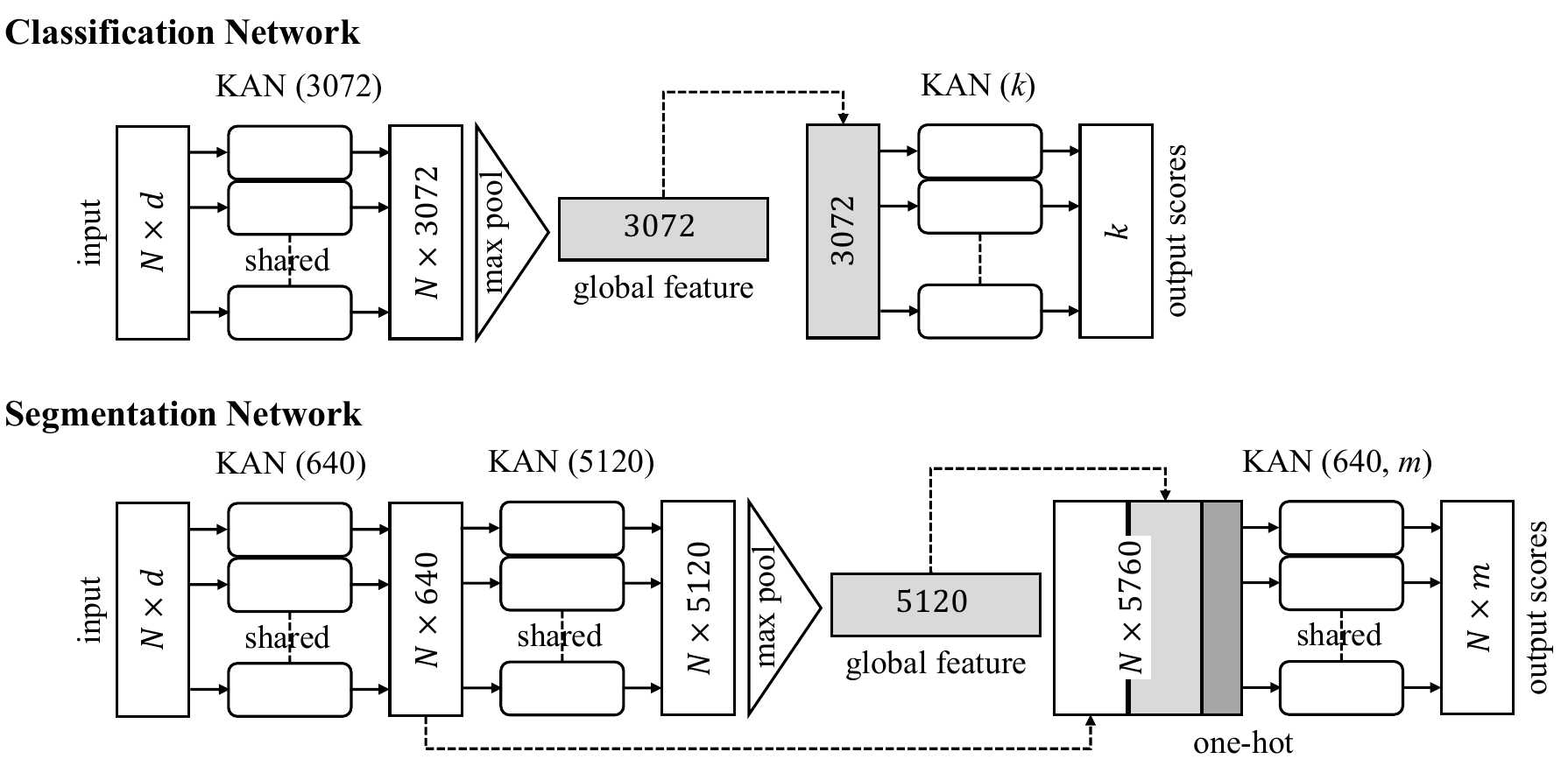}
\end{center}
\caption{Architecture of PointNet-KAN. The classification network is shown in the top panel, and the segmentation network is shown in the bottom panel. $N$ is the number of points in a point cloud. $d$ indicates the number of input point features (e.g., spatial coordinates, normal vectors, etc.). $k$ indicates the number of classes (e.g., for the ModelNet40 \citep{wu20153d} benchmark, $k=40$; see Sect. \ref{Sect41}). $m$ indicates the total number of possible parts (e.g., for the ShapeNet part \citep{yi2016scalable} benchmark, $m=50$; see Sect. \ref{Sect42}). In the semantic‐segmentation application, the one‐hot vector is removed and \(m\) denotes the total number of possible classes (e.g., in the Stanford 3D semantic parsing benchmark \citep{armeni20163d}, \(m=13\)).}
\label{Fig1}
\end{figure}

The top panel of Fig. \ref{Fig1} demonstrates the classification branch of PointNet-KAN. The architecture of the classification branch is explained as follows. The PointNet-KAN model accepts input with dimensionality corresponding to 3D spatial coordinates (i.e., $d=3$) and possibly the 3D normal vector as part of the point set representation (i.e., $d=6$). A shared KAN layer maps the input feature vector from its original space to an intermediate feature space of dimension 3072. Following the first shared KAN layer, batch normalization \citep{ioffe2015batch} is applied. After normalization, a max pooling operation is performed to extract global features by computing the maximum value across all points in the point cloud. Next, the global feature is passed through a KAN layer, which reduces the dimensionality to the number of output channels (i.e., $k$), corresponding to the classification task. A softmax function is applied to the output to convert the logits into class probabilities. The concept of shared KANs is analogous to the shared MLPs used in PointNet \citep{qi2017pointnet}. It means that the same functional tensor, $\tA$, is applied uniformly to the input or intermediate features in PointNet-KAN. The use of the shared KAN layers and the symmetric max-pooling function ensure that PointNet-KAN is invariant to the order of the points in the point cloud.

\paragraph{Part segmentation and semantic segmentation branch}

As shown in the bottom panel of Fig. \ref{Fig1}, the part segmentation branch of the PointNet-KAN is described as follows. The input is first passed through a shared KAN layer, transforming it to an intermediate feature space of size 640, followed by batch normalization. These local features are then processed by a second shared KAN layer, mapping them to a higher-dimensional space of size 5120, and another batch normalization step is applied. A max pooling operation extracts a global feature representing the entire point cloud, which is then expanded to match the number of points. The one-hot encoded class label, representing the object category, is concatenated with the local features and the global feature. This combined feature, consisting of local features of size 640, global features of size 5120, and the class label, is passed through a shared KAN layer to reduce the feature size to 640, followed by batch normalization. A final shared KAN layer generates the output, delivering point-wise segmentation predictions, followed by a softmax function to convert the logits into class probabilities. The architecture of PointNet-KAN for the semantic segmentation task is similar to that used for part segmentation (as shown in Fig. \ref{Fig1}), except that it does not include a one-hot vector.

\begin{table}
\caption{Classification results on ModelNet40 \citep{wu20153d}. In PointNet-KAN, the Jacobi polynomial degree is set to 4 (i.e., $n=4$) with $\alpha = \beta = 1.0$. In PointNet-KAN-MLP++, the Jacobi polynomial degree is set to 2 (i.e., $n=2$) with $\alpha = \beta = -0.5$. Time complexity for PointNet-KAN, PointNet-KAN-MLP++, PointNet++, DGCNN, PointMLP, and PointNet is provided. `M' and `G' stand for million and billion, respectively.}
\label{Table1}
\vspace{1mm}
\begin{adjustbox}{width=1\textwidth}
\begin{tabular}{c|ccccc}
\toprule
 & normal vector & number of points & Mean class accuracy & Overall accuracy & FLOPs/sample \\
\midrule
PointNet++ \citep{qi2017pointnet++}    & no & 2048   &  -   &  90.7 & 1.7G \\
PointNet++ \citep{qi2017pointnet++}    & yes & 2048   &  -   &  91.9 & 3.2G\\
DGCNN \citep{wang2019dynamic} & no & 2048   &  90.7   &  93.5 & 4.8G\\
Point Transformer  \citep{zhao2021point}   & yes & -   &  90.6   &  93.7 & - \\
PointMLP \citep{ma2022rethinking} & no & 1000   &  91.4   &  94.5 & 31.3G\\
ShapeLLM  \citep{qi2024shapellm} & no & 1000   &  94.8   &  95.0  & -\\
\midrule
PointNet (baseline) \citep{qi2017pointnet} & no   & 1024   & 72.6 & 77.4 &  148M \\
PointNet \citep{qi2017pointnet}   & no   &  1024   &  86.2   &  89.2  & 440M \\
\midrule
PointNet-KAN    &  no &  1024   &  82.7   &  87.5  & 60M\\
PointNet-KAN    & yes   &  1024   &  87.2   &  90.5 & 110M \\
\midrule
PointNet-KAN-MLP++   &  yes &  2048   &  90.6   &  92.9  & 2.7G \\
\bottomrule
\end{tabular}
\end{adjustbox}
\end{table}


\section{Experiment and discussion}
\label{Sect4}

\subsection{3D object classification}
\label{Sect41}

We evaluate PointNet-KAN on the ModelNet40 \citep{wu20153d} shape classification benchmark, which contains 12,311 models across 40 categories, with 9,843 models allocated for training and 2,468 for testing. Similar to \cite{qi2017pointnet}, we uniformly sample 1,024 points from the mesh faces and normalize them into a unit sphere. We also conduct an experiment with included normal vectors as input features, computed using the trimesh library \citep{trimesh}. Table \ref{Table1} presents the classification results of PointNet-KAN, with a polynomial degree of 4 (i.e., $n=4$ in Eq. \ref{Eq3}) and $\alpha = \beta = 1$. Training details are provided in \ref{ATD}. The obtained results can be interpreted from two perspectives.

First, comparing PointNet-KAN with PointNet (baseline) \citep{qi2017pointnet} and PointNet \citep{qi2017pointnet} shows that PointNet-KAN (with or without normal vectors) achieves higher accuracy than PointNet (baseline). Additionally, PointNet-KAN with normal vectors as input features outperforms PointNet. The number of trainable parameters for PointNet-KAN with $n=4$, PointNet (baseline), and PointNet in the classification branch is approximately 1M, 0.8M, and 3.5M, respectively. It is worth noting that PointNet-KAN with $n=2$ has only roughly 0.6M trainable parameters, making it lighter than PointNet (baseline) while still achieving an overall accuracy of 89.9 (see Table \ref{Table2}). Notably, despite its simpler architecture—lacking the input and feature transforms found in PointNet, as shown in Fig. 2 of \cite{qi2017pointnet}, and having only 3 hidden layers compared to the 8 hidden layers of PointNet—PointNet-KAN still delivers competitive results, with overall accuracy of 90.5\% versus 89.2\%. From a time complexity perspective, the number of floating-point operations required for one forward pass of the PointNet-KAN model is significantly lower than that of PointNet, as shown in Table \ref{Table1}.

From the second perspective, we observe that other advanced point-cloud-based deep learning frameworks, such as PointNet++ \citep{qi2017pointnet++}, DGCNN \citep{wang2019dynamic}, Point Transform \citep{zhao2021point}, PointMLP \citep{ma2022rethinking}, and ShapeLLM \citep{qi2024shapellm}, outperform PointNet-KAN, as listed in Table \ref{Table1}, though these models employ more advanced and complex architectures involving MLPs. This raises the question of whether redesigning these networks using KANs instead of MLPs could improve their accuracy. While the current article focuses on evaluating KAN within the simplest point-cloud-based neural network, PointNet, we hope that the promising results of PointNet-KAN motivate future efforts to embed KANs into more advanced architectures. As an example of progress in this direction, we investigate PointNet++ \citep{qi2017pointnet++} with KAN layers. The results are presented and discussed in Sect. \ref{Sect47}.


\begin{table}
\centering
\caption{Mean IoU results for part segmentation on ShapeNet part dataset \citep{yi2016scalable}. In PointNet-KAN and PointNet-KAN-MLP, the Jacobi polynomial degree is set to 2 (i.e., $n=2$) with $\alpha = \beta = -0.5$. Results of other models allocated, \cite{wu2014interactive}, 3DCNN \citep{qi2017pointnet}, \cite{yi2016scalable}, PointNet \citep{qi2017pointnet}, DGCNN \citep{wang2019dynamic}, KPConv \citep{thomas2019kpconv}, TAP \citep{wang2023take}. PN-KAN and PN-KAN-MLP stand for PointNet-KAN and PointNet-KAN-MLP in this table, respectively. No normal vectors are included in the input features of any PointNet version.}
\label{Table4}
\vspace{1mm}
\begin{adjustbox}{width=1\textwidth}
\begin{tabular}{l|c|cccccccccccccccc}
\toprule
& Mean & aero & bag & cap & car & chair & ear & guitar & knife & lamp & laptop & motor & mug & pistol & rocket & skate & table \\
& IoU &  &  & &  &  & phone &  &  &  &  &  & &  &  & board & \\
\midrule
\# shapes & & 2690 & 76 & 55 & 898 & 3758 & 69 & 787 & 392 & 1547 & 451 & 202 & 184 & 283 & 66 & 152 & 5271  \\
\midrule
Wu et al. & - & 63.2 & - & - & - & 73.5 & - & - & - & 74.4 & - & - & - & - & - & 74.8 \\
3DCNN & 79.4 & 75.1 & 72.8 & 73.3 & 70.0 & 87.2 & 63.5 & 88.4 & 79.6 & 74.4 & 93.9 & 58.7 & 91.8 & 76.4 & 51.2 & 65.3 & 77.1 \\
Yi et al. & 81.4 & 81.0 & 78.4 & 77.7 & 75.7 & 87.6 & 61.9 & 92.0 & 85.4 & 82.5 & 95.7 & 70.6 & 91.9 & 85.9 & 53.1 & 69.8 & 75.3 \\
DGCNN & 85.2 & 84.0 & 83.4 & 86.7 & 77.8 & 90.6 & 74.7 & 91.2 & 87.5 & 82.8 & 95.7 & 66.3 & 94.9 & 81.1 & 63.5 & 74.5 & 82.6 \\
KPConv & 86.4 & 84.6 & 86.3 & 87.2 & 81.1 & 91.1 & 77.8 & 92.6 & 88.4 & 82.7 & 96.2 & 78.1 & 95.8 & 85.4 & 69.0 & 82.0 & 83.6\\
TAP & 86.9 & 84.8 & 86.1 & 89.5 & 82.5 & 92.1 & 75.9 & 92.3 & 88.7 & 85.6 & 96.5 & 79.8 & 96.0 & 85.9 & 66.2 & 78.1 & 83.2 \\
\midrule
PointNet & 83.7 & 83.4 & 78.7 & 82.5 & 74.9 & 89.6 & 73.0 & 91.5 & 85.9 & 80.8 & 95.3 & 65.2 & 93.0 & 81.2 & 57.9 & 72.8 & 80.6 \\
\midrule
PN-KAN & 83.3 & 81.0 & 76.8 & 79.8 & 74.6 & 88.7 & 65.4 & 90.9 & 85.3 & 79.9 & 95.0 & 65.3 & 93.0 & 83.0 & 54.3 & 71.9 & 81.6\\
\midrule
PN-KAN-MLP & 83.9 & 82.3 & 76.3 & 86.0 & 75.8 & 89.3 & 72.7 & 91.0 & 85.3 & 80.5 & 95.3 & 65.5 & 93.4 & 82.4 & 55.7 & 74.6 & 82.0\\
\bottomrule
\end{tabular}
 \end{adjustbox}
\end{table}


\begin{figure}[]
    \centering
    \begin{subfigure}[b]{0.18\textwidth}
        \centering
        \includegraphics[width=\textwidth]{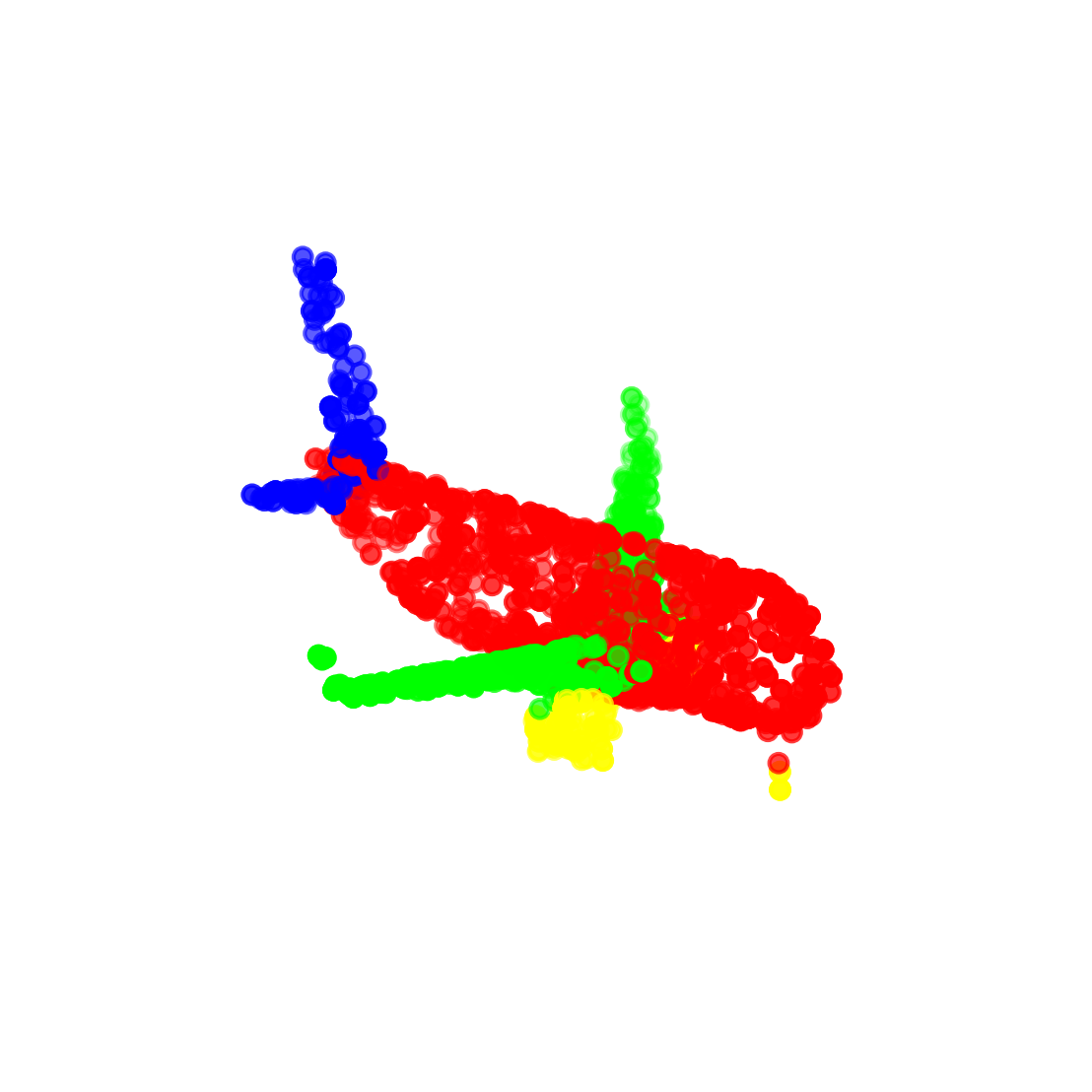}
        \caption*{Aero, gt}
    \end{subfigure}
    \hfill
    \begin{subfigure}[b]{0.18\textwidth}
        \centering
        \includegraphics[width=\textwidth]{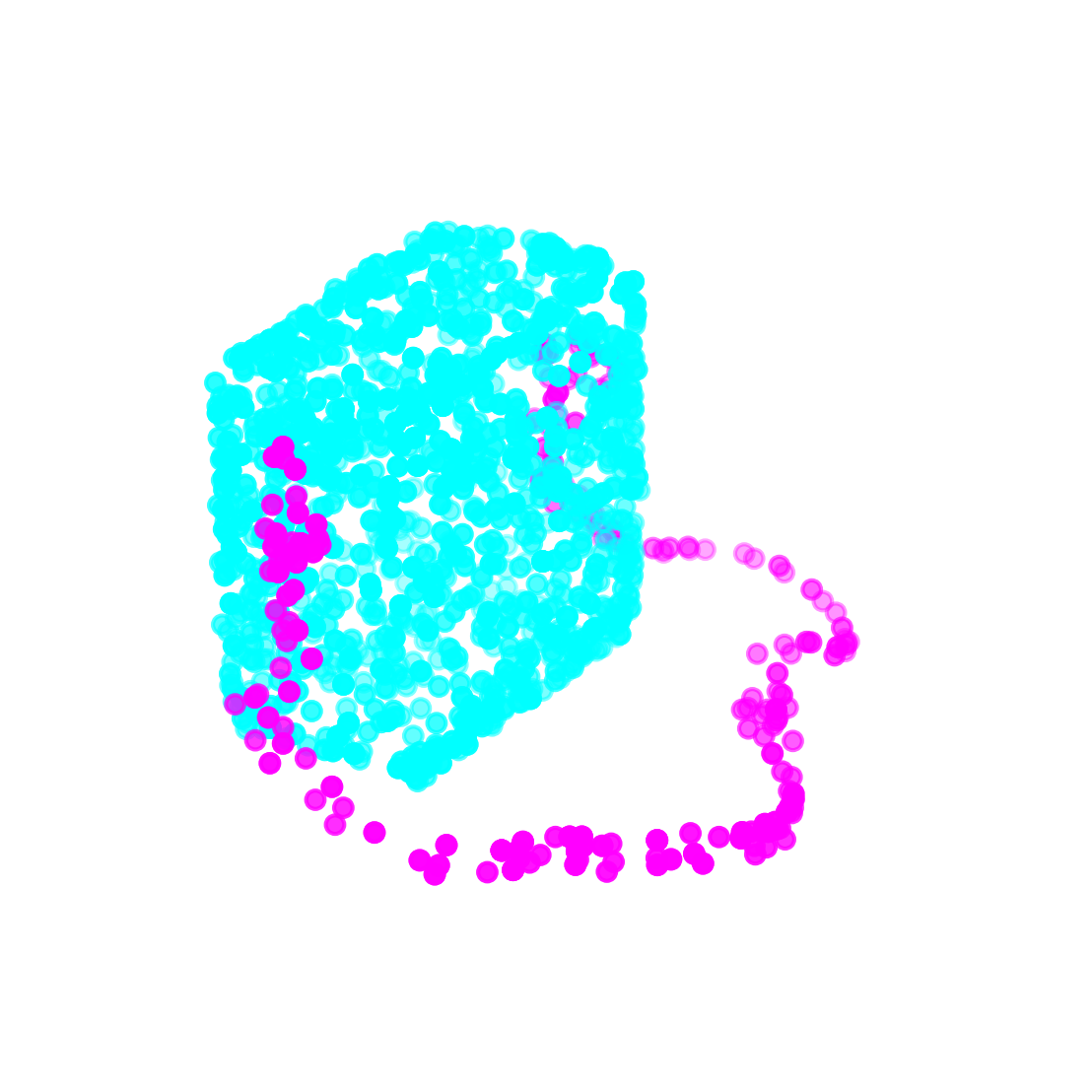}
        \caption*{Bag, gt}
    \end{subfigure}
    \hfill
    \begin{subfigure}[b]{0.18\textwidth}
        \centering
        \includegraphics[width=\textwidth]{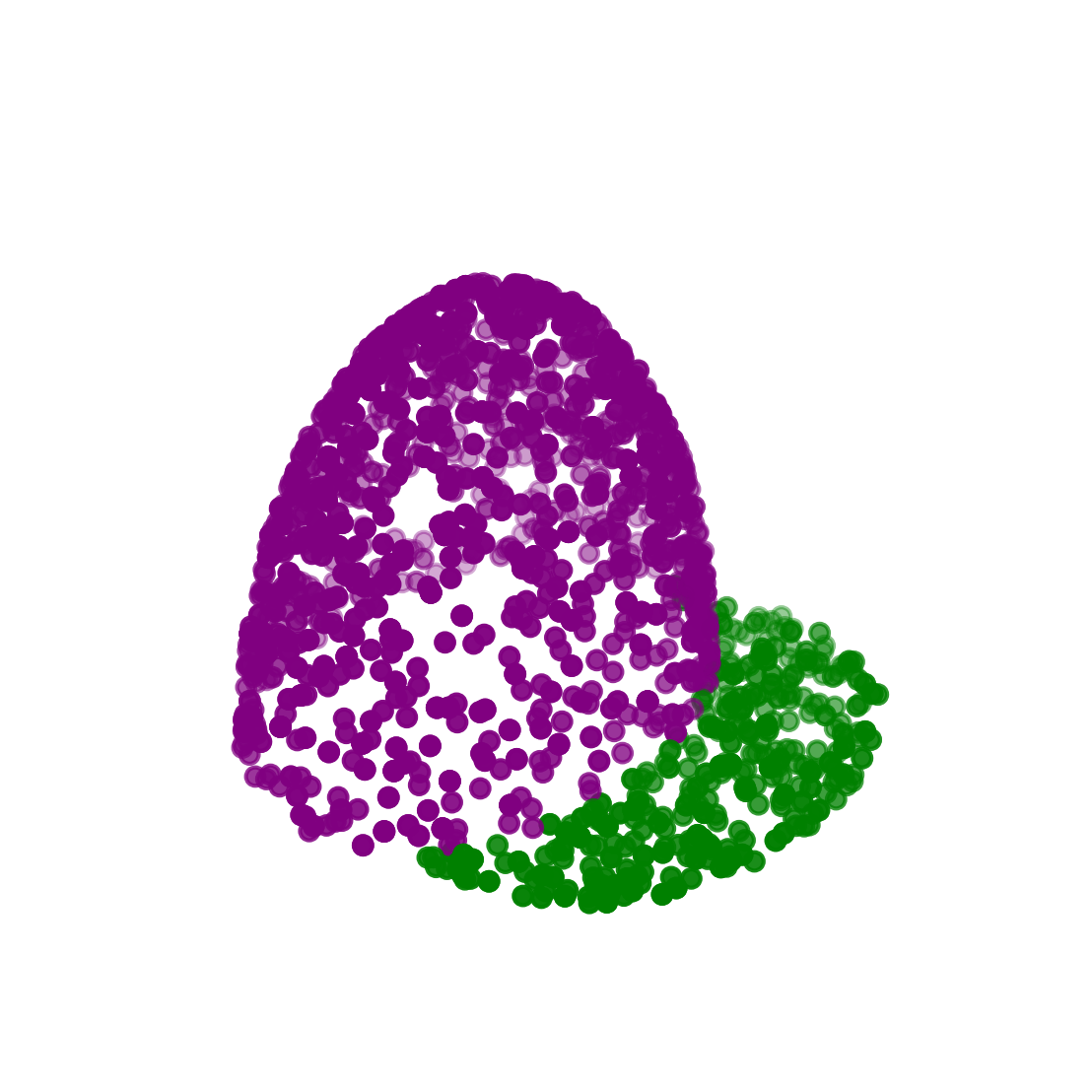}
        \caption*{Cap, gt}
    \end{subfigure}
    \hfill
    \begin{subfigure}[b]{0.18\textwidth}
        \centering
        \includegraphics[width=\textwidth]{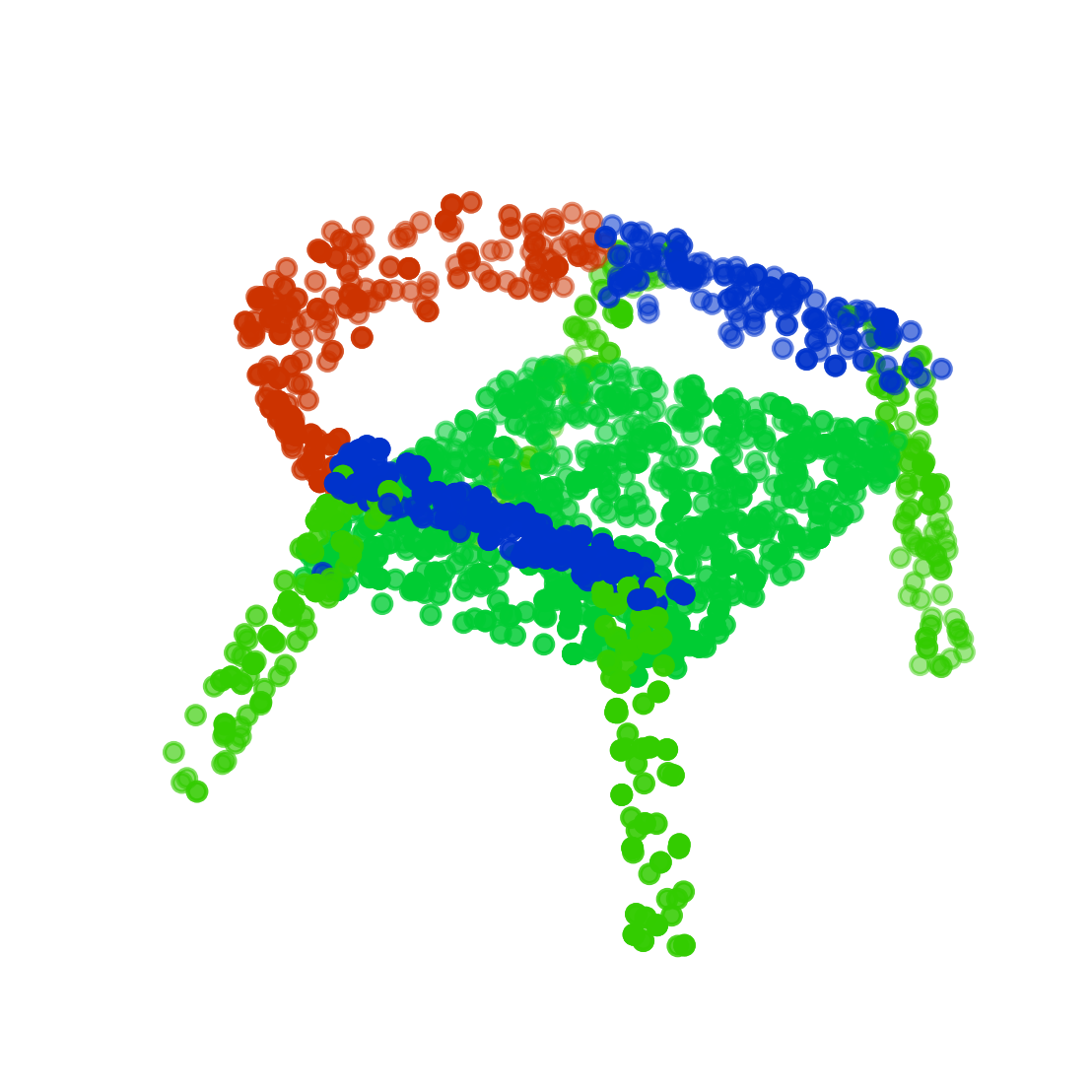}
        \caption*{Chair, gt}
    \end{subfigure}
    \hfill
    \begin{subfigure}[b]{0.18\textwidth}
        \centering
        \includegraphics[width=\textwidth]{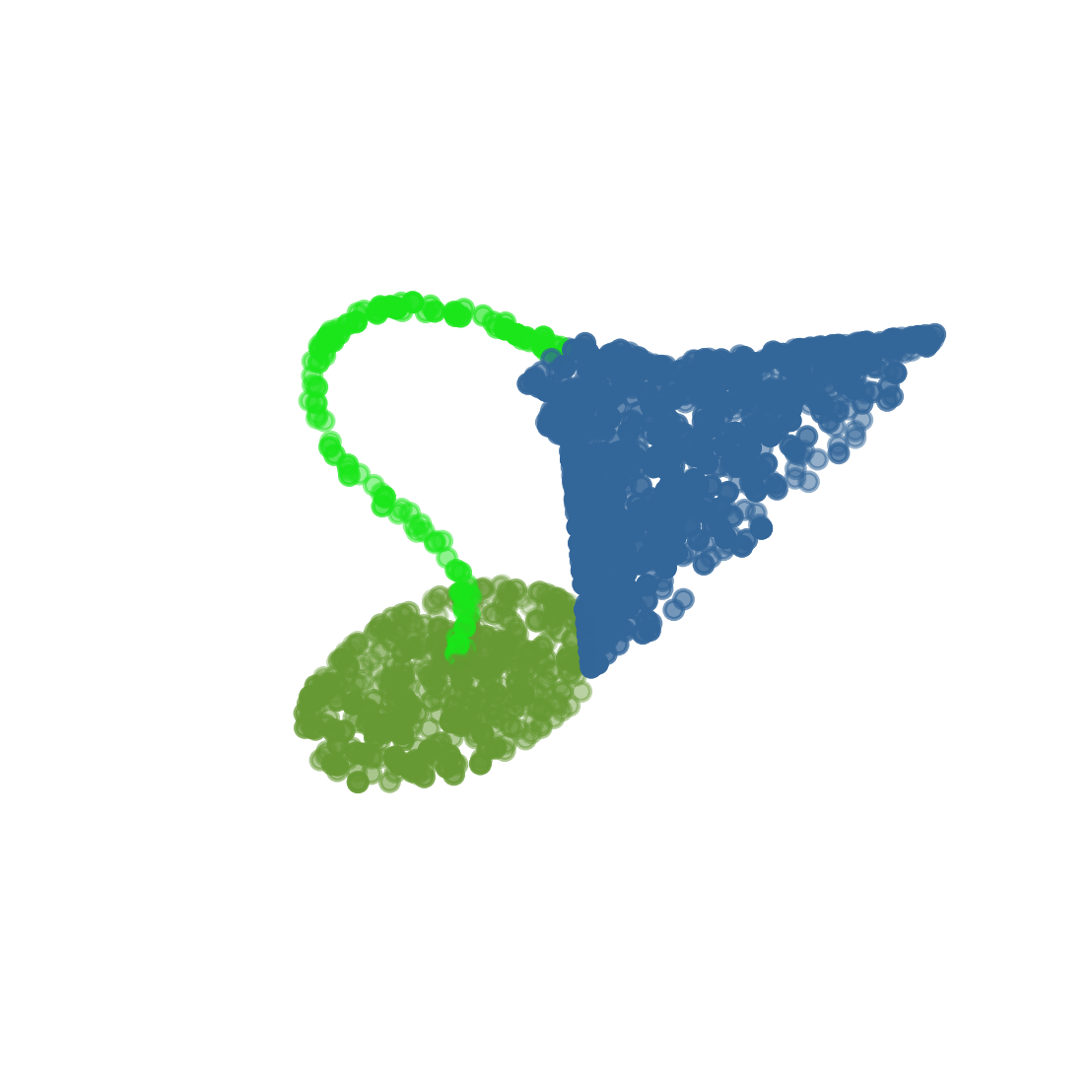}
        \caption*{Lamp, gt}
    \end{subfigure}

    \vspace{0.5cm}
    
    \begin{subfigure}[b]{0.18\textwidth}
        \centering
        \includegraphics[width=\textwidth]{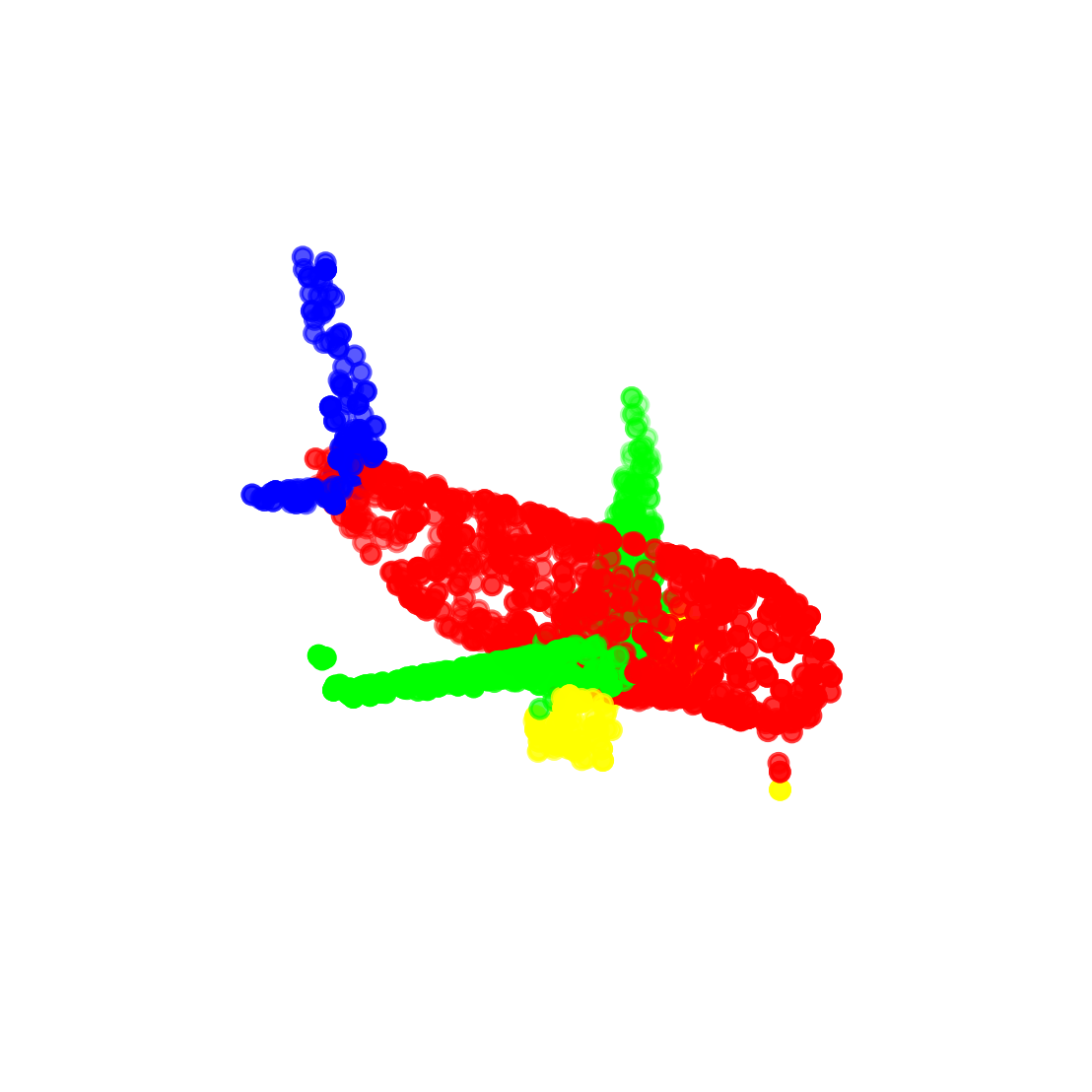}
        \caption*{Aero, p}
    \end{subfigure}
    \hfill
    \begin{subfigure}[b]{0.18\textwidth}
        \centering
        \includegraphics[width=\textwidth]{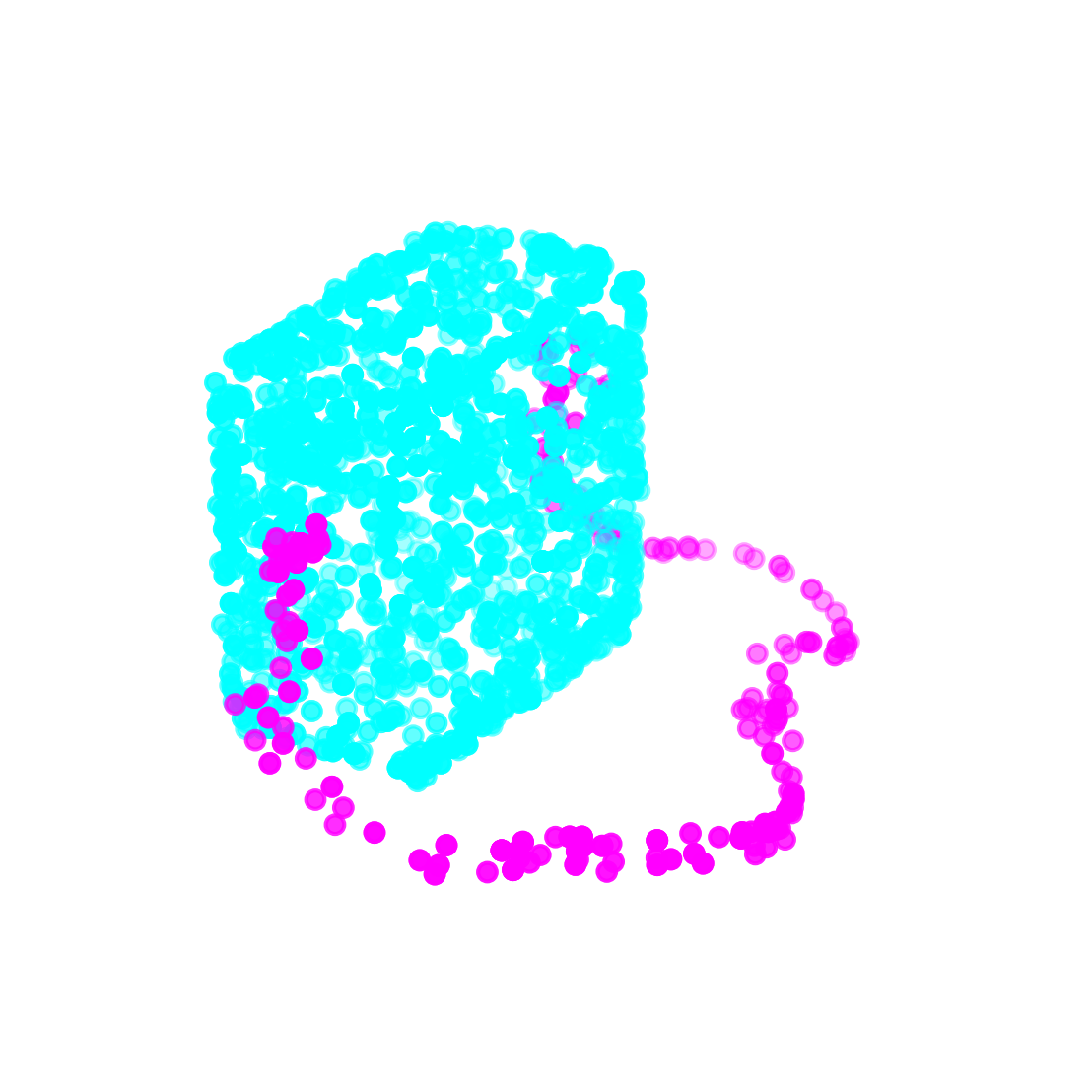}
        \caption*{Bag, p}
    \end{subfigure}
    \hfill
    \begin{subfigure}[b]{0.18\textwidth}
        \centering
        \includegraphics[width=\textwidth]{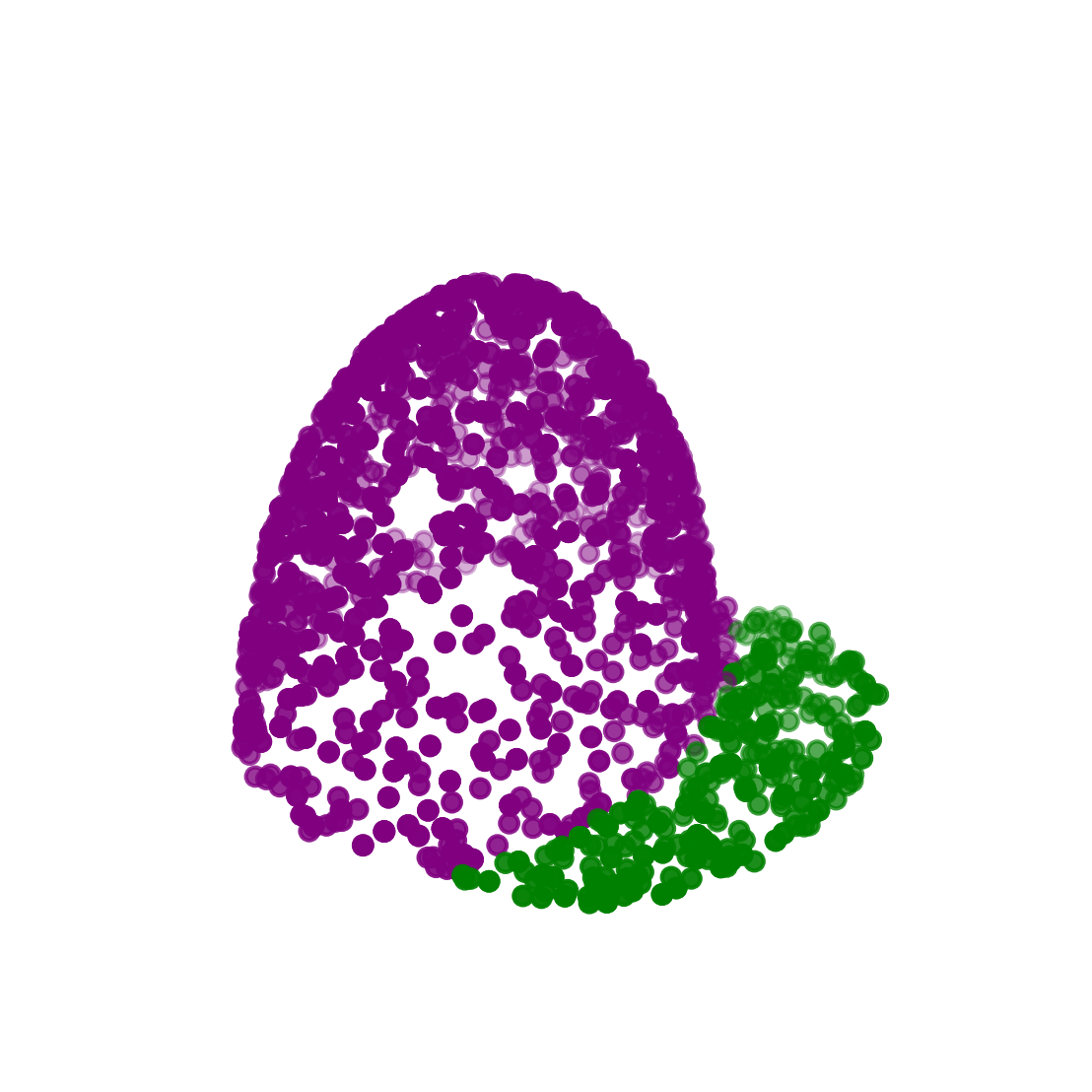}
        \caption*{Cap, p}
    \end{subfigure}
    \hfill
    \begin{subfigure}[b]{0.18\textwidth}
        \centering
        \includegraphics[width=\textwidth]{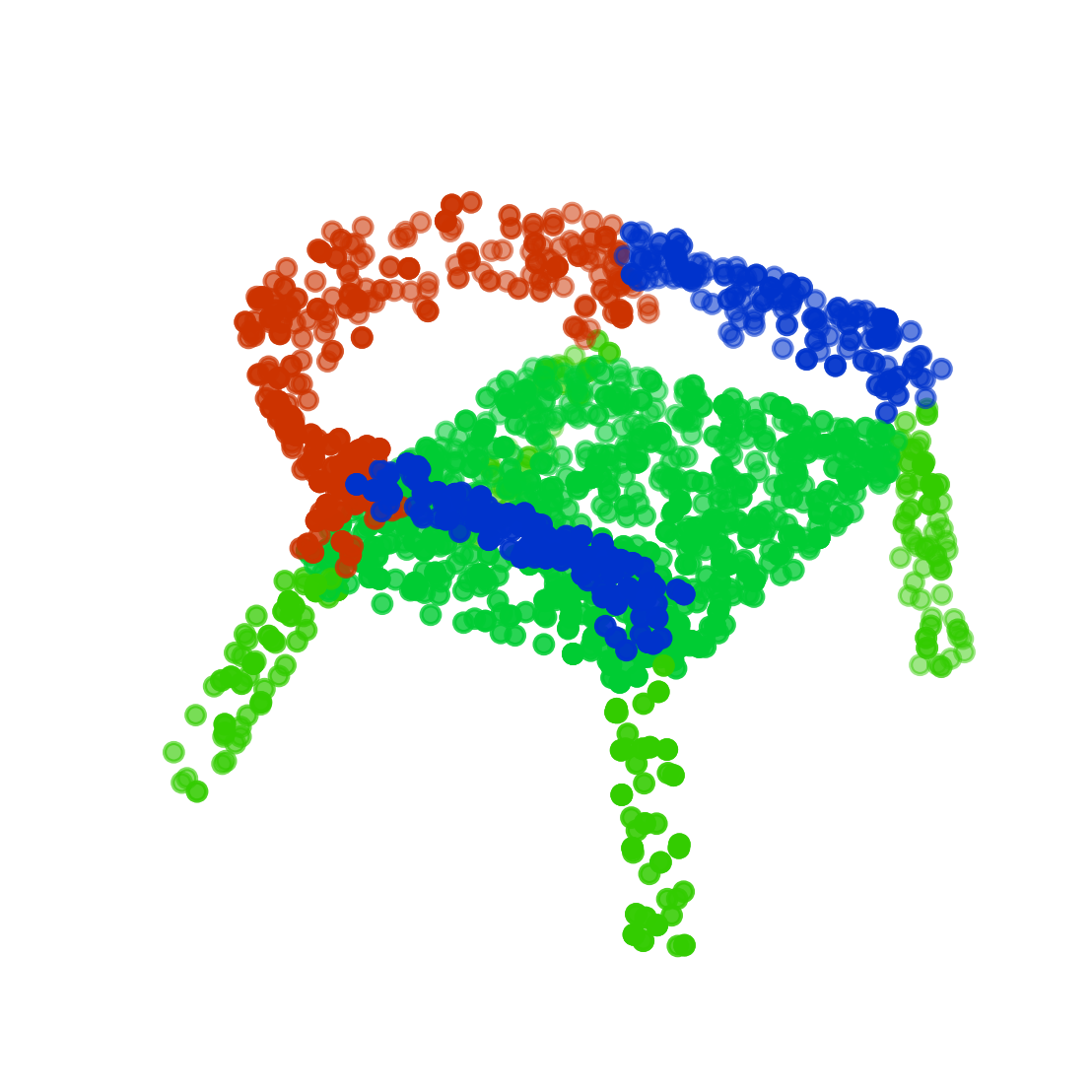}
        \caption*{Chair, p}
    \end{subfigure}
    \hfill
    \begin{subfigure}[b]{0.18\textwidth}
        \centering
        \includegraphics[width=\textwidth]{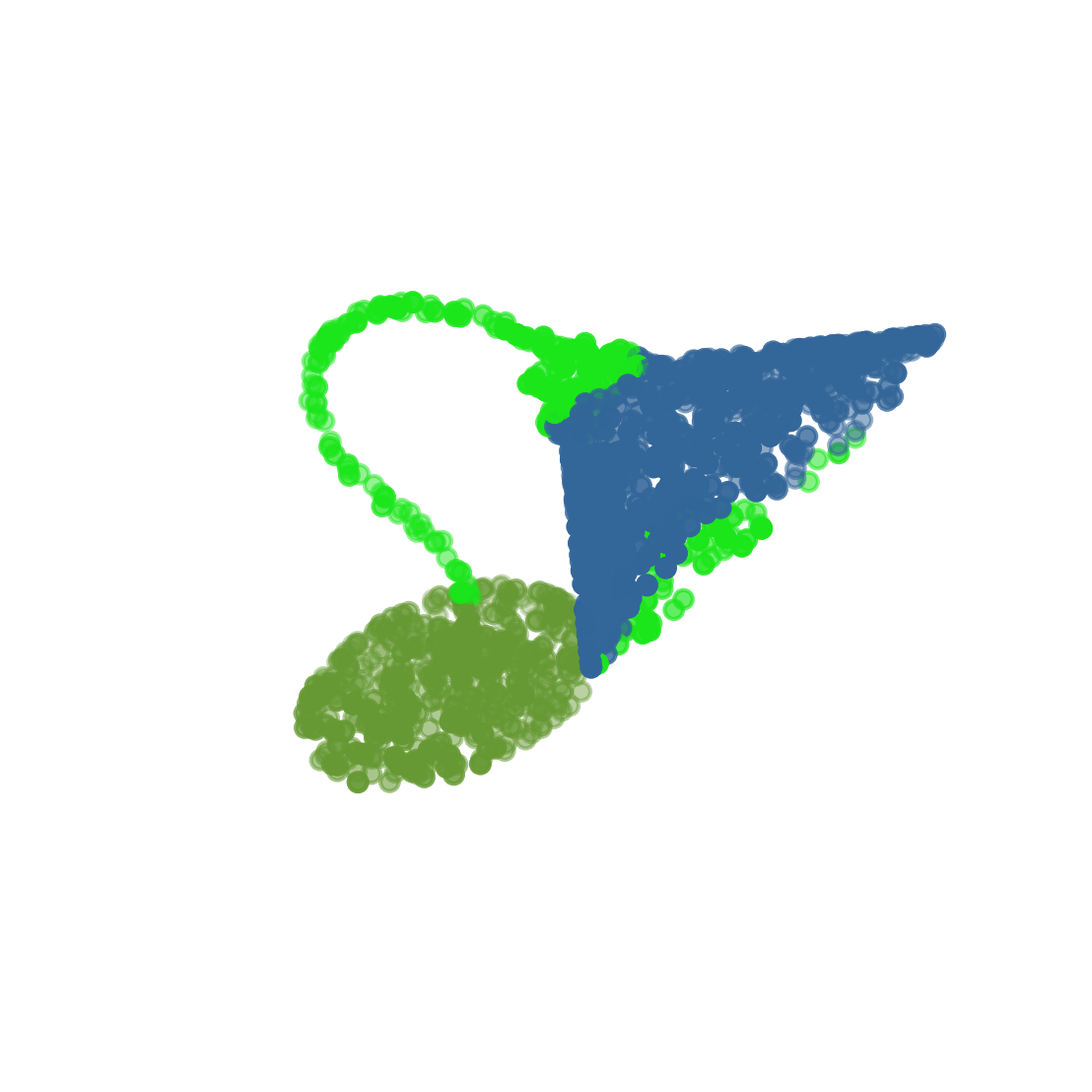}
        \caption*{Lamp, p}
    \end{subfigure}

    \vspace{0.5cm}
    
    \begin{subfigure}[b]{0.18\textwidth}
        \centering
       
        \includegraphics[width=\textwidth]{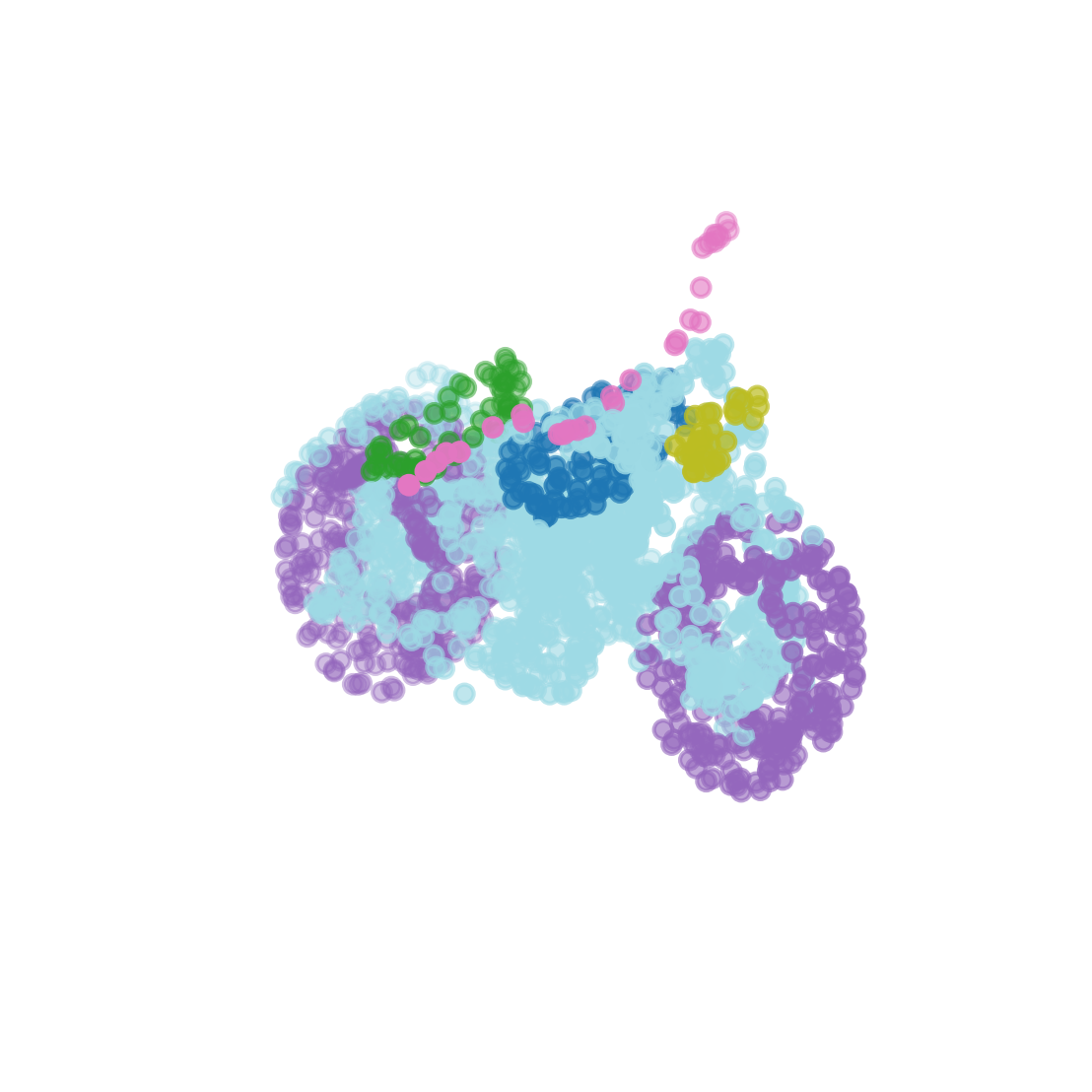}
        \caption*{Motor, gt}
    \end{subfigure}
    \hfill
    \begin{subfigure}[b]{0.18\textwidth}
        \centering
        \includegraphics[width=\textwidth]{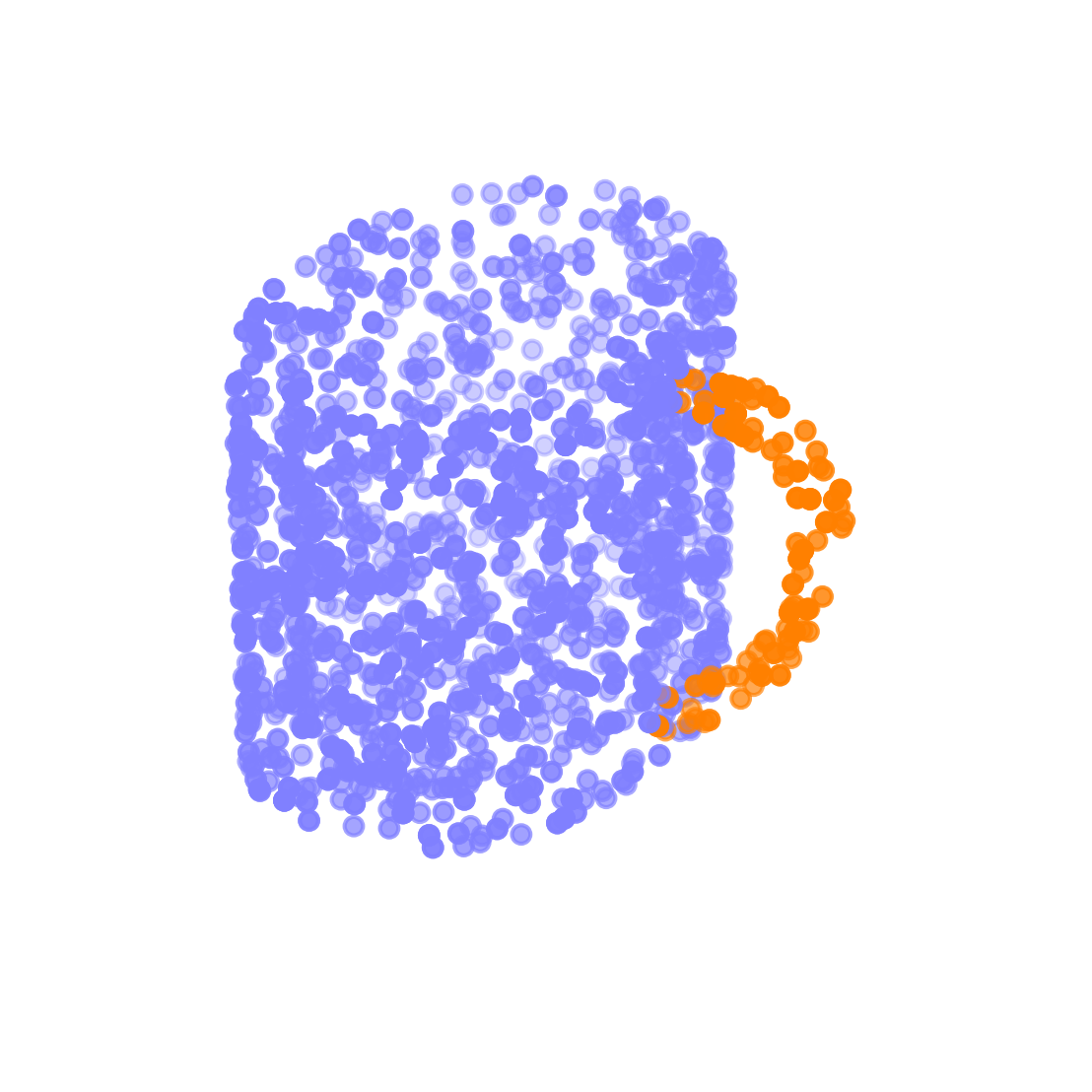}
        \caption*{Mug, gt}
    \end{subfigure}
    \hfill
    \begin{subfigure}[b]{0.18\textwidth}
        \centering
        \includegraphics[width=\textwidth]{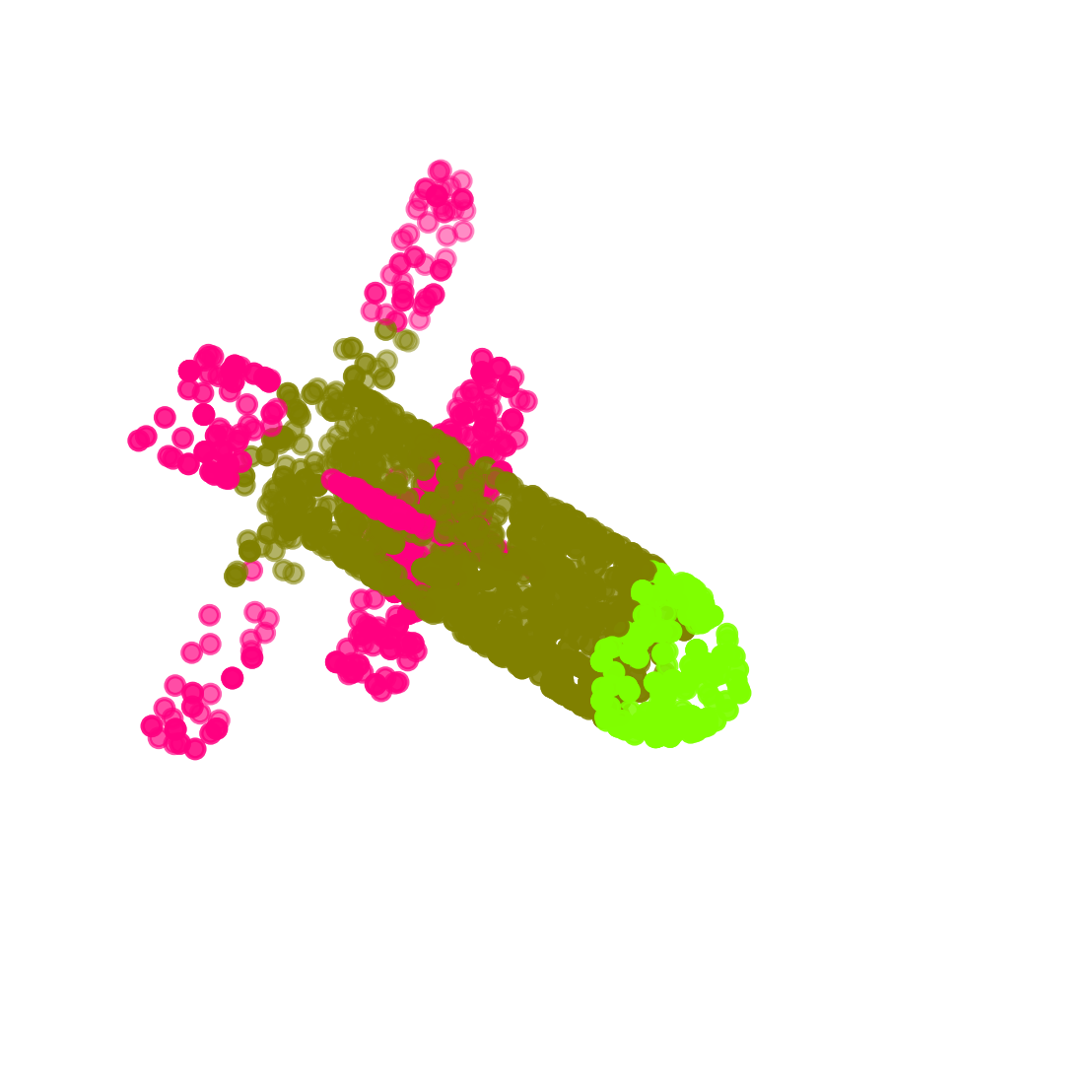}
        \caption*{Rocket, gt}
    \end{subfigure}
    \hfill
    \begin{subfigure}[b]{0.18\textwidth}
        \centering
          \includegraphics[width=\textwidth]{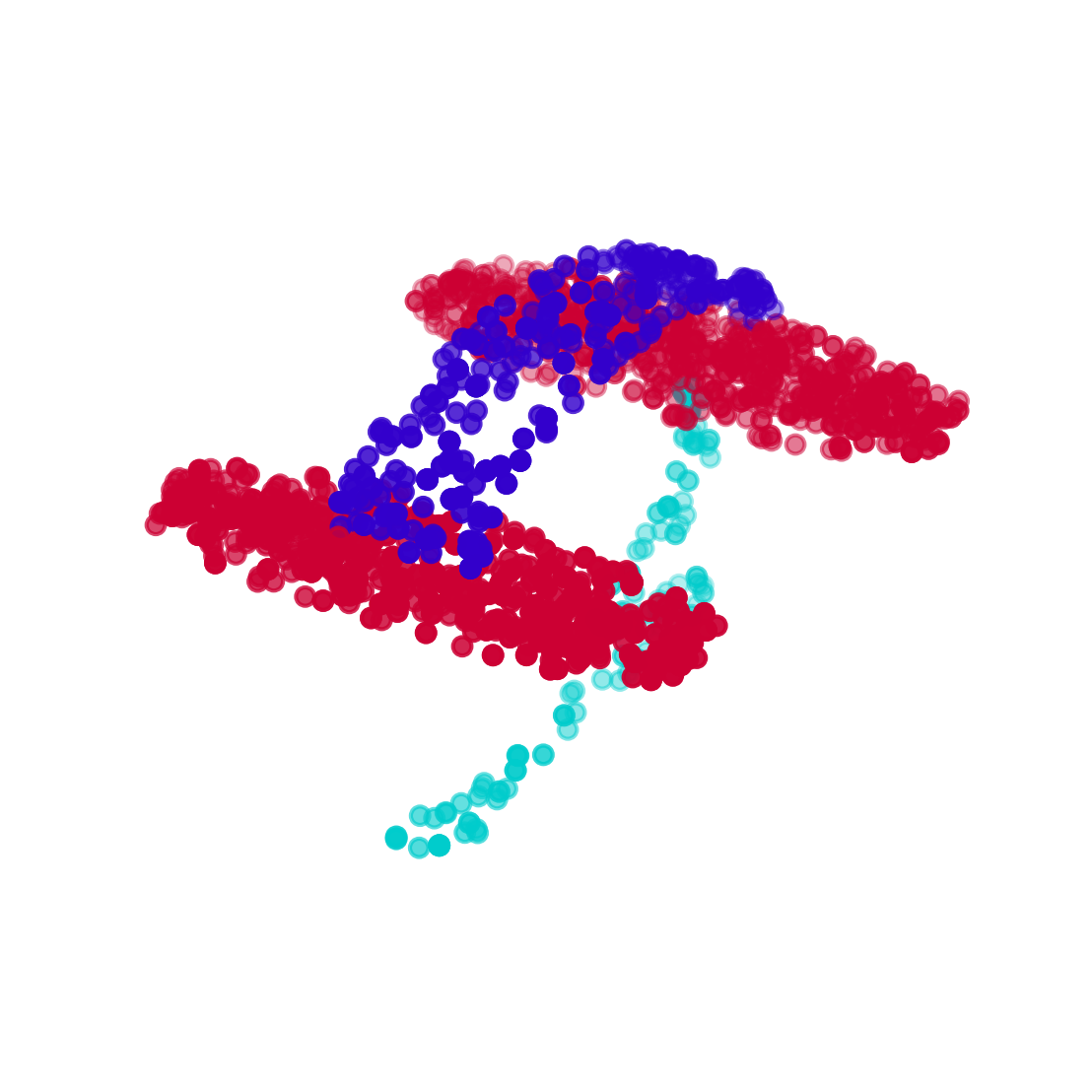}
        \caption*{Earphone, gt}
    \end{subfigure}
    \hfill
    \begin{subfigure}[b]{0.18\textwidth}
        \centering
        \includegraphics[width=\textwidth]{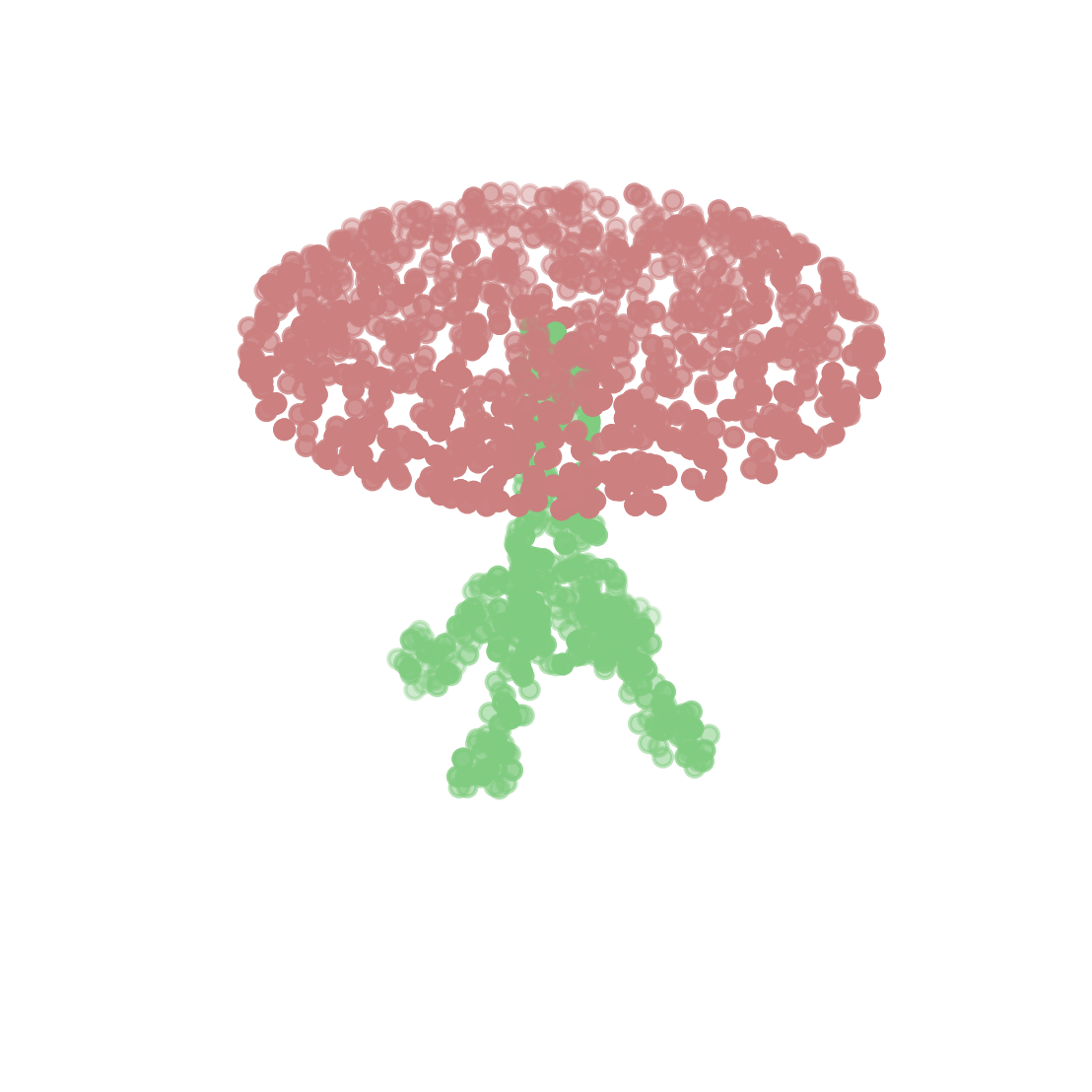}
        \caption*{Table, gt}
    \end{subfigure}

    \vspace{0.5cm}
    
    \begin{subfigure}[b]{0.18\textwidth}
        \centering
        \includegraphics[width=\textwidth]{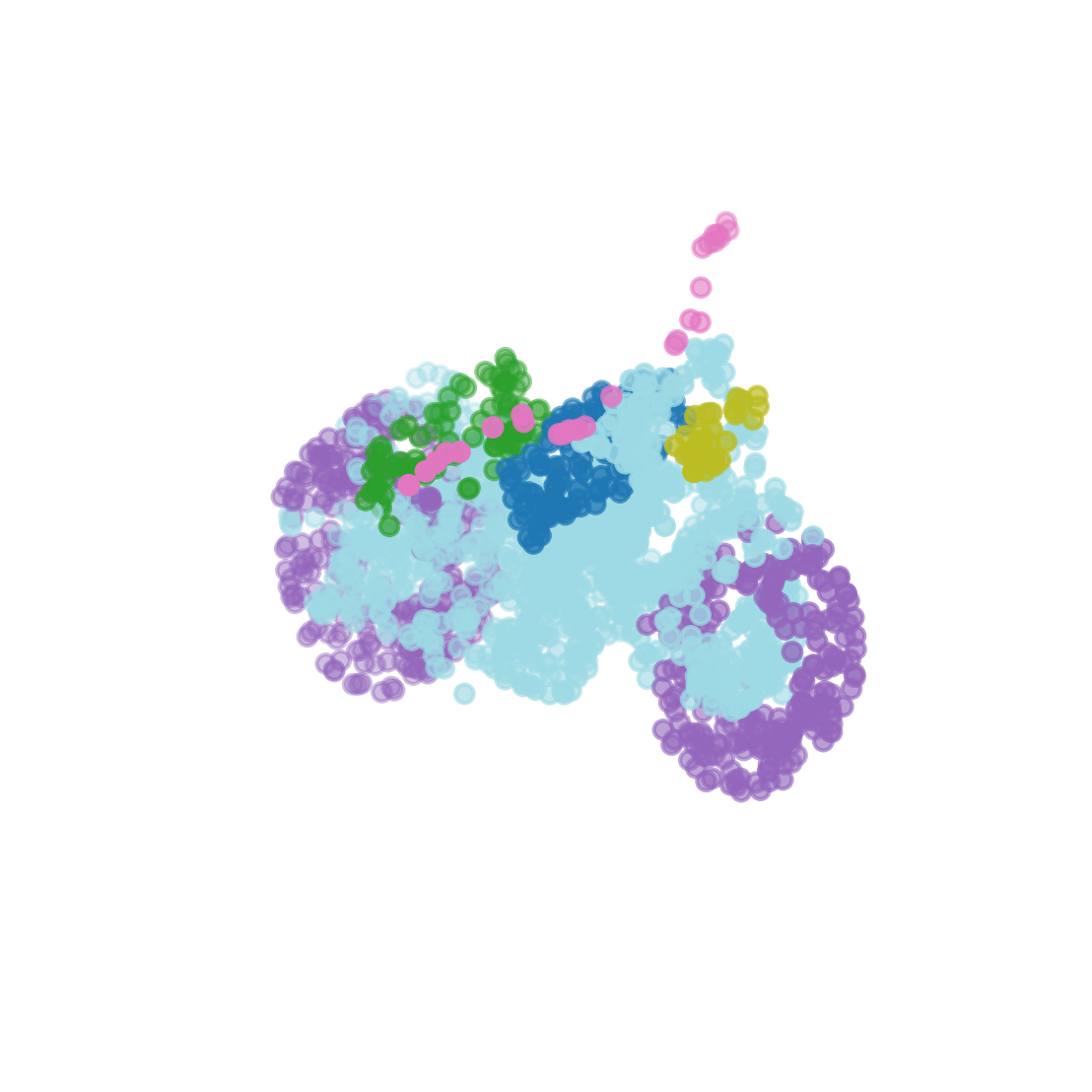}
        \caption*{Motor, p}
    \end{subfigure}
    \hfill
    \begin{subfigure}[b]{0.18\textwidth}
        \centering
        \includegraphics[width=\textwidth]{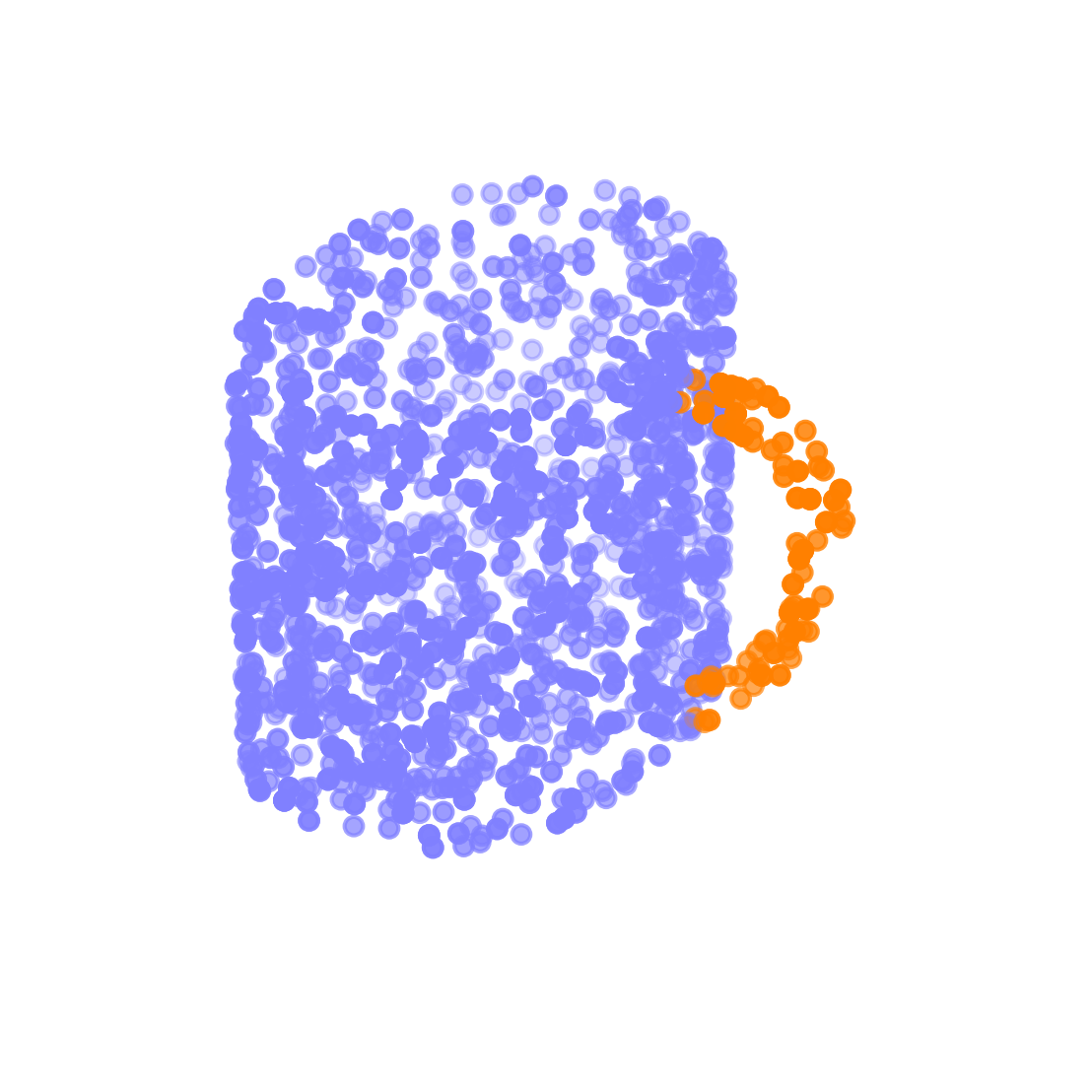}
        \caption*{Mug, p}
    \end{subfigure}
    \hfill
    \begin{subfigure}[b]{0.18\textwidth}
        \centering
        \includegraphics[width=\textwidth]{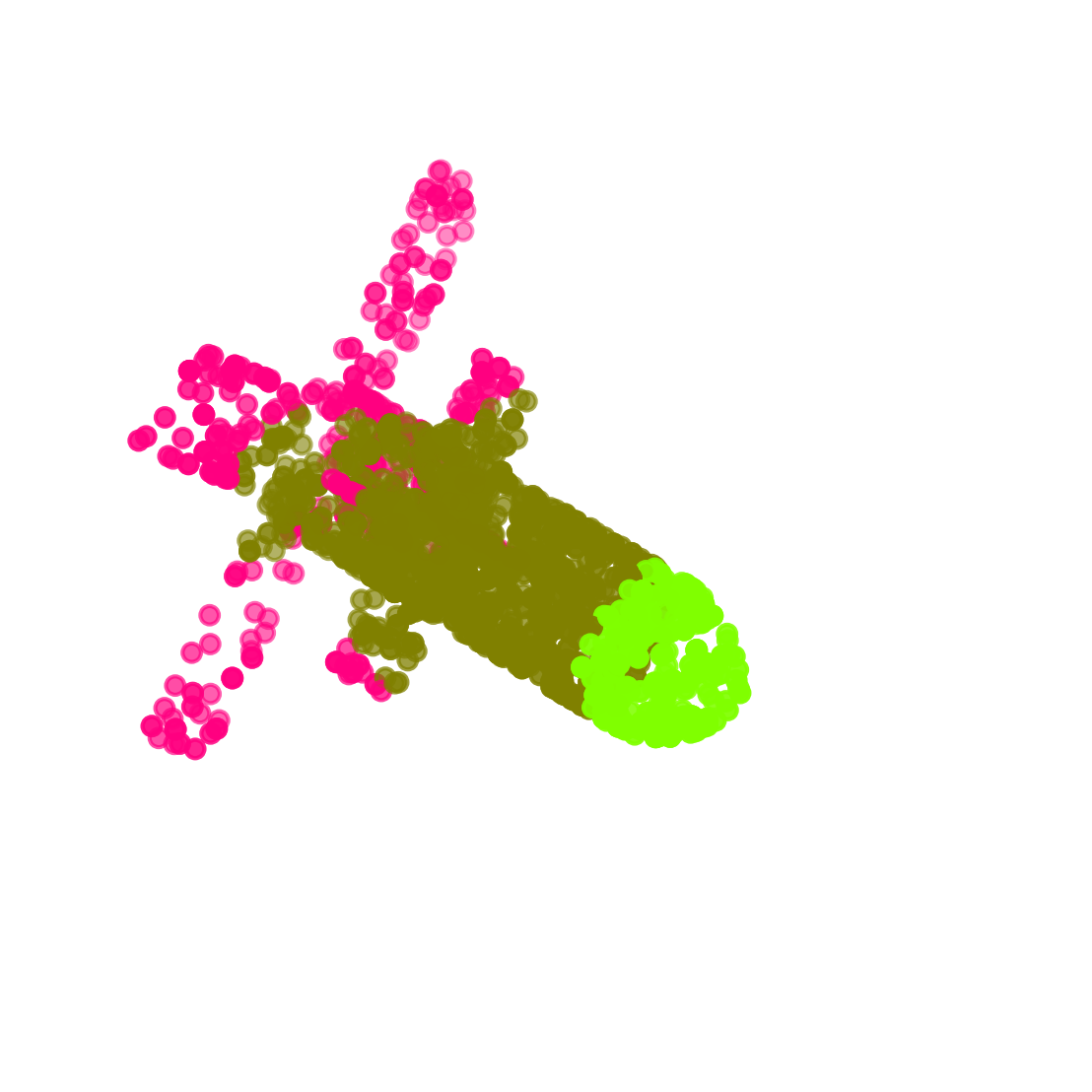}
        \caption*{Rocket, p}
    \end{subfigure}
    \hfill
    \begin{subfigure}[b]{0.18\textwidth}
        \centering
        \includegraphics[width=\textwidth]{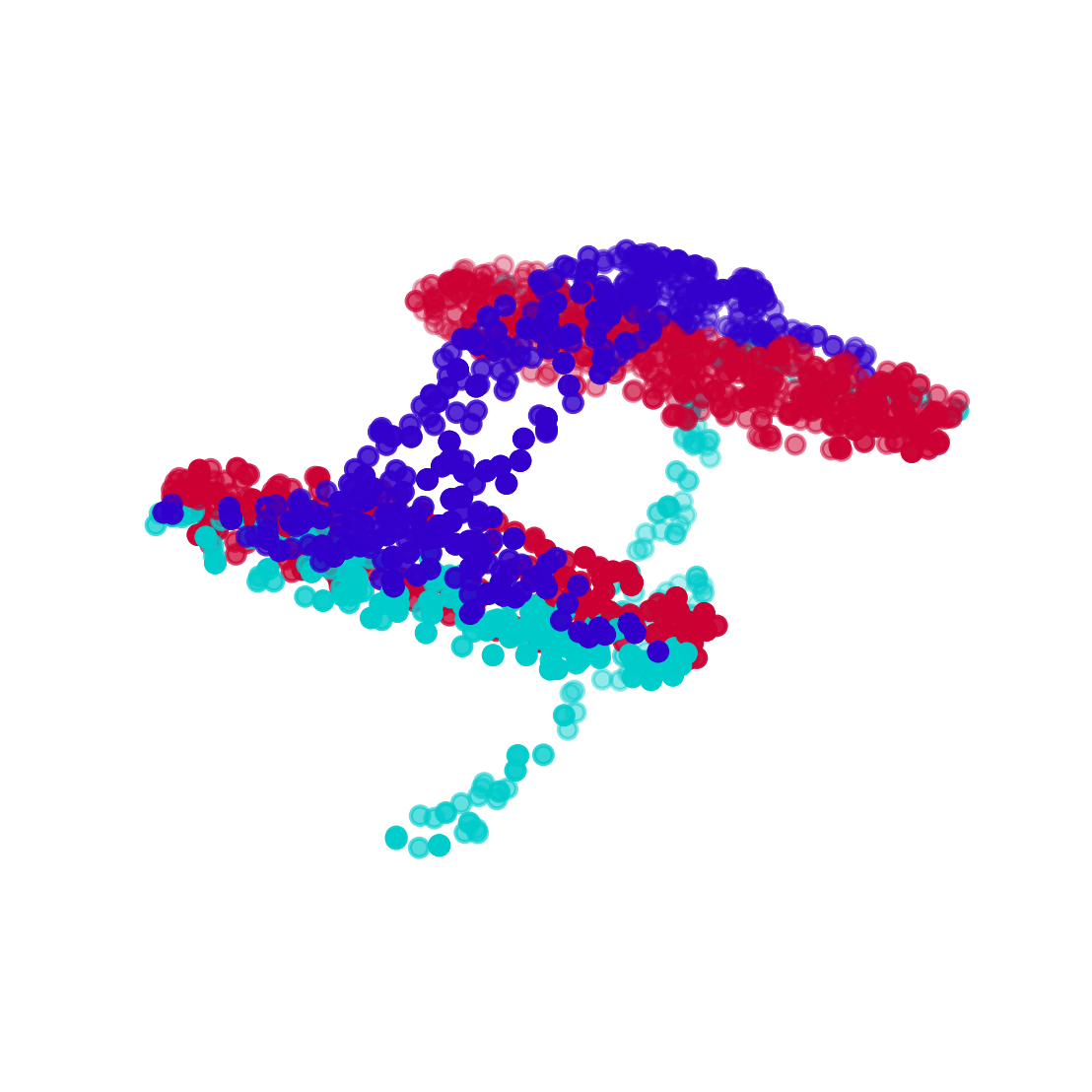}
        \caption*{Earphone, p}
    \end{subfigure}
    \hfill
    \begin{subfigure}[b]{0.18\textwidth}
        \centering
        \includegraphics[width=\textwidth]{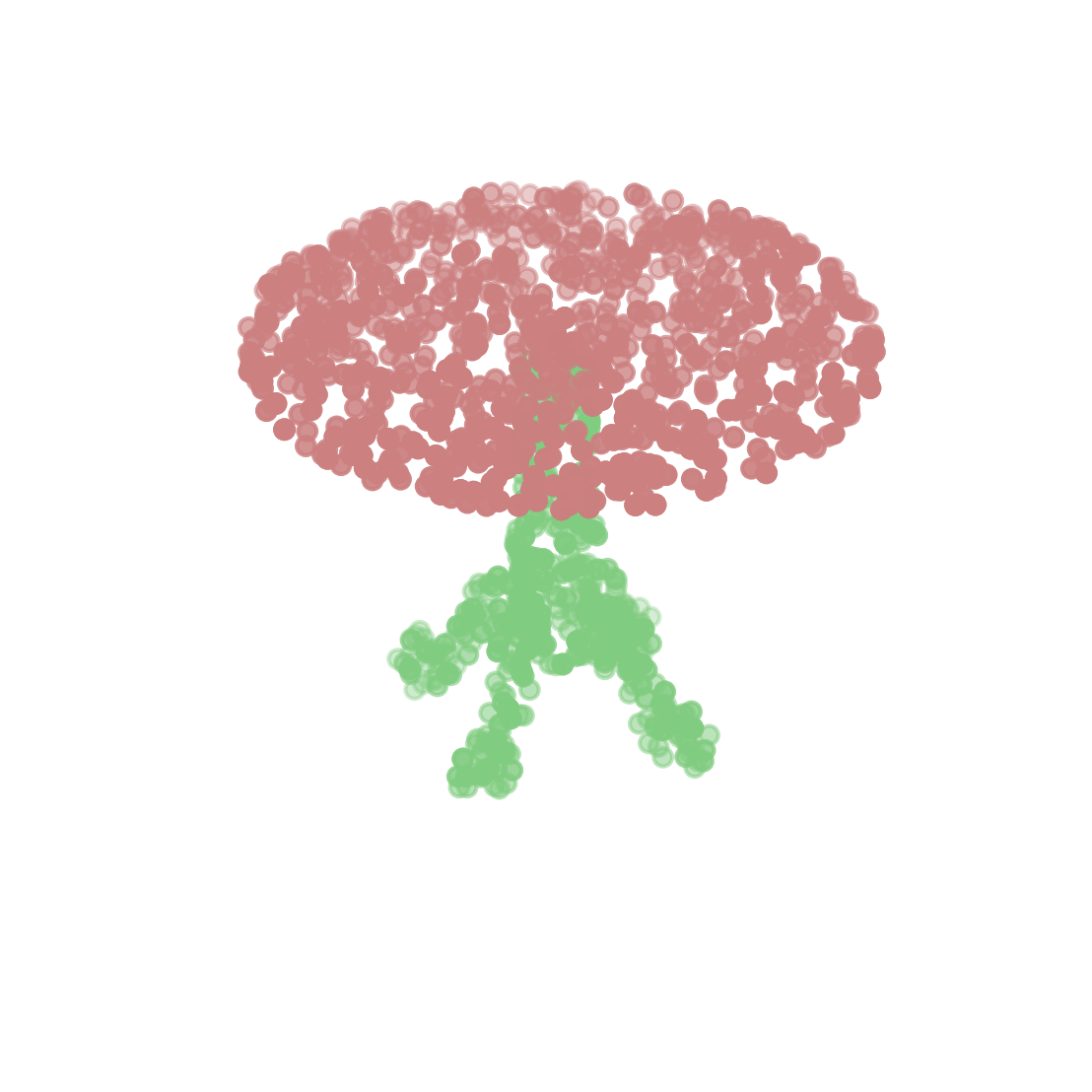}
        \caption*{Table, p}
    \end{subfigure}

    \begin{subfigure}[b]{0.18\textwidth}
        \centering
        \includegraphics[width=\textwidth]{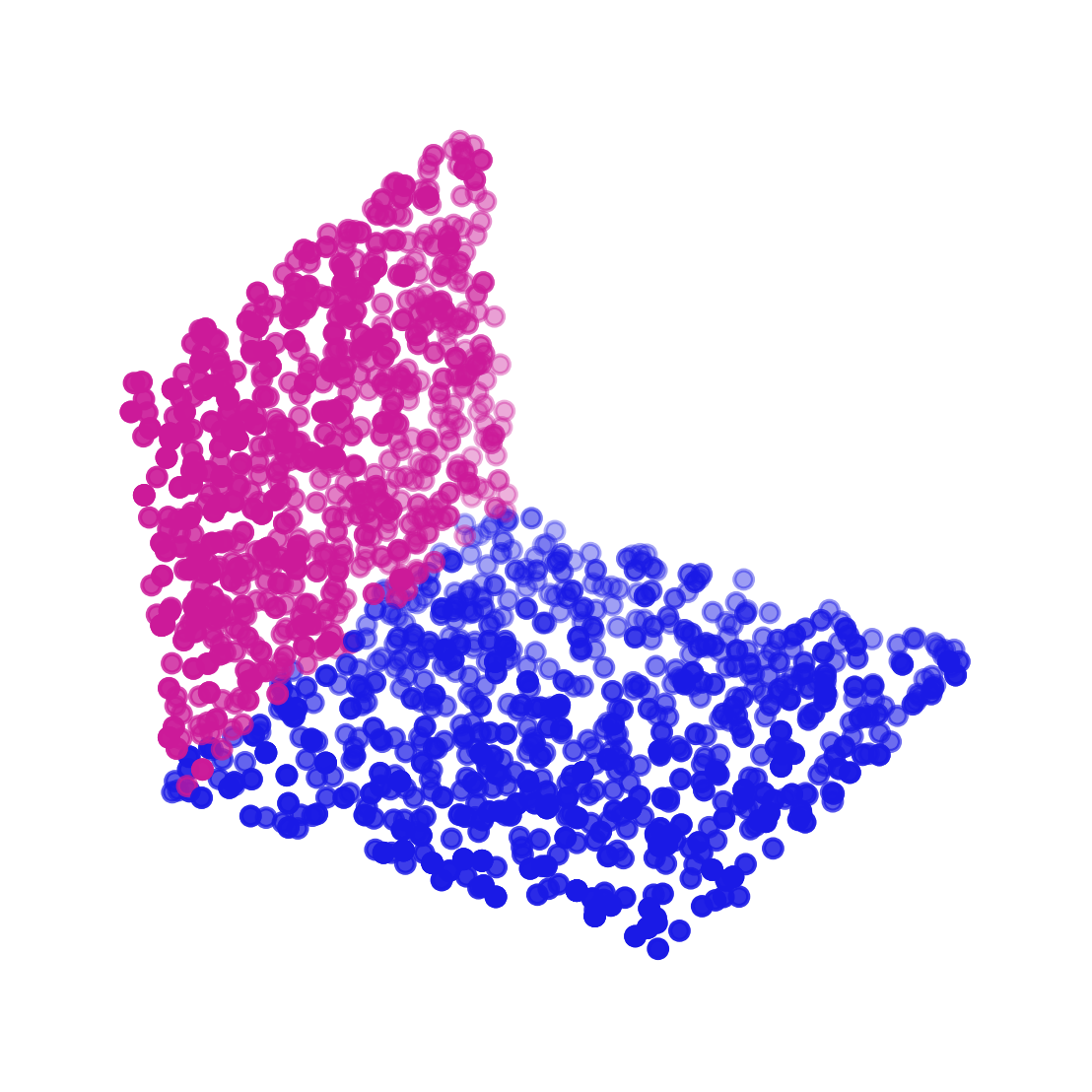}
        \caption*{Laptop, gt}
    \end{subfigure}
    \hfill
    \begin{subfigure}[b]{0.18\textwidth}
        \centering
         \includegraphics[width=\textwidth]{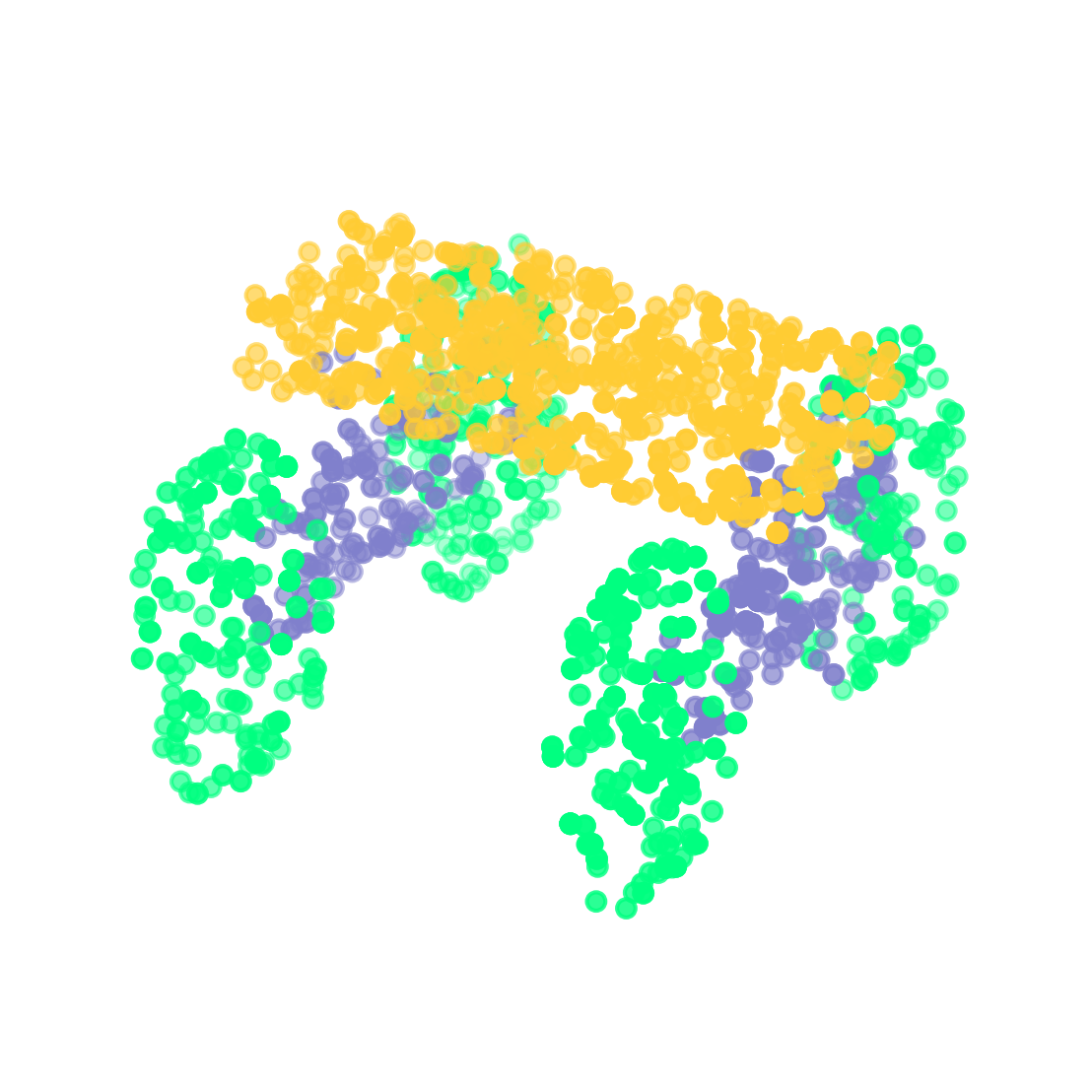}
        \caption*{Skateboard, gt}
    \end{subfigure}
    \hfill
    \begin{subfigure}[b]{0.18\textwidth}
        \centering
        \includegraphics[width=\textwidth]{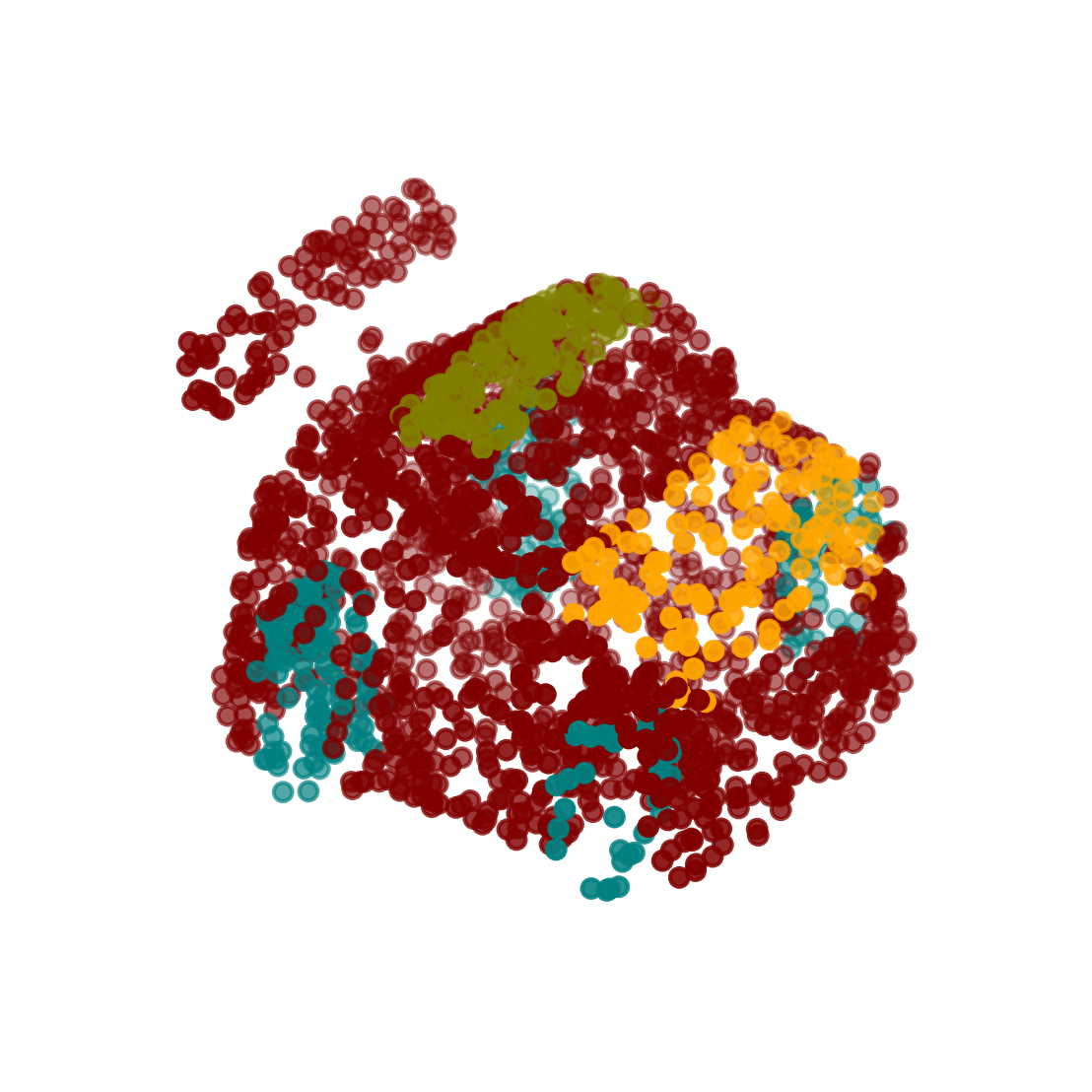}
        \caption*{Car, gt}
    \end{subfigure}
    \hfill
    \begin{subfigure}[b]{0.18\textwidth}
        \centering
        \includegraphics[width=\textwidth]{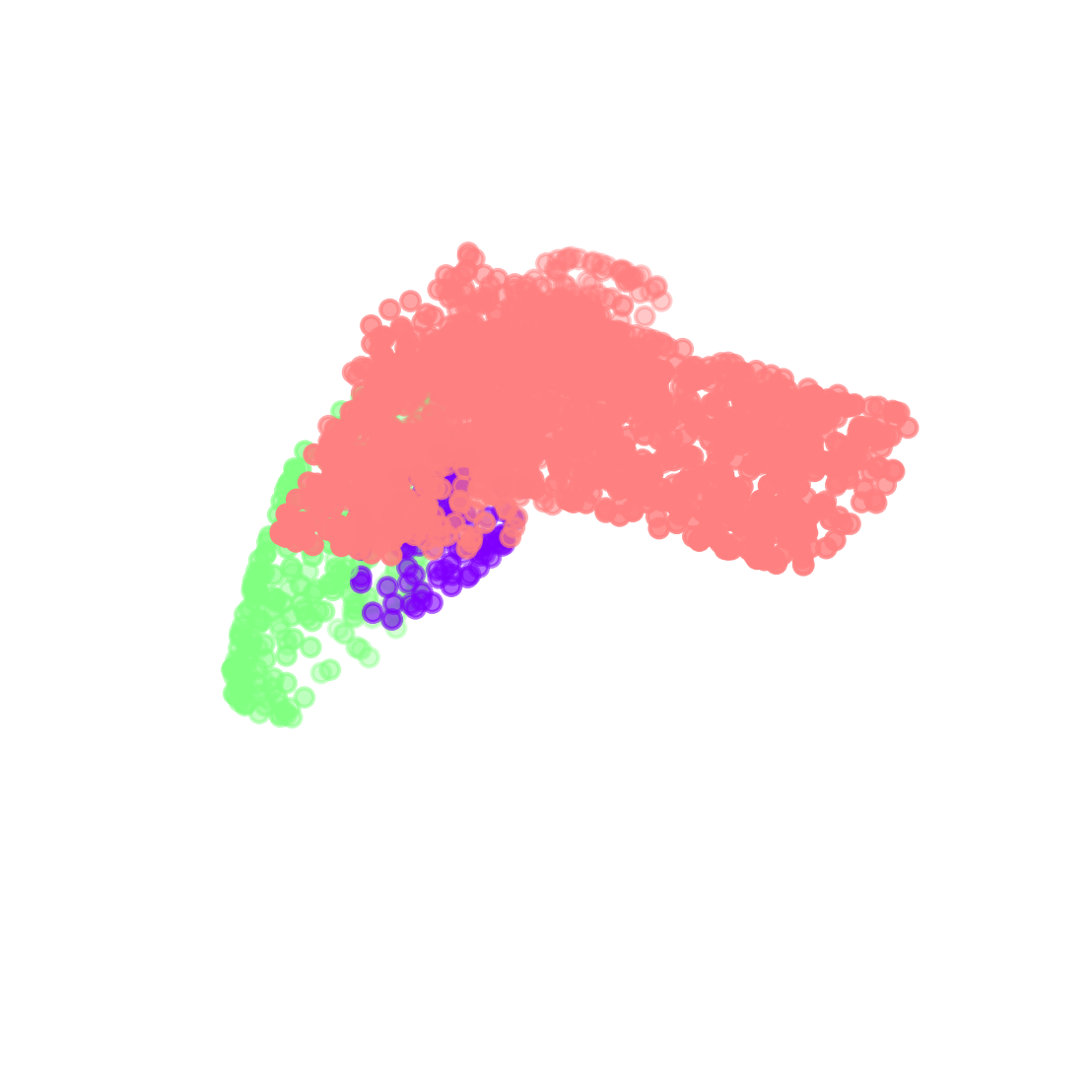}
        \caption*{Pistol, gt}
    \end{subfigure}
    \hfill
    \begin{subfigure}[b]{0.18\textwidth}
        \centering
        \includegraphics[width=\textwidth]{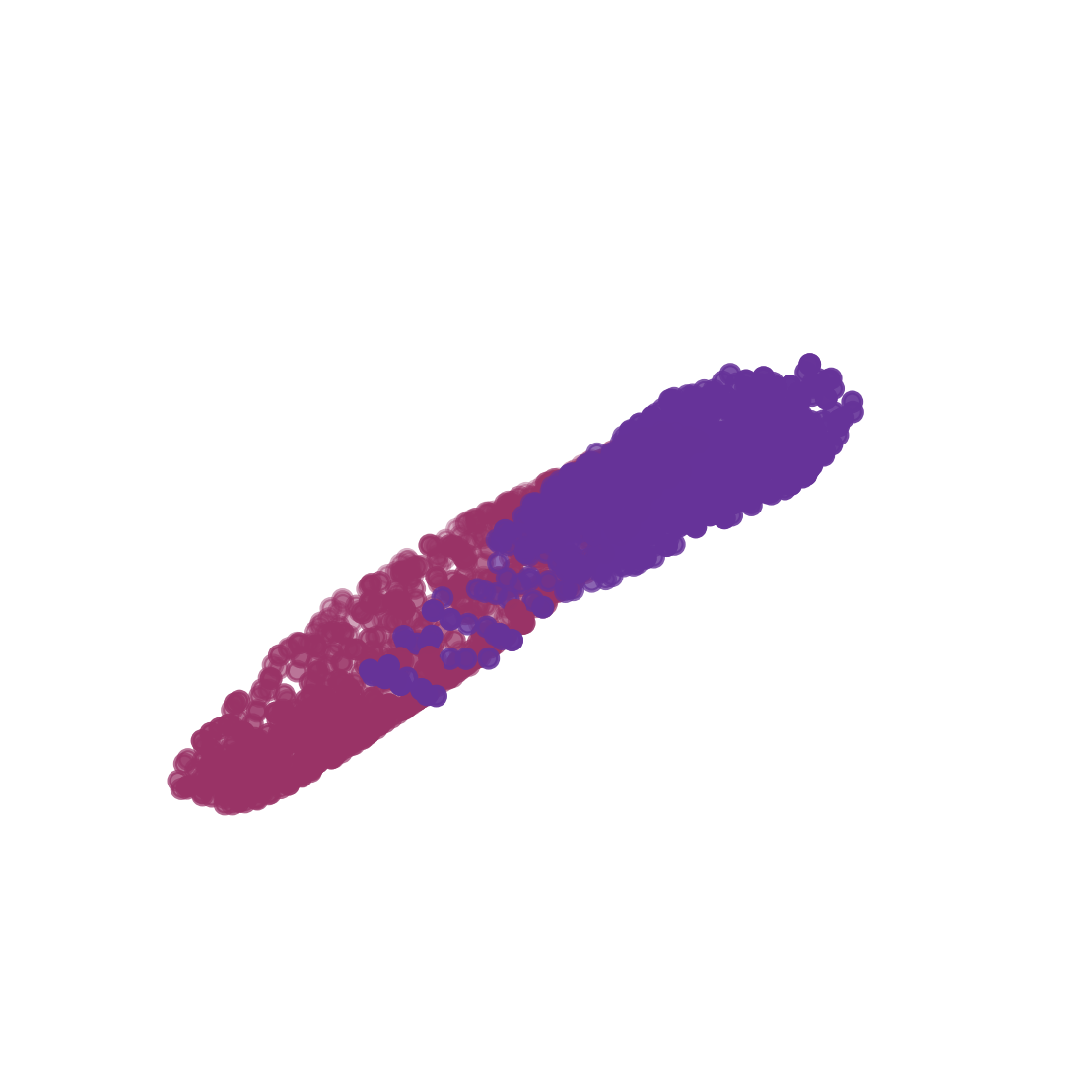}
        \caption*{Knife, gt}
    \end{subfigure}

    \vspace{0.5cm}
    
    \begin{subfigure}[b]{0.18\textwidth}
        \centering
        \includegraphics[width=\textwidth]{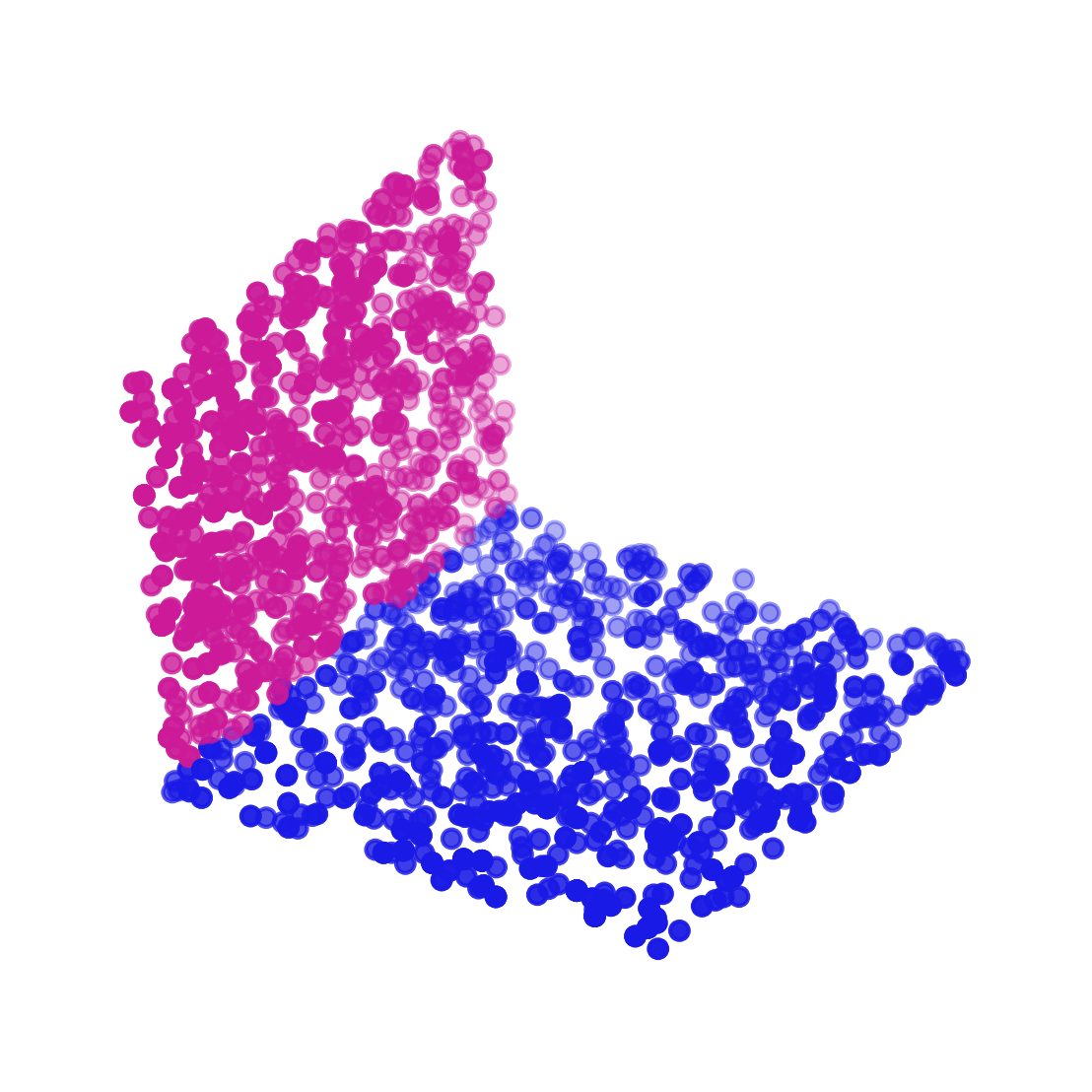}
        \caption*{Laptop, p}
    \end{subfigure}
    \hfill
    \begin{subfigure}[b]{0.18\textwidth}
        \centering
        \includegraphics[width=\textwidth]{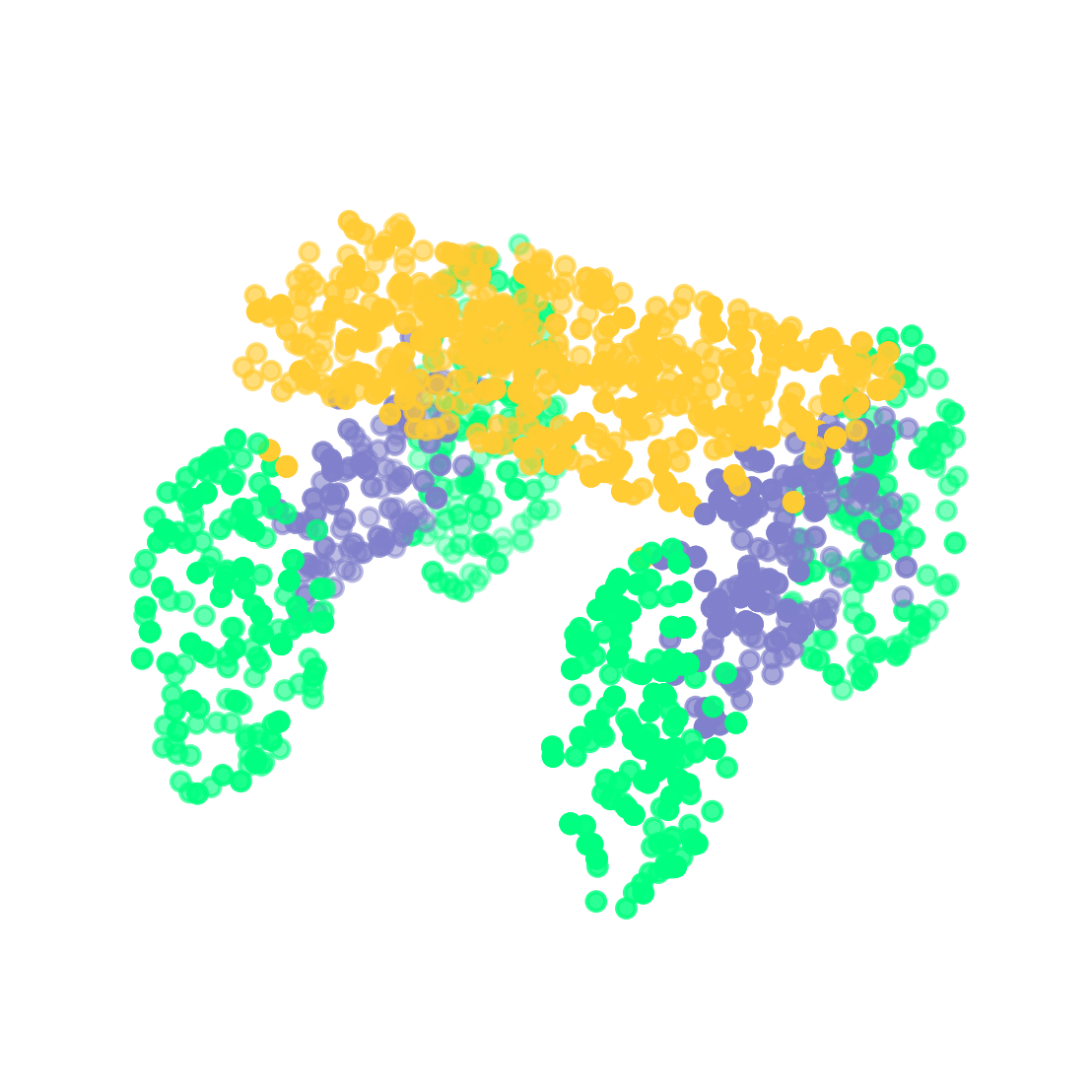}
        \caption*{Skateboard, p}
    \end{subfigure}
    \hfill
    \begin{subfigure}[b]{0.18\textwidth}
        \centering
        \includegraphics[width=\textwidth]{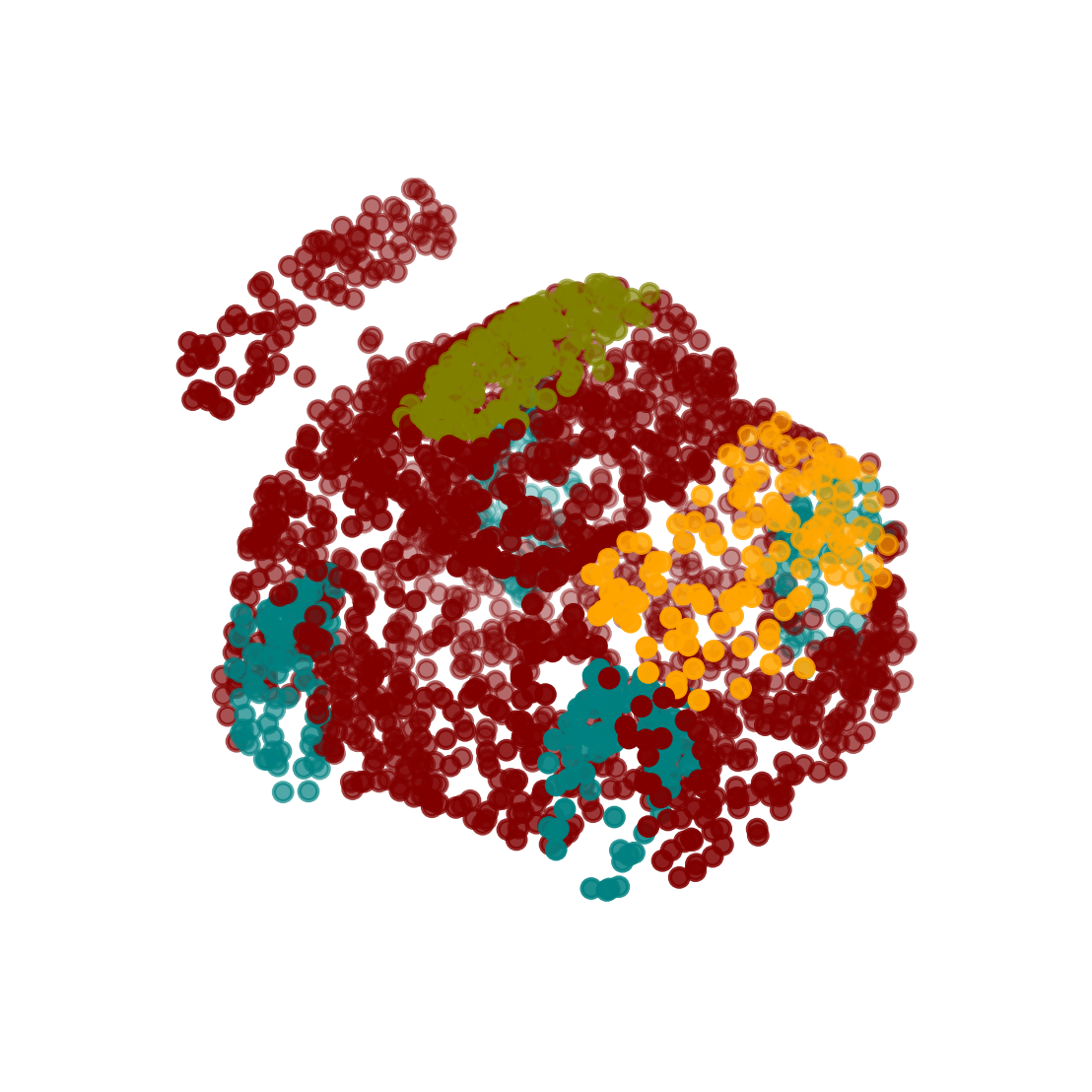}
        \caption*{Car, p}
    \end{subfigure}
    \hfill
    \begin{subfigure}[b]{0.18\textwidth}
        \centering
        \includegraphics[width=\textwidth]{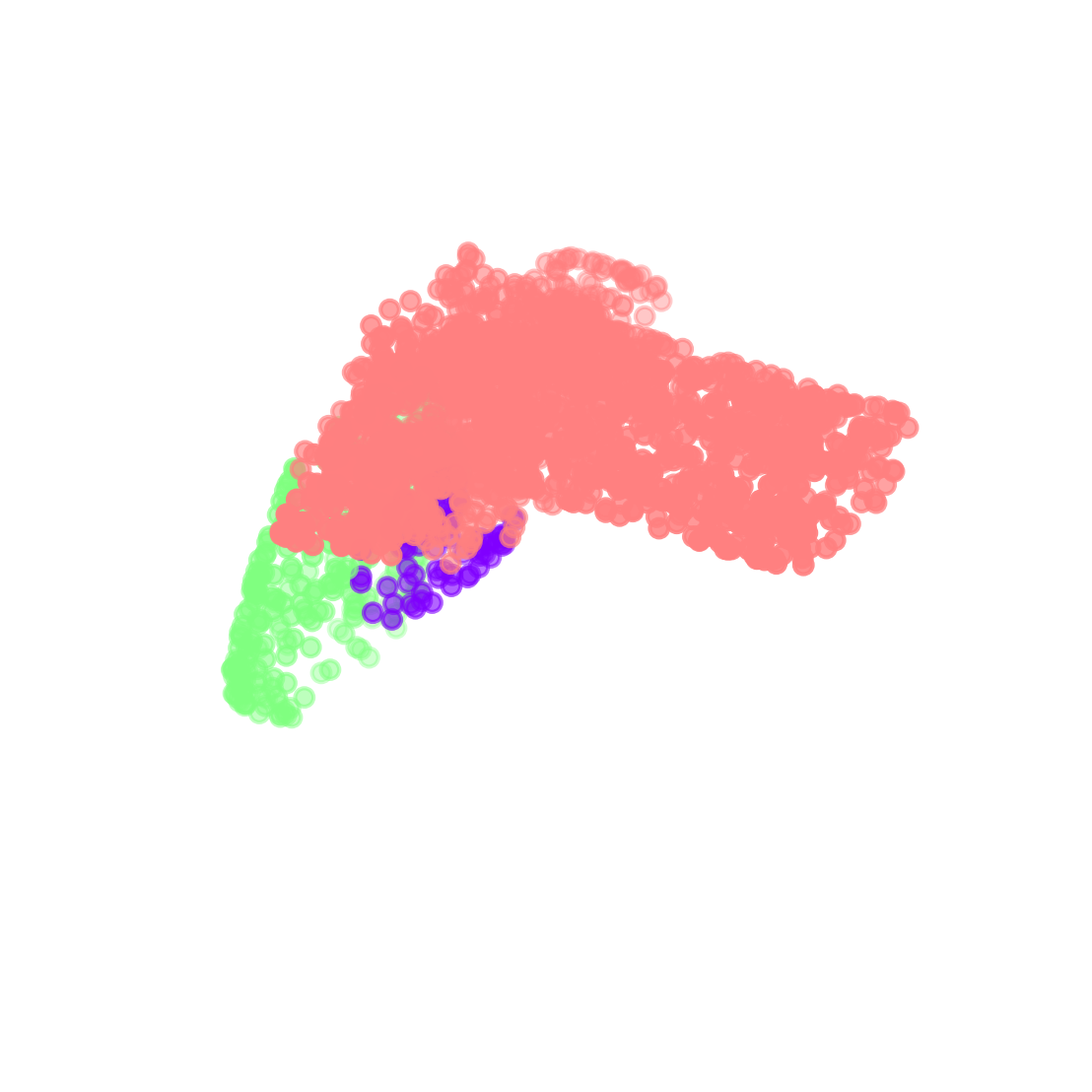}
        \caption*{Pistol, p}
    \end{subfigure}
    \hfill
    \begin{subfigure}[b]{0.18\textwidth}
        \centering
        \includegraphics[width=\textwidth]{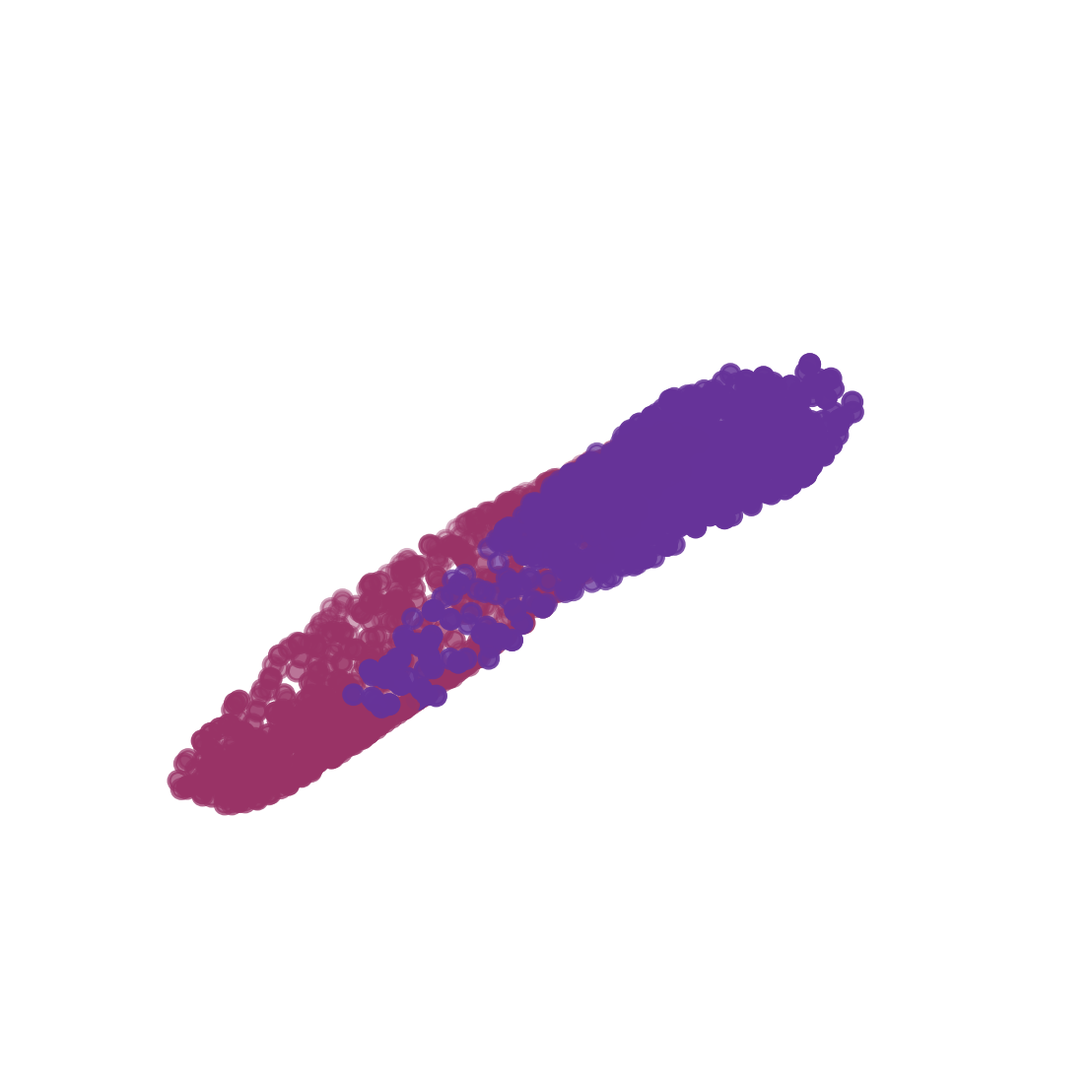}
        \caption*{Knife, p}
    \end{subfigure}

    \caption{A few qualitative results obtained by PointNet-KAN for part segmentation on the ShapeNet Part dataset \citep{yi2016scalable}. The results correspond to PointNet-KAN using a Jacobi polynomial of degree 2 with $\alpha = \beta = -0.5$. In the labels, `gt' represents the ground truth, and `p' represents prediction.}
    \label{Fig2}
\end{figure}

\subsection{3D object part segmentation}
\label{Sect42}

For the part segmentation task, we assess PointNet-KAN on the ShapeNet part dataset \citep{yi2016scalable}, which includes 16,881 shapes across 16 categories, with annotations for 50 distinct parts. The number of parts per category ranges from 2 to 6. We adhere to the official train, validation, and test splits as outlined in the literature \citep{chang2015shapenet,qi2017pointnet,wang2019dynamic}. In our experiment, we uniformly sample 2,048 points from each shape within a unit ball. The input features for PointNet-KAN consist solely of spatial coordinates, and normal vectors are not utilized (i.e., $d=3$). The evaluation metric used is Intersection-over-Union (IoU) on points, as described by \cite{qi2017pointnet}. Training details are provided in \ref{ATD}. Qualitative results for part segmentation are shown in Fig. \ref{Fig2}. The performance of PointNet-KAN compared to PointNet \cite{qi2017pointnet} is presented in Table \ref{Table4}. Accordingly, PointNet-KAN demonstrates competitive results compared to PointNet, with a mean IoU of 83.3\% versus 83.7\%. As shown in Table \ref{Table4}, for categories such as motorbike, pistol, and table, PointNet-KAN provides more accurate predictions than PointNet \cite{qi2017pointnet}. Based on our machine learning experiments, adding normal vectors as input features does not improve the performance of PointNet-KAN. A comparison between the segmentation branch of PointNet-KAN, shown in Fig. \ref{Fig1}, and the part segmentation branch of PointNet, shown in Fig. 9 of \cite{qi2017pointnet}, highlights the simplicity of the PointNet-KAN architecture, which consists of only 4 layers and uses a single local feature, whereas PointNet has 11 layers and uses 5 local features. Additionally, while PointNet includes input and feature transform networks, the PointNet-KAN architecture does not. Overall, PointNet-KAN outperforms earlier methodologies such as those in \cite{wu2014interactive}, 3DCNN \citep{qi2017pointnet}, and \cite{yi2016scalable}. However, more recent architectures, including DGCNN \citep{wang2019dynamic}, KPConv \citep{thomas2019kpconv}, and TAP \citep{wang2023take}, surpass PointNet-KAN. As discussed in Sect. \ref{Sect41}, incorporating KANs into the core of these networks as a replacement for MLPs could potentially enhance their performance.

We further investigate the slight performance decrease of PointNet-KAN compared to PointNet (83.3\% versus 83.7\%) in the part segmentation task. This observation can be explained as follows. KAN layers have been shown to be more prone to overfitting in different applications, including both computational physics and computer graphics \citep{zhang2024generalizationKAN,ji2024comprehensiveKAN,koenig2024kan,cang2024can}. While this characteristic is not necessarily problematic in physics-informed machine learning for solving inverse problems \citep{WANG2025kan,shukla2024comprehensive}, where the learnable activation functions in KAN layers can better fit sparse sensor data, it may negatively impact generalization in computer graphics tasks, potentially reducing performance on the test set. To overcome this challenge, we propose PointNet-KAN-MLP, in which we simply replace the shared KAN decoder layers of size $(640,m)$ (see the bottom panel of Fig. \ref{Fig1}) with shared MLP layers of the same size. The results are presented in Table \ref{Table4}. We observe a mean IoU of 83.9\%, which surpasses the performance of both PointNet (83.3\%) and PointNet-KAN (83.7\%). It is concluded that the shared KAN layers effectively extract global features as the encoder, while the shared MLP layers are well-suited for mapping these global features to predict the label of each part as the decoder.


\begin{table}[]
\caption{Effect of Jacobi polynomial degree on classification performance of PointNet-KAN with the choice of $\alpha = \beta = 1.0$ on ModelNet40 \citep{wu20153d}. Normal vectors are included as part of the input features.}
\label{Table2}
\begin{center}
\begin{tabular}{c|ccccc}
\toprule
Jacobi polynomial degree $(n)$ & 2 & 3 & 4 & 5 & 6 \\
\midrule
Number of trainable parameters & 620928   &  823680 & 1026432   & 1229184   &  1431936 \\
Mean class accuracy & 86.7   &  87.0   & 87.2   & 86.8   &  86.1 \\
Overall accuracy   & 89.9   &  90.4   &  90.5   &  89.9   &  89.9 \\
\bottomrule
\end{tabular}
\end{center}
\end{table}


\begin{table}[]
\caption{Effect of the choice of $\alpha$ and $\beta$ in Jacobi polynomials on the classification performance of PointNet-KAN, using a polynomial of degree 2 (i.e., $n=2$ in Eq. \ref{Eq3}), on ModelNet40 \citep{wu20153d}. Note that $\alpha = \beta = 0$ corresponds to the Legendre polynomial, $\alpha = \beta = -0.5$ corresponds to the Chebyshev polynomial of the first kind, $\alpha = \beta = 0.5$ corresponds to the Chebyshev polynomial of the second kind, and, in general, $\alpha = \beta$ corresponds to the Gegenbauer polynomial. Normal vectors are included as part of the input features.}
\label{Table3}
\vspace{1mm}
\begin{adjustbox}{width=1\textwidth}
\begin{tabular}{c|cccccc}
\toprule
Polynomial type & $\alpha=\beta=0$ & $\alpha=\beta=-0.5$ & $\alpha=\beta=0.5$ & $\alpha=\beta=1$ & $2\alpha=\beta=2$ & $\alpha=2\beta=2$ \\
\midrule
Mean class accuracy & 85.6   &  86.0   & 86.7   & 86.7   &  85.4 & 86.2 \\
Overall accuracy   & 89.5   &  89.9   &  90.1   &  89.9   &  89.4 &  89.8 \\
\bottomrule
\end{tabular}
\end{adjustbox}
\end{table}


\subsection{Semantic segmentation in 3D scenes}
\label{Sect400}

\begin{figure}[h]
    \centering
    \begin{subfigure}[b]{0.24\textwidth}
        \centering
        \includegraphics[width=\textwidth]{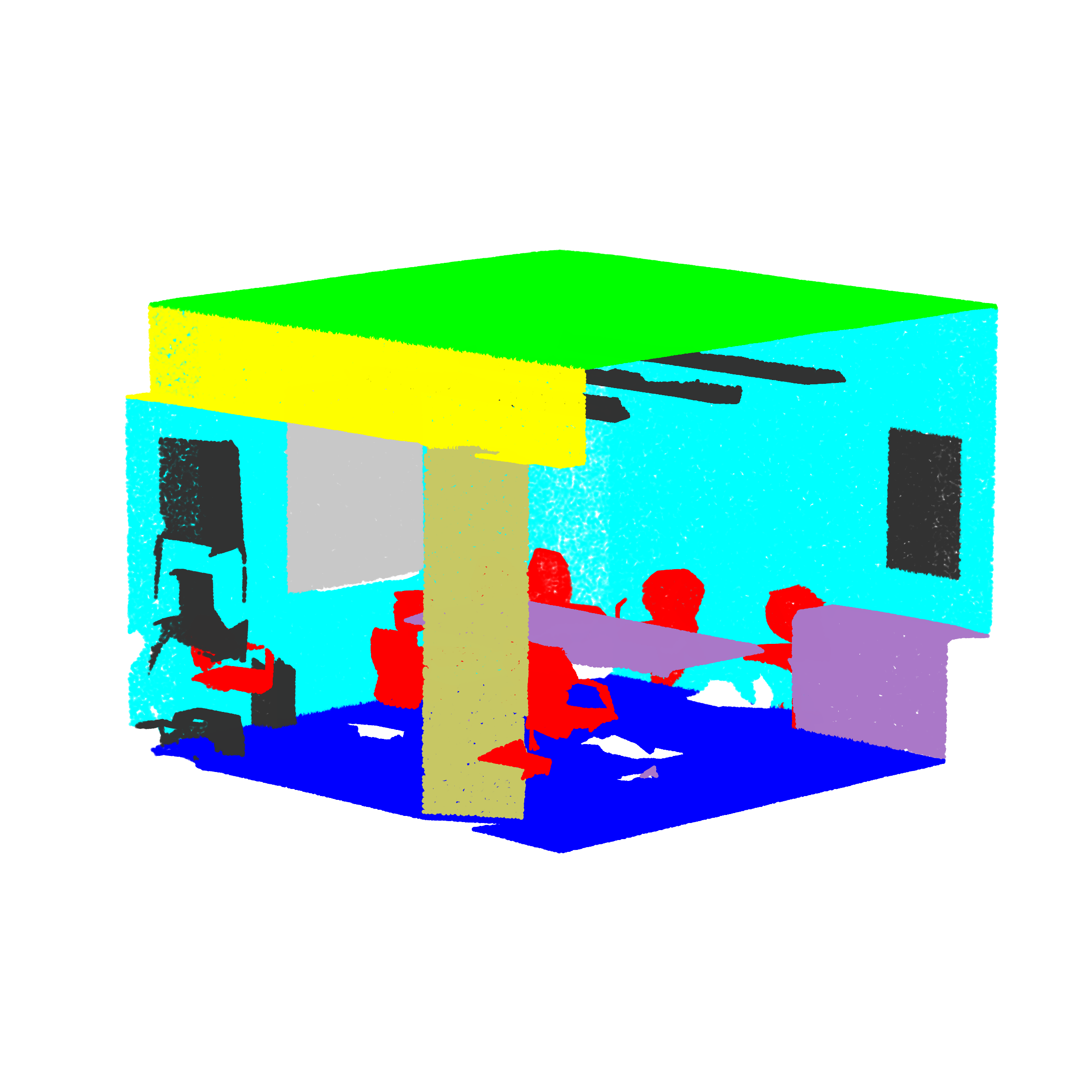}
    \end{subfigure}
    \hfill
    \begin{subfigure}[b]{0.24\textwidth}
        \centering
        \includegraphics[width=\textwidth]{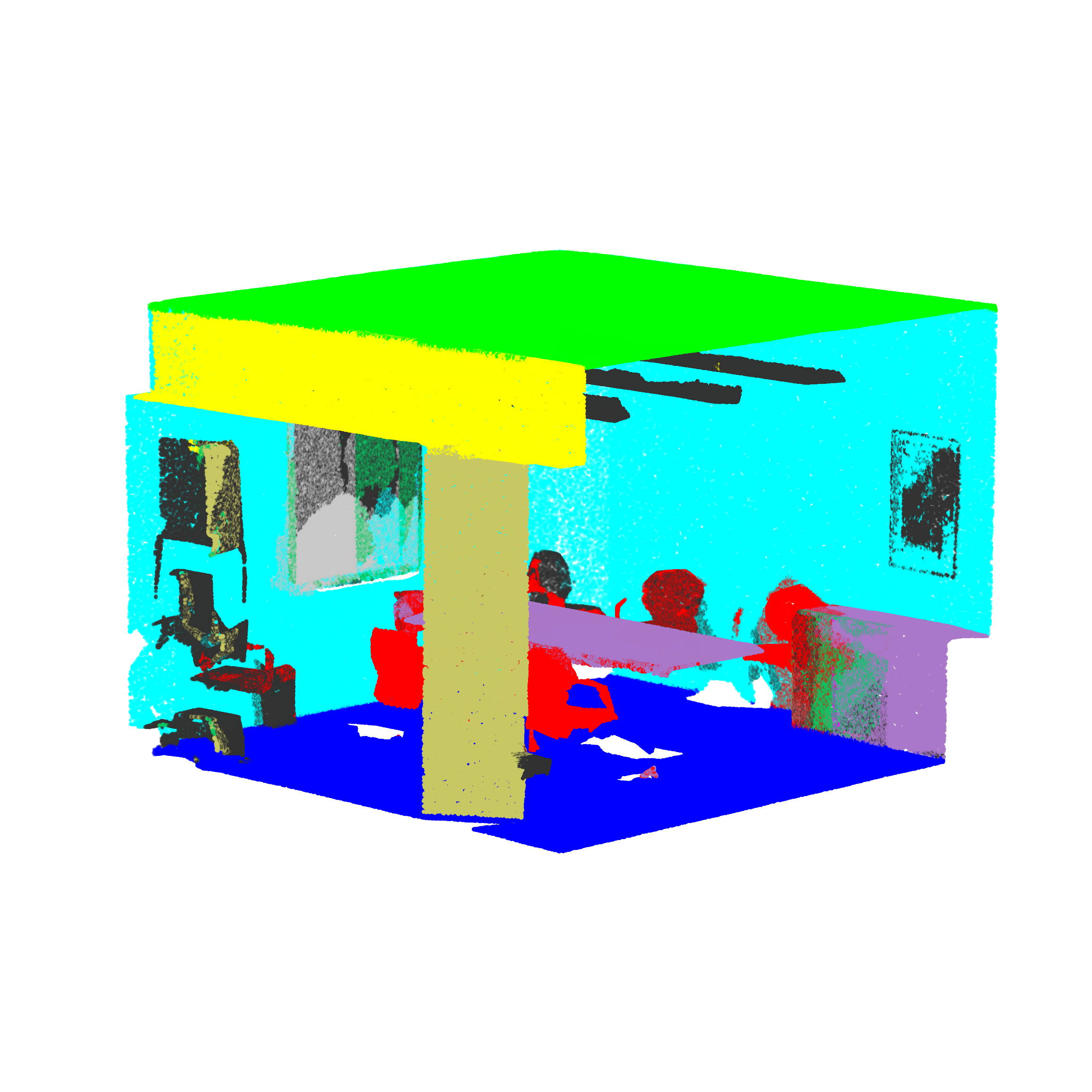}
    \end{subfigure}
    \hfill
    \begin{subfigure}[b]{0.24\textwidth}
        \centering
        \includegraphics[width=\textwidth]{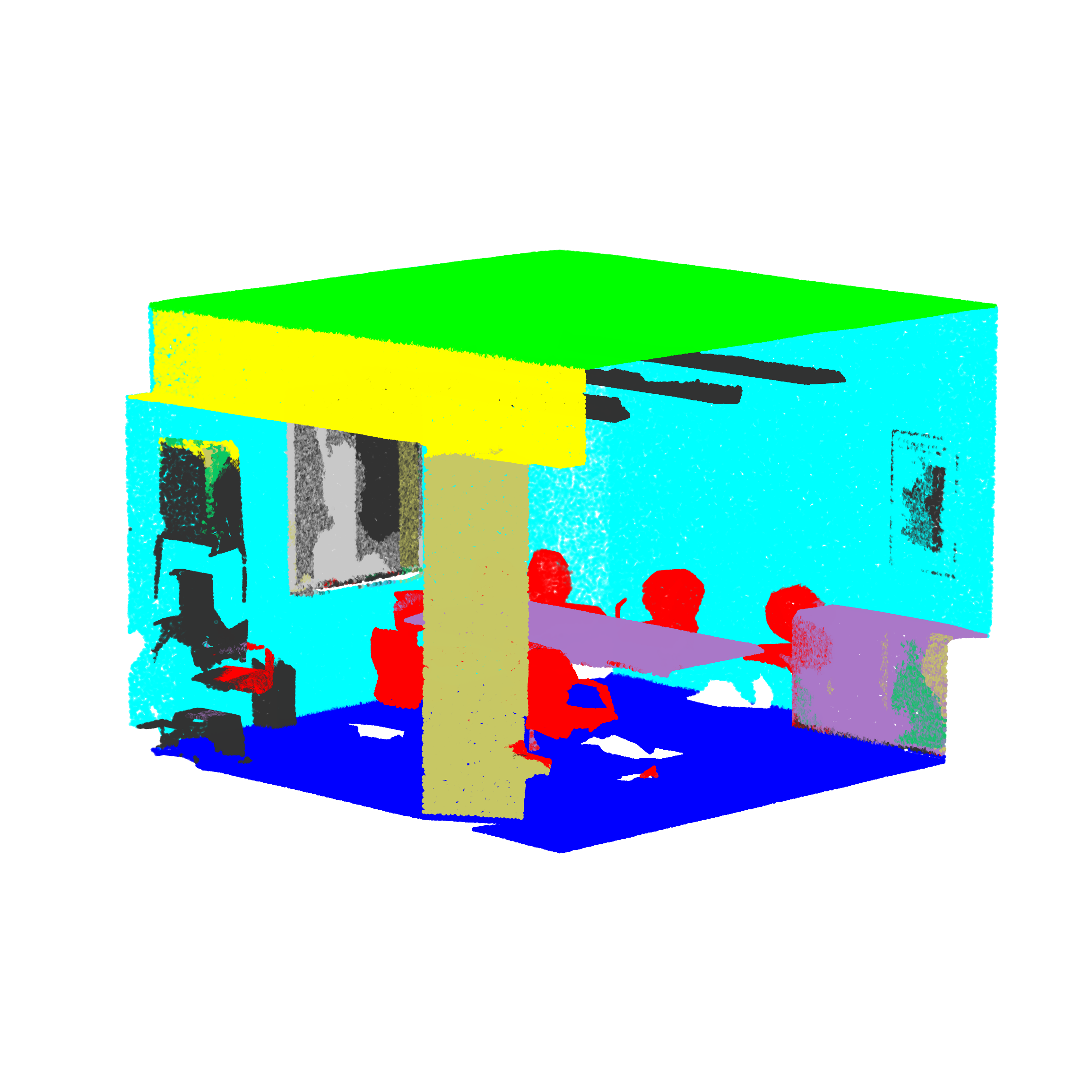}
    \end{subfigure}
    \hfill
    \begin{subfigure}[b]{0.24\textwidth}
        \centering
        \includegraphics[width=\textwidth]{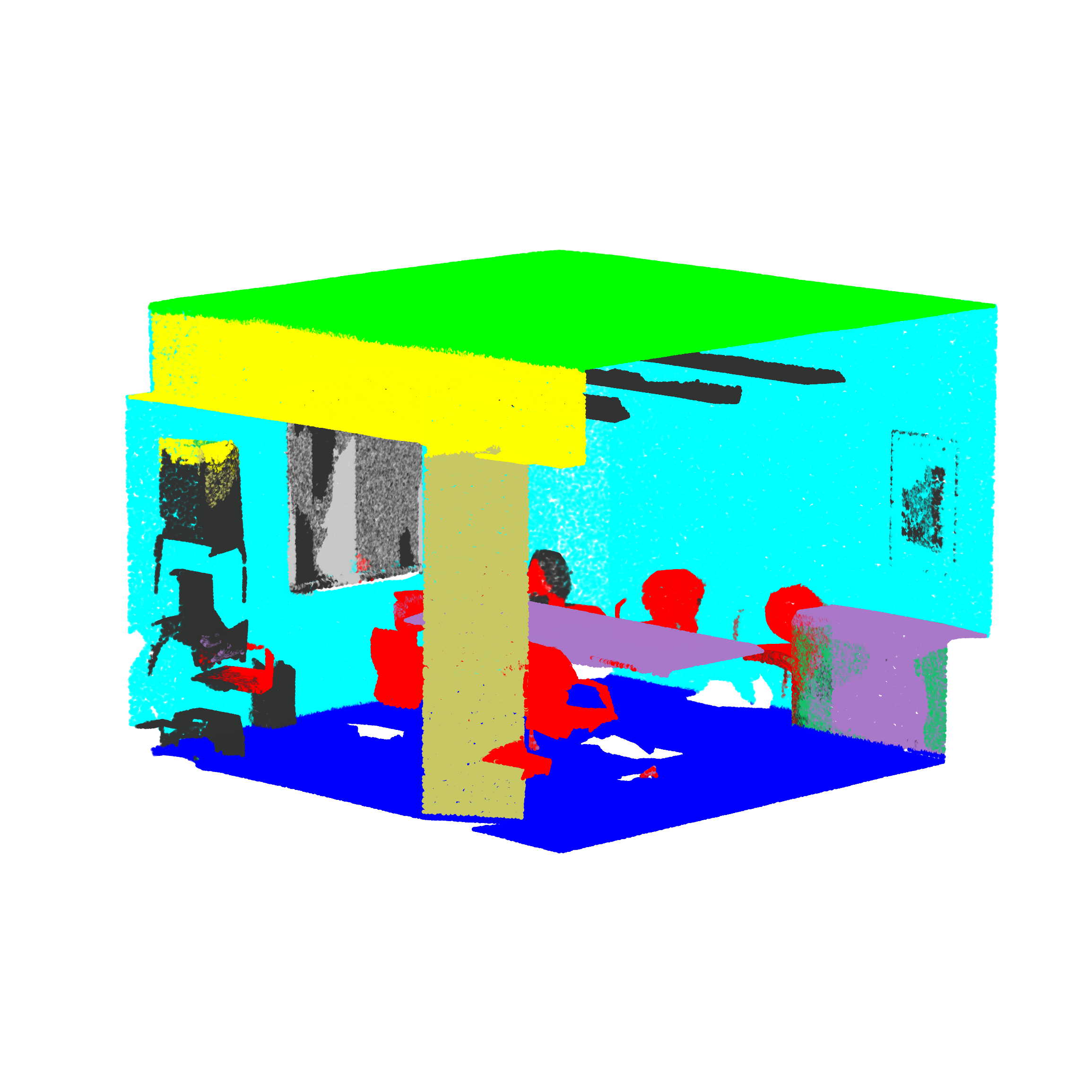}
    \end{subfigure}

    \begin{subfigure}[b]{0.24\textwidth}
        \centering
        \includegraphics[width=\textwidth]{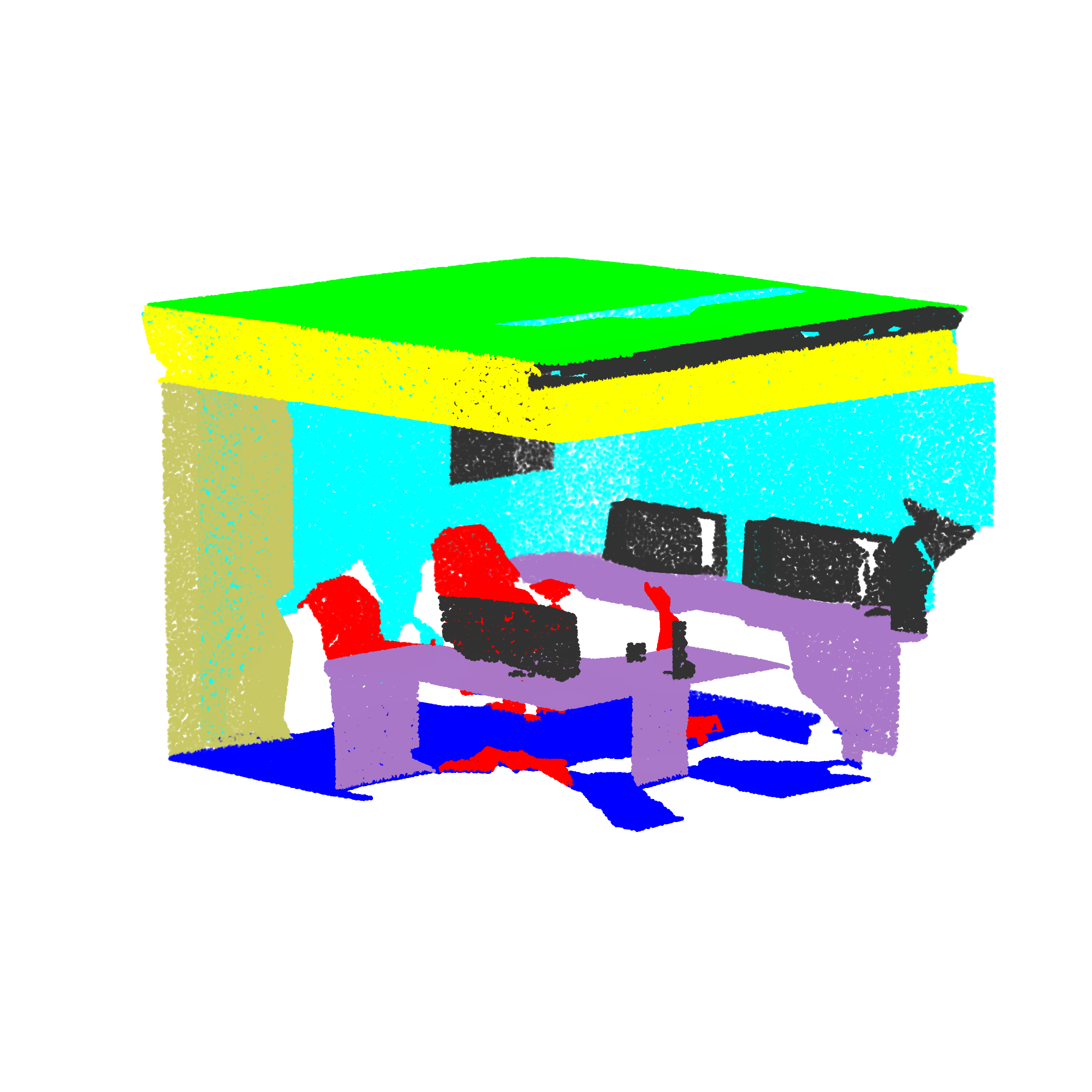}
    \end{subfigure}
    \hfill
    \begin{subfigure}[b]{0.24\textwidth}
        \centering
        \includegraphics[width=\textwidth]{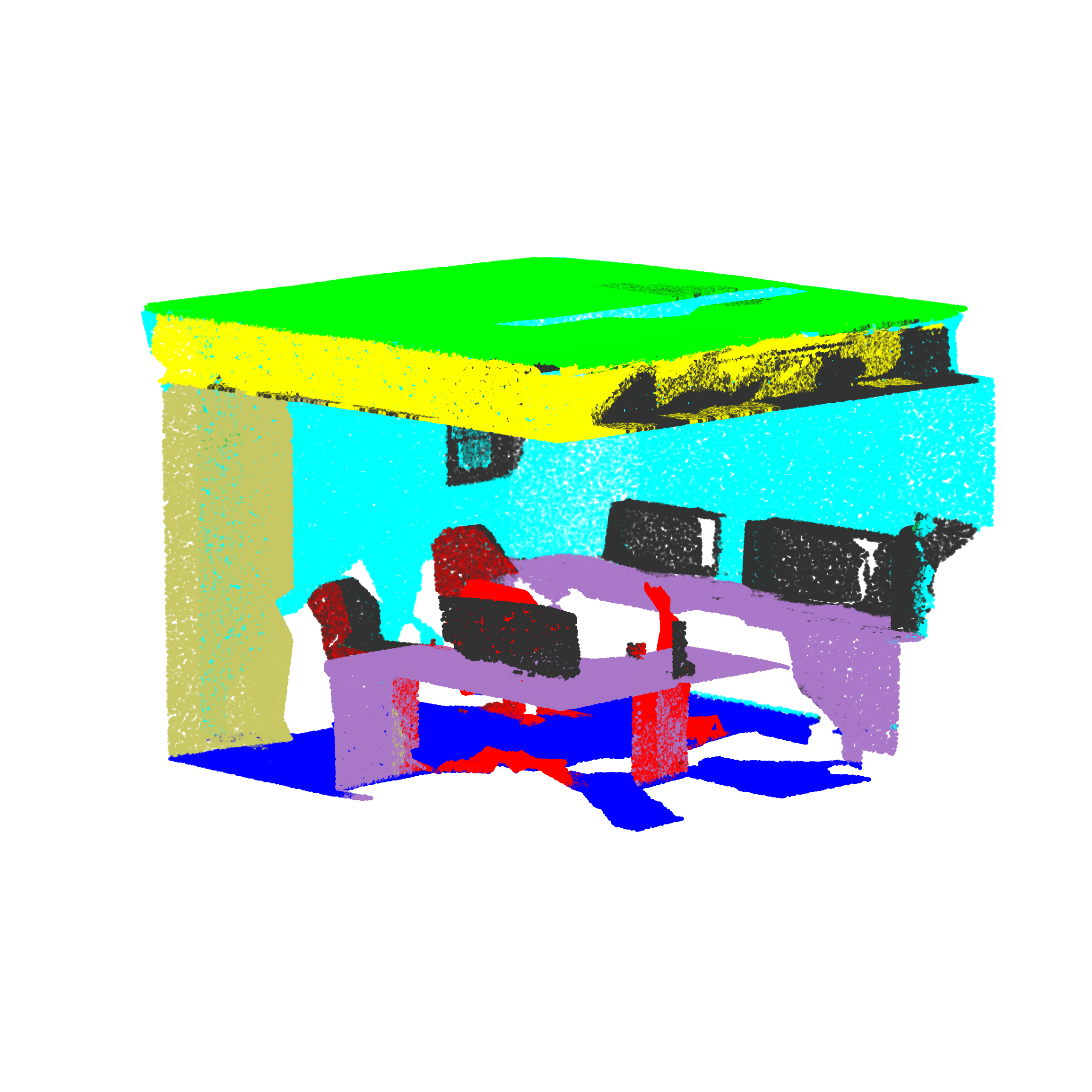}
    \end{subfigure}
    \hfill
    \begin{subfigure}[b]{0.24\textwidth}
        \centering
        \includegraphics[width=\textwidth]{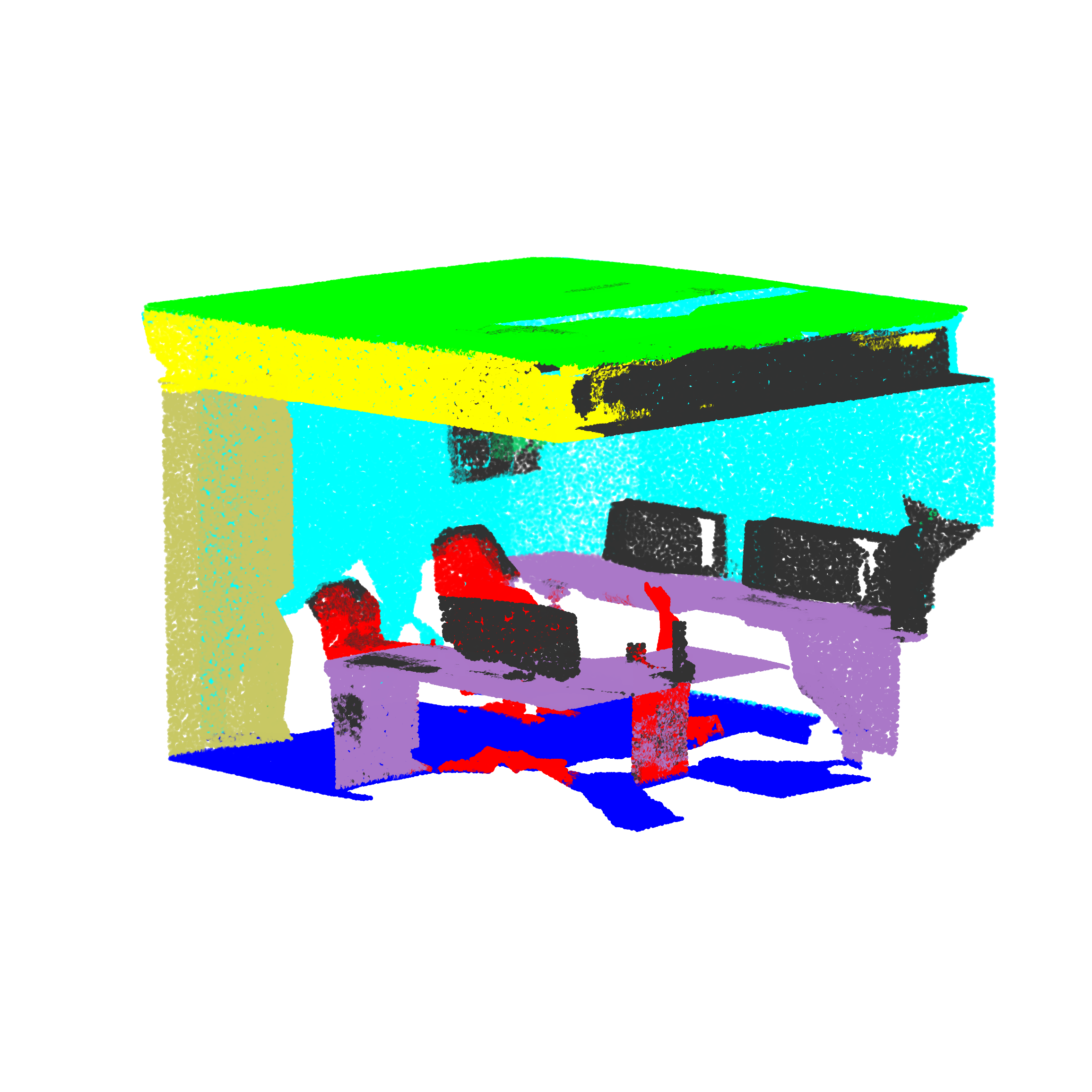}
    \end{subfigure}
    \hfill
    \begin{subfigure}[b]{0.24\textwidth}
        \centering
        \includegraphics[width=\textwidth]{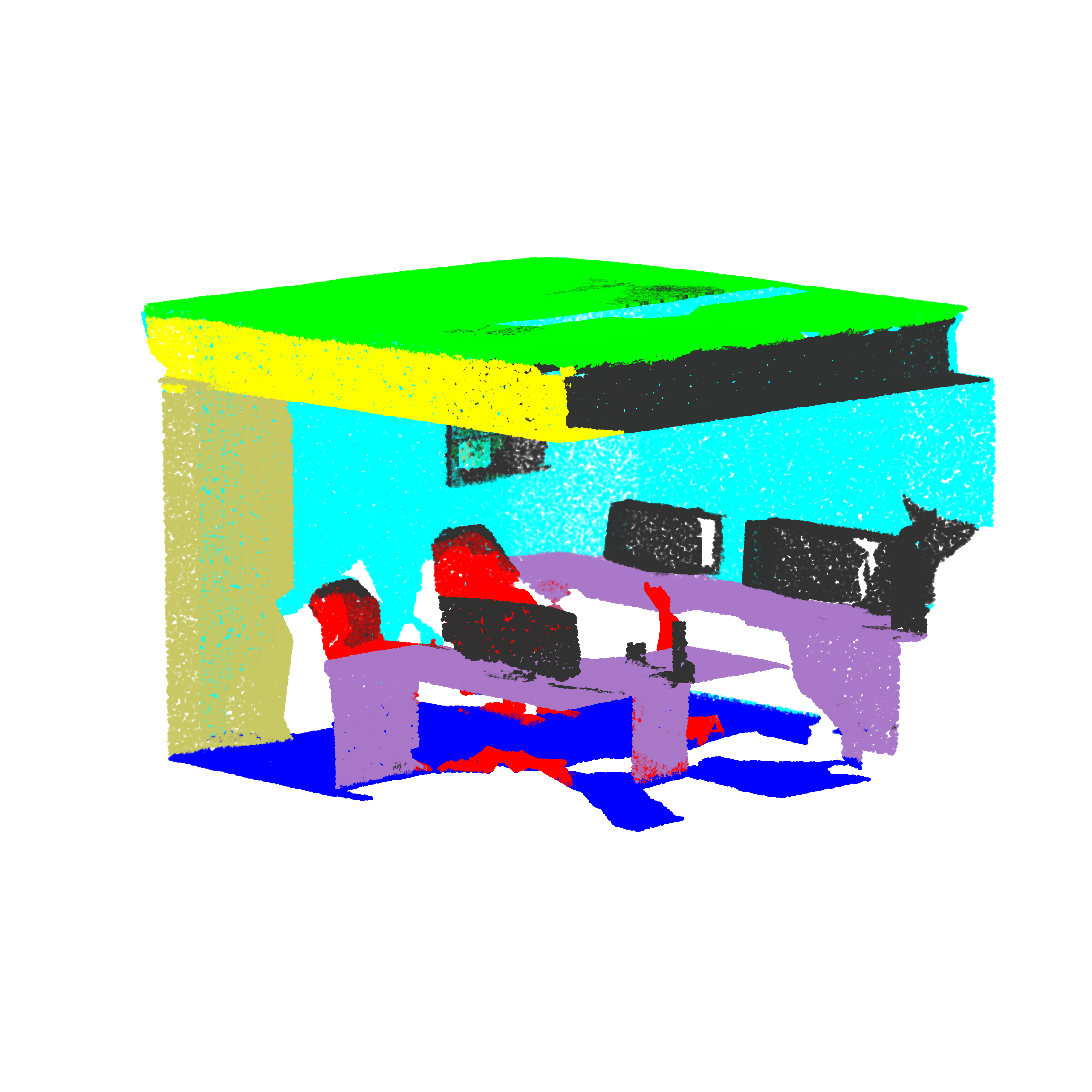}
    \end{subfigure}
    
    \begin{subfigure}[b]{0.24\textwidth}
        \centering
        \includegraphics[width=\textwidth]{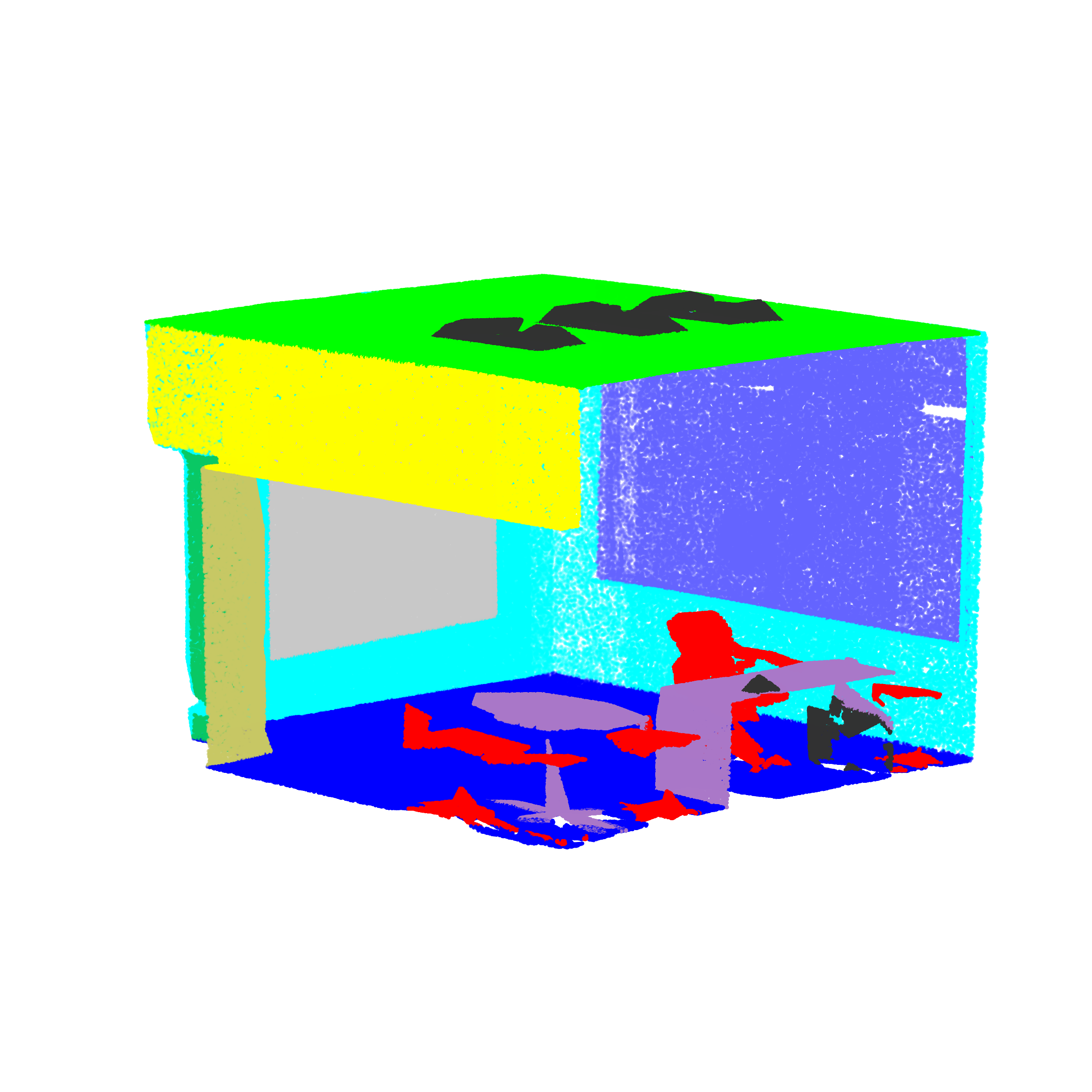}
        \caption*{Ground Truth}
    \end{subfigure}
    \hfill
    \begin{subfigure}[b]{0.24\textwidth}
        \centering
        \includegraphics[width=\textwidth]{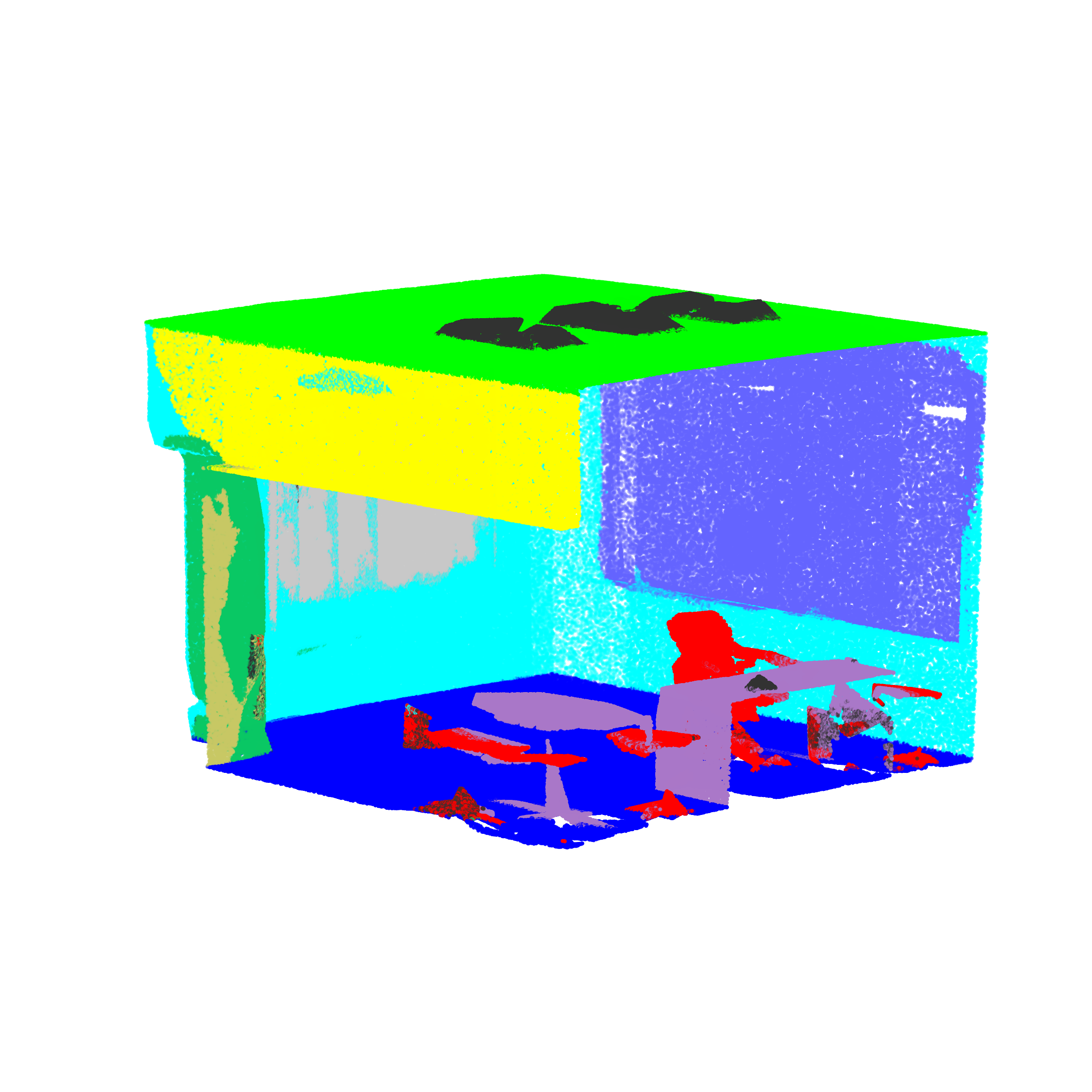}
        \caption*{PointNet}
    \end{subfigure}
    \hfill
    \begin{subfigure}[b]{0.24\textwidth}
        \centering
        \includegraphics[width=\textwidth]{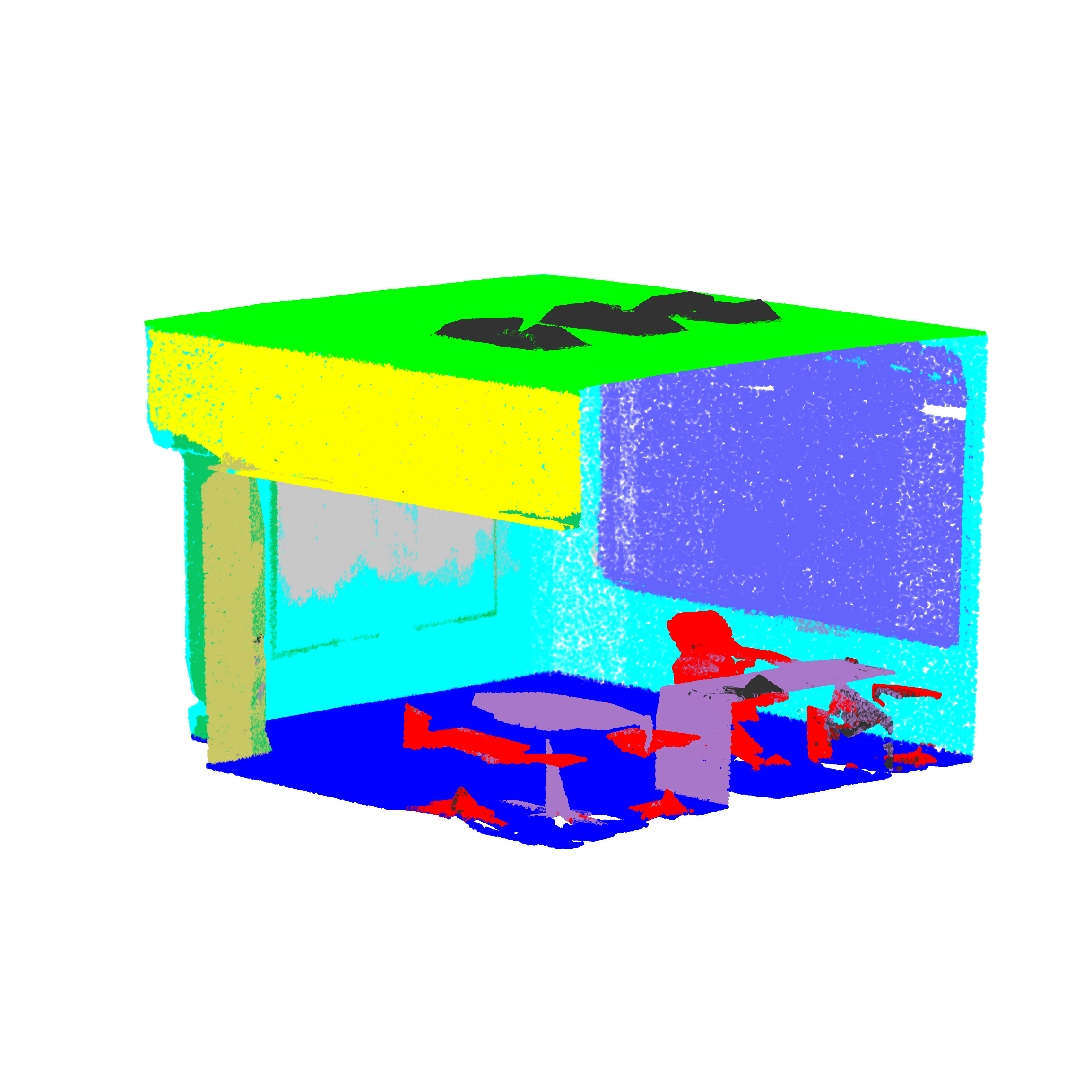}
        \caption*{PointNet-KAN}
    \end{subfigure}
    \hfill
    \begin{subfigure}[b]{0.24\textwidth}
        \centering
          \includegraphics[width=\textwidth]{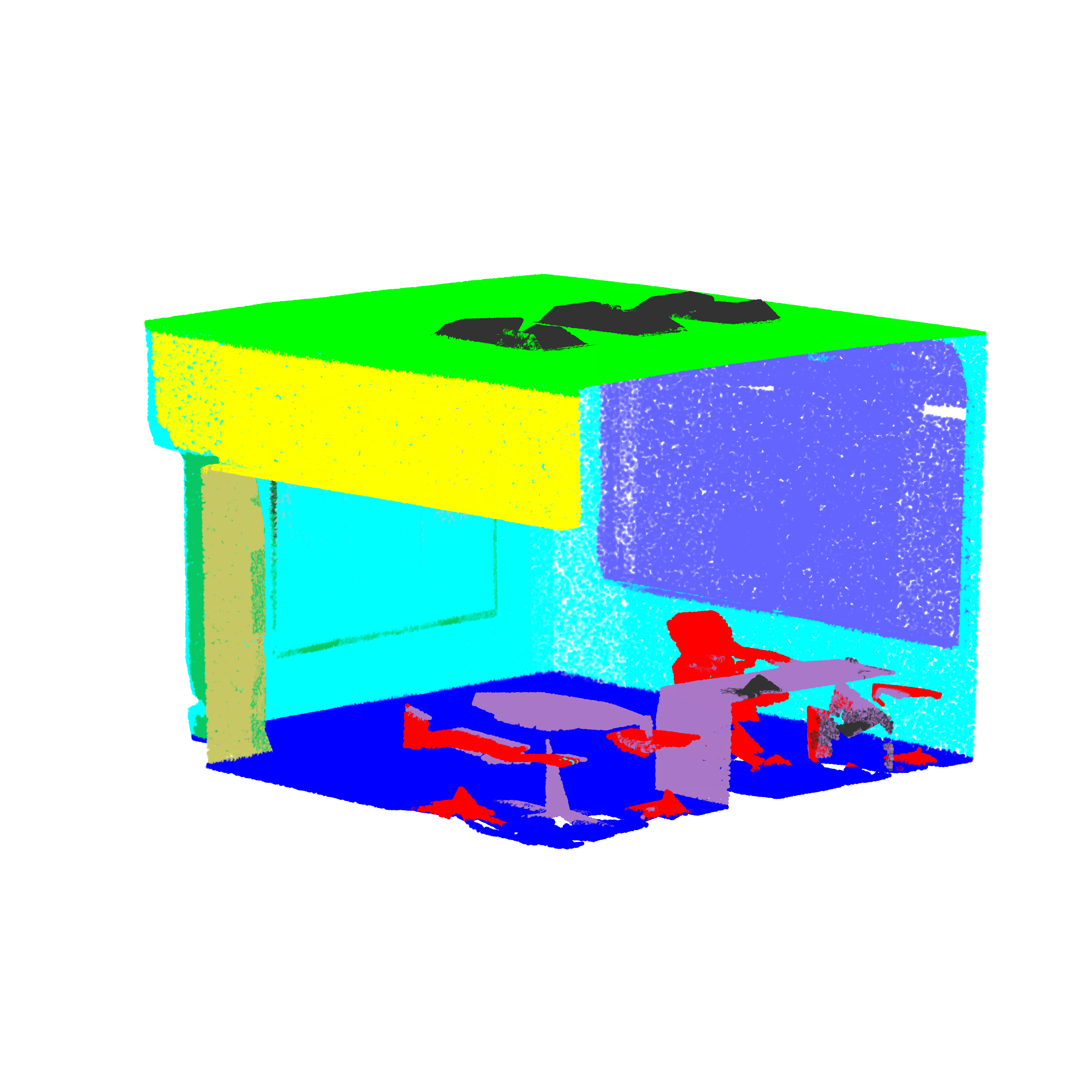}
        \caption*{PointNet-KAN-MLP}
    \end{subfigure}

\caption{A few qualitative results obtained by PointNet, PointNet-KAN, and PointNet-KAN-MLP for semantic segmentation on the Stanford 3D semantic parsing (S3DIS) dataset \citep{armeni20163d}. In PointNet-KAN and PointNet-KAN-MLP, a Jacobi polynomial of degree 2 with $\alpha = \beta = -0.5$ is used.}
    \label{Fig30}
\end{figure}

For the semantic segmentation task, we evaluate PointNet-KAN and PointNet-KAN-MLP (discussed in Sect. \ref{Sect42}) on the Stanford 3D semantic parsing (S3DIS) dataset \citep{armeni20163d}, which consists of 3D scans captured using scanners across six areas, encompassing a total of 271 rooms. Each point in the dataset is labeled with one of 13 semantic categories, including common objects such as chairs, tables, floors, walls, as well as a category for clutter. In line with the PointNet \citep{qi2017pointnet} methodology, we partition the point cloud by room and further divide each room into 1-meter by 1-meter blocks. Each point is represented by 9 features of three-dimensional spatial coordinates, RGB color values, and a normalized location within the room, ranging from 0 to 1. Furthermore, 4096 points are sampled from each block. Following the protocol established by \citet{armeni20163d} and \citet{qi2017pointnet}, we use the same 6-fold cross-validation strategy for training and testing. In particular, we use a Jacobi polynomial of degree 2 with $\alpha = \beta = -0.5$ (i.e., the Chebyshev polynomial of the first kind) to construct KAN layers for this experiment. The results are tabulated in Table \ref{TableX}. Accordingly, we observe that both PointNet-KAN and PointNet-KAN-MLP outperform PointNet in terms of accuracy and mean IoU for both Area 5 and the 6-fold cross-validation criterion. PointNet-KAN-MLP shows a slight improvement over PointNet-KAN, for the same reason discussed in Sect.\ref{Sect42}. Moreover, a qualitative comparison of the semantic segmentation predictions by PointNet, PointNet-KAN, and PointNet-KAN-MLP, along with the ground truth for a conference room and two offices, is shown in Fig. \ref{Fig30}. For instance, compared to PointNet, PointNet-KAN and PointNet-KAN-MLP demonstrate, overall, more accurate segmentations of tables and chairs, as shown in Fig. \ref{Fig30}. This machine learning experiment demonstrates the capacity of KAN layers, when embedded in point-cloud-based neural networks, to perform semantic segmentation in large-scale scenes.

\begin{table}[]
\caption{Results for semantic segmentation on the Stanford 3D semantic parsing (S3DIS) dataset \citep{armeni20163d}. Evaluations are reported using 6-fold cross-validation or on Area 5. In PointNet-KAN and PointNet-KAN-MLP, a polynomial of degree 2 (i.e., $n = 2$ in Eq. \ref{Eq3}) with $\alpha = \beta = -0.5$ is used.}
\label{TableX}
\vspace{1mm}
\begin{adjustbox}{width=1\textwidth}
\begin{tabular}{c|cc|cc}
\toprule
 & \multicolumn{2}{c|}{6-Fold} & \multicolumn{2}{c}{Area 5} \\
& \multicolumn{2}{c|}{------------------------------------} & \multicolumn{2}{c}{------------------------------------} \\
 & mean IoU & overall accuracy & mean IoU & overall accuracy \\
\midrule
PointCNN \citep{li2018pointcnn} & 65.4  & - & - & - \\
DGCNN \citep{wang2019dynamic} & 56.1 & 84.1 & - & - \\
Point Transformer \citep{zhao2021point} & 73.5 & 90.2 & 70.4 & 90.8  \\
PointNeXt-XL \citep{qian2022pointnext} & 74.9 & 90.3 & 70.5 & 90.6 \\
MH-V \citep{feng2024shape2scene} & - & - & 70.3 & -  \\
\midrule
PointNet (baseline) \citep{qi2017pointnet} & 20.1 & 53.2 & - & -  \\
PointNet  \citep{qi2017pointnet} &  47.7 & 78.6 &  41.1 & - \\
\midrule
PointNet-KAN &  51.3 &  81.8  &  44.7 & 83.9  \\
PointNet-KAN-MLP &  51.5  &  82.4 &   44.5  &  84.3   \\
\bottomrule
\end{tabular}
\end{adjustbox}
\end{table}


\subsection{A more advanced network: PointNet++ with KAN and MLP}
\label{Sect47}

In the previous sections, we analyzed PointNet-KAN, which incorporates two key principles for processing unordered point sets: shared layers and symmetric functions. In this section, we further investigate the capability of KAN layers by integrating them into PointNet++, a more advanced point-based neural network. PointNet++, proposed by \citep{qi2017pointnet++}, is a hierarchical architecture that outperforms the original PointNet by capturing local geometric features at multiple scales via set abstraction modules (see Fig. 2 in \citep{qi2017pointnet++}). As discussed earlier, more sophisticated networks often include a variety of nonlinear and hierarchical operators, which can potentially obscure the isolated effect of substituting KAN layers for traditional MLPs. With this in mind, we evaluate the performance of PointNet++ with KAN layers for a classification task. Our results indicate that the standalone replacement of MLPs with KANs in the standard PointNet++ architecture \citep{qi2017pointnet++} does not yield a notable improvement in classification accuracy on the ModelNet40 dataset \citep{wu20153d}. However, our machine learning experiments show that combining KAN layers with MLP layers enhances overall performance. Specifically, we propose a hybrid architecture, PointNet-KAN-MLP++, in which shared KAN layers are used within the set abstraction modules for feature extraction, while MLP layers are retained in the classification head.

The results are presented in Table \ref{Table1}. The overall accuracy of PointNet-KAN-MLP++, 92.9\%, surpasses that of PointNet (89.2\%), PointNet-KAN (90.5\%), and PointNet++ (91.9\%). This observation closely aligns with the findings discussed in the final paragraph of Sect. \ref{Sect42}, where we analyzed the performance of PointNet-KAN-MLP in detail. In fact, KAN layers appear to perform better than MLP layers within the set abstraction modules for feature extraction. On the other hand, shared MLPs are less prone to overfitting and are more effective at mapping global features to classification scores. This explains why the performance of PointNet++ with KAN layers alone is comparable to the original PointNet++ with MLPs (91.8\% versus 91.9\%). As a result, combining the two can enhance the performance of PointNet-KAN-MLP++ compared to PointNet++ architectures that rely solely on either MLPs or KANs.

Finally, for the sake of completeness, we provide a brief overview of the PointNet‑KAN‑MLP++ architecture. PointNet‑KAN‑MLP++ augments the hierarchical feature‑learning paradigm of PointNet++ by replacing each shared MLP in its set abstraction (SA) modules with a shared KAN layer. In shared KAN layers, the Jacobi polynomial degree is set to 2 (i.e., $n=2$) with $\alpha = \beta = -0.5$. In the first SA stage, 512 centroids are sampled via farthest‑point sampling. For each centroid, using single‑scale grouping via a fixed‑radius ball query, up to 32 neighboring points within a radius of $0.2$ are grouped and normalized relative to the centroid’s coordinates. Each grouped patch, with shape $[\mathcal{B} \times d \times n_b]$, is then processed by three stacked shared KAN layers with output dimensions $[64,64,128]$, each followed by batch normalization. Here, $\mathcal{B}$ denotes the batch size and $n_b$ is the number of points in each local neighborhood (e.g., 32). A max‑pooling operation along the neighbor dimension yields a 128‑dimensional feature vector per centroid. This process is repeated at increasingly coarser scales: 128 centroids with a radius of $0.4$ and KAN layers $[128,128,256]$, followed by a global grouping using KAN layers $[256,512,1024]$, resulting in a 1024‑dimensional global descriptor for the entire point cloud. After building this SA hierarchy, the resulting $\mathcal{B} \times1024$ feature tensor is fed into a compact classifier head comprising three sequential MLP layers. The first layer projects from 1024 to 512 coefficients, followed by batch normalization and 40\% dropout; the second layer reduces dimensionality from 512 to 256 with the same normalization and dropout scheme. A final MLP layer then maps these 256‑dimensional embeddings to the desired number of output classes. For further explanation of technical terms such as set abstraction, centroid, single‑scale grouping, ball query, and farthest‑point sampling, see \cite{qi2017pointnet++}.


\begin{table}[h]
\caption{Mean IoU results of PointNet-KAN for part segmentation on ShapeNet part dataset \citep{yi2016scalable} for different Jacobi polynomial degrees ($n$) with $\alpha = \beta = 1$.}
\label{TableB}
\vspace{1mm}
\begin{adjustbox}{width=1\textwidth}
\begin{tabular}{l|c|cccccccccccccccc}
\toprule
& Mean & aero & bag & cap & car & chair & ear & guitar & knife & lamp & laptop & motor & mug & pistol & rocket & skate & table \\
& IoU &  &  & &  &  & phone &  &  &  &  &  & &  &  & board & \\
\midrule
\# shapes & & 2690 & 76 & 55 & 898 & 3758 & 69 & 787 & 392 & 1547 & 451 & 202 & 184 & 283 & 66 & 152 & 5271  \\
\midrule
$n = 2$ & 82.8 & 81.1 & 76.8 & 78.7 & 74.4 & 88.4 & 64.8 & 90.5 & 84.5 & 78.8 & 95.0 &  66.9 & 93.0 & 82.3 & 56.8 & 73.5 & 80.7 \\
$n = 3$ & 81.8 & 80.0 & 76.3 & 79.6 & 72.1 & 88.0 & 69.4 & 89.0 & 83.0 & 79.4 & 95.0 & 61.5 & 91.3 & 81.0 & 55.3 & 70.0 & 79.0 \\
$n = 4$ & 82.4 & 81.2 & 71.2 & 75.6 & 70.7 & 87.9 & 68.3 & 90.0 & 81.8 & 78.4 & 94.0 & 60.7 & 90.7 & 80.1 & 51.3 & 70.8 & 81.7 \\
$n= 5$ & 80.7 & 78.2 & 72.0 & 79.0 & 67.8 & 87.5 & 68.9 & 87.6 & 81.3 & 76.6 & 94.5 & 60.8 & 88.0 & 81.0 & 47.3 & 69.3 & 79.3 \\
$n = 6$ & 82.2 & 80.5 & 70.8 & 78.0 & 71.7 & 87.5 & 62.5 & 88.0 & 82.7 & 76.8 & 94.6 & 62.8 & 92.0 & 78.9 & 48.7 & 65.9 & 81.6 \\
\bottomrule
\end{tabular}
\end{adjustbox}
\end{table}


\begin{table}
\caption{Mean IoU results of PointNet-KAN for part segmentation on ShapeNet part dataset \citep{yi2016scalable} for different values of $\alpha$ and $\beta$. In PointNet-KAN, the Jacobi polynomial degree is set to 2 (i.e., $n=2$). Note that $\alpha = \beta = 0$ corresponds to the Legendre polynomial, $\alpha = \beta = -0.5$ corresponds to the Chebyshev polynomial of the first kind, $\alpha = \beta = 0.5$ corresponds to the Chebyshev polynomial of the second kind, and, in general, $\alpha = \beta$ corresponds to the Gegenbauer polynomial.}
\label{TableC}
\vspace{1mm}
\begin{adjustbox}{width=1\textwidth}
\begin{tabular}{l|c|cccccccccccccccc}
\toprule
& Mean & aero & bag & cap & car & chair & ear & guitar & knife & lamp & laptop & motor & mug & pistol & rocket & skate & table \\
& IoU &  &  & &  &  & phone &  &  &  &  &  & &  &  & board & \\
\midrule
\# shapes & & 2690 & 76 & 55 & 898 & 3758 & 69 & 787 & 392 & 1547 & 451 & 202 & 184 & 283 & 66 & 152 & 5271  \\
\midrule
$\alpha = \beta = 0$ & 83.1 & 82.0 & 73.5 & 80.2 & 75.4 & 88.5 & 68.9 & 90.4 & 83.9 & 80.6 & 95.2 & 65.3 & 92.7 & 81.2 & 56.9 & 72.4 & 80.9 \\

$\alpha = \beta = -0.5$ & 83.3 & 81.0 & 76.8 & 79.8 & 74.6 & 88.7 & 65.4 & 90.9 & 85.3 & 79.9 & 95.0 & 65.3 & 93.0 & 83.0 & 54.3 & 71.9 & 81.6\\

$\alpha = \beta = 0.5$ & 81.7 & 80.5 & 74.9 & 78.9 & 69.3 & 87.5 & 66.3 & 89.5 & 84.1 & 77.3 & 95.0 & 64.5 & 92.0 & 81.7 & 53.1 & 71.3 & 79.7 \\

$\alpha = \beta = 1$ & 82.8 & 81.1 & 76.8 & 78.7 & 74.4 & 88.4 & 64.8 & 90.5 & 84.5 & 78.8 & 95.0 &  66.9 & 93.0 & 82.3 & 56.8 & 73.5 & 80.7 \\

$2\alpha = \beta = 2$ & 82.6 & 81.0 & 75.8 & 81.5 & 72.1 & 88.1 & 68.0 & 90.9 & 83.5 & 79.5 & 95.2 & 63.2 & 91.2 & 80.5 & 58.2 & 74.0 & 80.8 \\

$\alpha = 2\beta = 2$ & 82.5 & 81.0 & 73.3 & 82.4 & 71.6 & 88.3 & 68.5 & 90.7 & 84.3 & 79.3 & 95.4 & 64.2 & 91.3 & 81.9 & 54.6 & 70.4 & 80.5 \\
\bottomrule
\end{tabular}
\end{adjustbox}
\end{table}


\subsection{Ablation studies}
\label{Sect44}

\paragraph{Influence of polynomial type and polynomial degree}

Concerning the classification task discussed in Sect. \ref{Sect41}, Table \ref{Table2} illustrates the effect of varying the polynomial degree from 2 to 6, with $\alpha = \beta = 1$ held constant. While increasing the degree does not significantly affect accuracy, it does increase the number of trainable parameters. Moreover, Table \ref{Table3} reports the results of varying $\alpha$ and $\beta$ with a fixed polynomial degree of 2, showing that different Jacobi polynomial types do not significantly impact performance. Concerning the segmentation task discussed in Sect. \ref{Sect42}, we investigate the effect of the Jacobi polynomial degree and the roles of $\alpha$ and $\beta$ on performance. The results are tabulated in Table \ref{TableB} and \ref{TableC}. Similar to the classification task discussed in Sect. \ref{Sect41}, no significant differences are observed. As shown in Table \ref{TableB}, increasing the degree of the Jacobi polynomial does not improve prediction accuracy. According to Table \ref{TableC}, the best performance is achieved with the Chebyshev polynomial of the first kind when $\alpha = \beta = -0.5$.


\paragraph{Influence of the size of tensors and global feature}

We investigate the effect of the size of the tensor $\tA$ (see Eq. \ref{Eq2}) and, consequently, the size of the global feature on prediction accuracy. In the classification branch (see Fig. \ref{Fig1}), choosing the shared KAN layer with the size of 1024 (i.e., $\tA_{1024\times 6}$ and global feature size of 1024) and 2048 (i.e., $\tA_{2048\times 6}$ and global feature size of 2048) results in the overall accuracy of 89.7\% and 90.3\%, respectively, for the ModelNet40 \citep{wu20153d} benchmark. In the segmentation branch (see Fig. \ref{Fig1}), there are four shared KAN layers, each corresponding to a tensor. From left to right, we refer to them as $\tB$, $\tC$, $\tD$, and $\tE$. For example, selecting the sets $\tB_{128 \times 3}$, $\tC_{1024 \times 128}$, $\tD_{128 \times1153}$, $\tE_{50 \times128}$ and $\tB_{384 \times 3}$, $\tC_{3072\times 384}$, $\tD_{3457\times 3457}$, $\tE_{50 \times 384}$, respectively, results in a mean IoU of 82.6\% and 82.2\% for the ShapeNet part \citep{yi2016scalable} benchmark. Note that the size of the global feature in the segmentation branch is determined by the number of rows ($d_{\text{output}}$) in tensor $\tC$.

\begin{wrapfigure}{r}{5.5cm}
    \centering
    \vspace{-2mm}
    \includegraphics[width=1\linewidth]{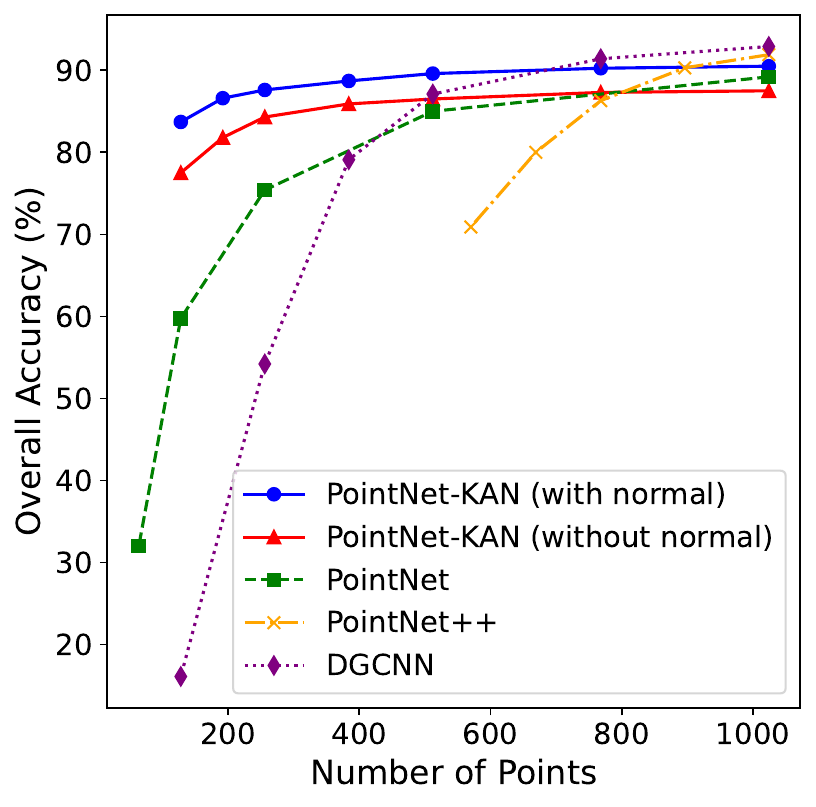}
    \vspace{-6mm}
    \caption{Robustness test for PointNet-KAN on the ModelNet40 \citep{wu20153d} test case, where input points are randomly dropped. See text for details.}
    \vspace{-2mm}
    \label{Fig3}
\end{wrapfigure}

\paragraph{Robustness} 

Figure \ref{Fig3} shows the overall accuracy on the ModelNet40 \citep{wu20153d} benchmark when input points from the test set are randomly dropped. PointNet-KAN (with $\alpha = \beta = 1$, $n = 4$) demonstrates relatively stable performance as the number of points decreases from 1024 to 128, with accuracy gradually dropping from 90.5\% to 83.7\% when using normal vectors ($d = 6$), and from 87.5\% to 77.5\% without normal vectors ($d = 3$). Interestingly, PointNet-KAN shows stronger stability compared to other models \citep{qi2017pointnet,qi2017pointnet++,wang2019dynamic}, as indicated in Fig. \ref{Fig3}.


\paragraph{Influence of input and feature transform networks and deeper architectures}

In Sect. \ref{Sect41} and Sect. \ref{Sect42}, we pointed out that PointNet-KAN is effective, despite its simple and shallow architecture, and the absence of input and feature transform networks. A question arises: if such a simple structure performs well, why not improve PointNet-KAN's performance by deepening the network and adding input and feature transform networks to achieve even better results? To answer this question, a straightforward approach is to replace all MLPs in the PointNet architecture (see Fig. 2 of \cite{qi2017pointnet} for the classification branch and Fig. 9 of \cite{qi2017pointnet} for the segmentation branch) with KAN to create an equivalent model. We conduct this experiment as follows. We utilize KAN layers with a Jacobi polynomial degree of 2 (i.e., $n = 2$) and parameters $\alpha = \beta = 1$. The size of the sequential KAN layers is chosen to match the corresponding size of the MLPs in PointNet, such as (64, 64), (64, 128, 1024), and so on, as illustrated in \cite{qi2017pointnet}. To conserve space, we omit sketching the full network architecture again. Interestingly, the network's performance does not improve. The overall accuracy of classification on ModelNet40 \citep{wu20153d} is 88.9\% and the mean IoU on the ShapeNet part dataset \citep{yi2016scalable} is 82.1\%.


\section{Summary}
\label{Sect6}

In this work, we proposed, for the first time, PointNet with shared KANs (i.e., PointNet-KAN) and compared its performance to PointNet with shared MLPs. Our results demonstrated that PointNet-KAN achieved competitive performance to PointNet in both classification and segmentation tasks, while using a simpler and much shallower network compared to the deep PointNet with shared MLPs. In our implementation of shared KAN, we compared various families of the Jacobi polynomials, including Lagrange, Chebyshev, and Gegenbauer polynomials, and observed no significant differences in performance among them. Additionally, we found that a polynomial degree of 2 was sufficient. PointNet-KAN exhibited greater stability and robustness than PointNet \citep{qi2017pointnet}, PointNet++ \citep{qi2017pointnet++}, and DGCNN \citep{wang2019dynamic} as the number of input points decreased in classification tasks. Thus, its lightweight design relative to these networks makes it better suited for real-time inference under sparse‐input conditions. Additionally, we explored the use of KANs as encoders and MLPs as decoders or classifiers within PointNet, motivated by the desire to leverage the strengths of both components in a hybrid architecture, and found that this version achieves better performance than both PointNet and PointNet-KAN in classification and segmentation tasks. As an example of an advanced point cloud neural network, we evaluated the performance of PointNet++ \citep{qi2017pointnet++} integrated with KANs in the encoder and MLPs in the classification head for object classification, and found that it outperforms the original PointNet++ architecture. We hope this work lays a foundation and offers insights for incorporating KANs, as an alternative to MLPs, into more advanced architectures for point cloud deep learning frameworks.


\subsubsection*{Reproducibility Statement}

The Python code is available on the following GitHub repository at \url{https://github.com/Ali-Stanford/PointNet_KAN_Graphic}.

\subsubsection*{Acknowledgment}


We express our gratitude to the Sherlock Computing Center at Stanford University for providing the computational resources that supported this research. We sincerely thank the reviewers for their insightful comments and suggestions, which have contributed to enhancing the quality of this article.

\bibliography{iclr2025_conference}
\bibliographystyle{iclr2025_conference}

\newpage

\appendix
\section{Supplementary materials}

\subsection{Training Details}
\label{ATD}

The models for both classification and segmentation are implemented using PyTorch. For classification tasks, a batch size of 64 is used, while part and semantic segmentation uses a batch size of 32. The training process employs the Adam optimizer, configured with $\beta_1=0.9$, $\beta_2 = 0.999$, and $\hat{\epsilon} = 10^{-8}$. An initial learning rate of 0.0005 and 0.001 is chosen respectively for the classification and part and semantic segmentation tasks. To progressively decrease the learning rate during training, a learning rate scheduler is applied, which reduces the learning rate by a factor of 0.5 after every 20 epochs. The cross-entropy loss function is used. All experiments run on an NVIDIA A100 Tensor Core GPU with 80 GB of RAM.


\end{document}